%% file: SMVAGC-SF.tex
\documentclass[lettersize,journal]{IEEEtran}
\usepackage{amsmath,amsfonts}
\usepackage{algorithmic}
\usepackage{algorithm}
\usepackage{array}
\usepackage{textcomp}
\usepackage{stfloats}
\usepackage{url}
\usepackage{verbatim}
\usepackage{graphicx}
\usepackage{cite}
\usepackage[colorlinks,linkcolor=black,anchorcolor=black,citecolor=black,]{hyperref}
\usepackage{times}
\usepackage{url}
\usepackage{ulem}
\usepackage[utf8]{inputenc}
\usepackage[small]{caption}
\usepackage{algorithm}
\usepackage{algorithmic}
\usepackage{amsmath}
\usepackage{bm}
\usepackage{booktabs,amssymb}
\usepackage{multirow}
\usepackage{subfigure}
\usepackage[table,xcdraw]{xcolor}
\usepackage{epstopdf}
\usepackage{float}
\usepackage{stackengine}
\usepackage[table,xcdraw]{xcolor}
\usepackage{cite}
\useunder{\uline}{\ul}{}

\renewcommand{\algorithmicrequire}{\textbf{Input:}} 
\renewcommand{\algorithmicensure}{\textbf{Output:}} 


\begin{document}

\title{Fast Continual Multi-View Clustering with Incomplete Views}

\author{~Xinhang~Wan,~Bin~Xiao,~Xinwang~Liu$^{\dagger}$,~\IEEEmembership{Senior~Member,~IEEE},~Jiyuan~Liu,\\~Weixuan~Liang,~En~Zhu$^{\dagger}$,
\thanks{X. Wan, X. Liu, J. Liu, W. Liang and E. Zhu are with School of Computer, National University of Defense Technology, Changsha, 410073, China. (E-mail: \{wanxinhang,\, xinwangliu,\, weixuanliang,\, liujiyuan13,\, enzhu\} @nudt.edu.cn).}
\thanks{B. Xiao is with School of Computer Science and Technology, Chongqing University of Posts and Telecommunications, Chongqing, 400065, China. (E-mail: xiaobin@cqupt.edu.cn).}}
\markboth{Submitted to IEEE Transactions on Image Processing, Month June, Year 2023}%
{Shell \MakeLowercase{\textit{ Wan et al.}}: Fast Continual Multi-View Clustering with Incomplete Views}


\maketitle
\begin{abstract}
Multi-view clustering (MVC) has gained broad attention owing to its capacity to exploit consistent and complementary information across views. This paper focuses on a challenging issue in MVC called the incomplete continual data problem (ICDP). In specific, most existing algorithms assume that views are available in advance and overlook the scenarios where data observations of views are accumulated over time. Due to privacy considerations or memory limitations, previous views cannot be stored in these situations. Some works are proposed to handle it, but all fail to address incomplete views. Such an incomplete continual data problem (ICDP) in MVC is tough to solve since incomplete information with continual data increases the difficulty of extracting consistent and complementary knowledge among views. We propose Fast Continual Multi-View Clustering with Incomplete Views (FCMVC-IV) to address it. Specifically, it maintains a consensus coefficient matrix and updates knowledge with the incoming incomplete view rather than storing and recomputing all the data matrices. Considering that the views are incomplete, the newly collected view might contain samples that have yet to appear; two indicator matrices and a rotation matrix are developed to match matrices with different dimensions. Besides, we design a three-step iterative algorithm to solve the resultant problem in linear complexity with proven convergence. Comprehensive experiments on various datasets show the superiority of FCMVC-IV.

\end{abstract}


\section{Introduction}
\begin{figure*}[htbp]
 	\centering
 	\subfigure{
 		\includegraphics[width=0.75\textwidth]{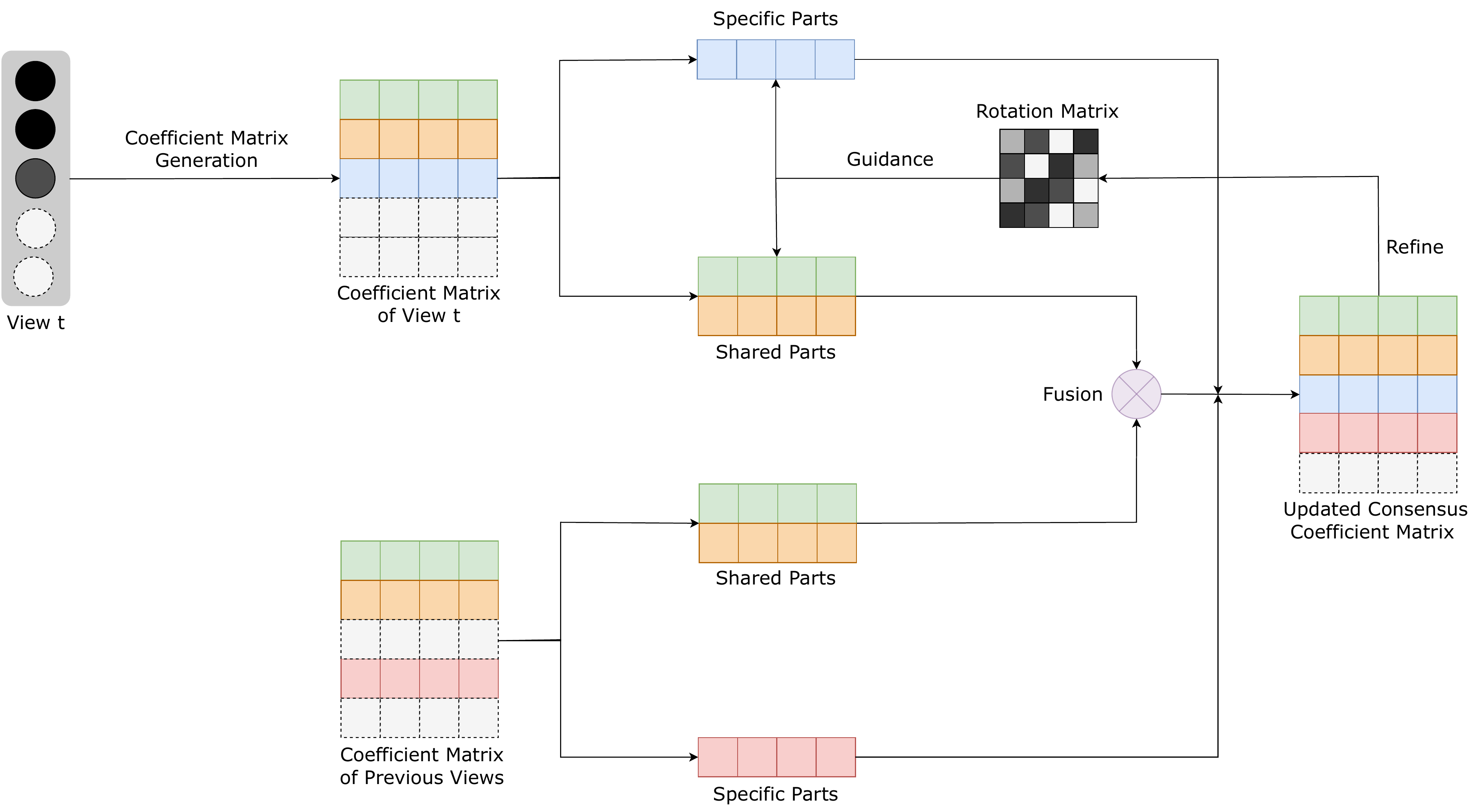}}
 	\caption{The basic framework of our proposed algorithm. Once the $t$-th view is available, the consensus coefficient matrix will be updated by the original and incoming views. Considering that FCMVC-IV can handle incomplete views, the size of the consensus matrix ought to be scalable if the newly collected view contains samples that do not exist in the previous views.
 	}
 	\label{alg_fig}
\end{figure*}

\IEEEPARstart{W}ITH the rapid development of multimedia techniques, data can be described from various modalities \cite{XiaWGZG22, KGESymCL, Zhang2020ConsensusOM,9718038,huang2021joint,9226062,8506433}. For instance, an object could be recognized by an intelligent robot from its eyes (image), ears (sound), and past knowledge (knowledge graph). Integrating multimodal information to make decisions play an essential role in the robot taking action. Also, in many situations, such as the recommender and decision support systems, how to uncover items' intrinsic structure and label them with multi-view data is crucial \cite{zhang2022efficient,yang2022interpolation, XiaWGYG21,9653838,wang2021generative,9623366,9318542}. Multi-view clustering is a promising method to solve the above problems by exploring the consistent and complementary information of different data views and discovering the underlying data structure.

To the best of our knowledge, existing multi-view clustering algorithms can be roughly divided into three popular paradigms, including multi-view subspace clustering (MVSC) \cite{7410839,liu2022efficient,DBLP:conf/acml/SunCMJ19,Jing2015StratifiedFS,wang2015robust,liu2019robust,9447203,8703424,7297854,9390340}, multi-view graph clustering (MVGC) \cite{liu-1,liu-3,9200798,9769920,zhan2018multiview,9780880} and multiple kernel clustering (MKC) \cite{8627941,liang2022LSWMKC}. Among them, MVSC first computes an independent subspace of each view and then fuses them into a unified subspace. For example, Liu \textit{et al.} \cite{9399655} obtain robust representations by performing eigendecomposition on the original data observations, then seek a consensus subspace among them. Nevertheless, most MVGC methods \cite{KANG2020105102,10.5555/2886521.2886704,9774970} transform the data samples of multiple views into corresponding undirected graphs based on their pairwise similarities, then perform spectral clustering and graph fusion simultaneously.
In MKC \cite{10.5555/2832581.2832733,9212617}, one learns the weights of base kernels to obtain a linear kernel combination as the optimal kernel for the final clustering results.

Although the above three types of algorithms have achieved great success, there are still plenty of challenging problems. One primary concern is the relatively high time and space complexity. Denote the number of data samples as $n$, most MVSC methods require $\mathcal{O}\left(n^{2}\right)$ space to store the consensus subspace structure and $\mathcal{O}\left(n^{3}\right)$ time to perform subspace decomposition, making it hard to be applied on large-scale tasks. To handle large-scale data, a series of MVC methods based on non-negative matrix factorization (NMF) is proposed \cite{liu2013multi, Gao2019MultiviewLM,yang2020fast}. As pointed out by \cite{10.1007/978-3-319-23528-8_20}, NMF aims to factorize data matrices from multiple sources into two non-negative components, i.e., a consensus coefficient matrix and view-specific base matrices. The authors in \cite{Jing2017Diverse} 
develop a diverse NMF method to enhance the diversity among multi-view representations. In \cite{2020Auto}, Nie \textit{et al.} propose a new co-clustering method to shrink the information loss in matrix factorization.

Despite existing NMF-based MVC methods demonstrating promising clustering performance, they still encounter two main problems, leading to their limited applications. First, real-time data is ignored (the data observations of new views are accumulated over time). In a brain-computer interface system, the signal data at various moments will constitute a new view \cite{YIN2021260}. Under face recognition, the images of different perspectives or times will also contribute to a new view. To deal with these real-time data, one prominent approach to current NMF methods is recomputing with all data once a new view arrives, wasting computation resources and time. Second, the assumption that the views are complete. However, in real applications like Alzheimer’s disease prediction \cite{Xiang2013MultisourceLW} and cardiac disease discrimination \cite{6556587}, it is not uncommon to see that some views suffer from the missing information. In a disease diagnosis, some patients may fail to have all necessary medical checkups because of limited financial support. When a new medical device is developed, a new test metric emerges; in other words, a fresh incomplete view arrives. The violation of the assumption that the views are complete and attained in advance prevents these methods from being applied to many situations. 

Unfortunately, solving the incomplete continual data problem (ICDP) is challenging. We summarize the challenges in two areas: (1) How to preserve and exploit the information of the previous views since it is not available with time due to privacy policies or memory burdens? (2) How to handle incomplete views and integrate three parts of information, i.e., the shared items between the previous and incoming views and the specific samples of them, respectively?

In light of this, we propose a novel method termed Fast Continual Multi-View Clustering with Incomplete Views (FCMVC-IV), which combines continual learning and matrix factorization into a unified framework. The illustration of our framework is shown in Figure \ref{alg_fig}. Specifically, FCMVC-IV only maintains a consensus coefficient matrix, which integrates all previous views information. Once a newly collected view is collected, FCMVC-IV updates the consensus coefficient matrix with the incoming view rather than keeping and re-computing all of the previously learned data matrices. Given that the views are incomplete, the newly collected view might contain samples that have yet to appear. We leverage two indicator matrices and a rotation matrix to match matrices of different dimensions. FCMVC-IV efficiently fuses the data because only two matrices are integrated per time. Since no prior knowledge is brought with the incoming view, allocating the two pieces of knowledge with the same weight is reasonable. Overall, the main contributions are summarized as follows:

\begin{enumerate}
\item We study a new paradigm for large-scale multi-view clustering called incomplete continual data problem (ICDP). The incomplete continual views make consistent and complementary information among views hard to extract and affect the clustering performance.
\item We propose FCMVC-IV to tackle the ICDP problem and will inspire future research. By maintaining a consensus coefficient matrix and fusing only two matrices each time, FCMVC-IV is flexible and effectual.
\item We propose an efficient three-step alternate optimization strategy with proven convergence. Extensive experiments on multiple datasets demonstrate the superiority of FCMVC-IV.
\end{enumerate}

We end this section by discussing the differences between our work and the work in \cite{10.1145/3503161.3547864} termed continual multi-view clustering (CMVC). Both methods combine multi-view clustering with continual learning in a novel way. However, the significant differences can be summarized as follows: 1) CMVC combines late fusion and continual learning into a unified step. Since it utilizes kernel matrices to obtain partition matrices, the time complexity of initialization is $\mathcal{O}\left(n^{2}\right)$. FCMVC-IV is under a matrix-factorization framework, and the time complexity is linear to the sample number. 2) FCMVC-IV extends the range of applications of CMVC and can handle both complete and incomplete real-time multi-view data, while CMVC merely deals with complete real-time multi-view data. 3) We conduct more comprehensive experiments on both complete and incomplete datasets, verifying the excellence of FCMVC-IV. More importantly, FCMVC-IV almost consistently exceeds CMVC on complete datasets.

\section{Related Work}
In this section, we will briefly introduce the most related research, including NMF-based multi-view clustering and NMF-based incomplete multi-view clustering.

\subsection{NMF-Based Multi-View Clustering}
Given a multi-view dataset with $n$ instances and $V$ views $\left\{\mathbf{X}^{v}\right\}_{v=1}^{V}$, in which $\mathbf{X}^{v} \in \mathbb{R}^{d_{v} \times n}$ and $d_{v}$ denotes its feature dimension, NMF-based multi-view clustering aims to factorize the data matrices $\left\{\mathbf{X}^{v}\right\}_{v=1}^{V}$ into two nonnegative components, i.e., view-specific base matrices and coefficient matrices. One of the most representative formulas is as follows \cite{8030316},

\begin{equation}\label{two_step_MFMVC}
\begin{aligned}
&\min _{\mathbf{H}, \mathbf{Z}}\sum_{v=1}^{V}\left\|\mathbf{X}^{v}-\mathbf{H}^{v} \mathbf{Z}^{v}\right\|_{F}^{2}
&\text { s.t. }   \mathbf{H}^{v}, \mathbf{Z}^{v} \geq \mathbf{0},
\end{aligned}
\end{equation}
where $ \mathbf{H}^{v}, \mathbf{Z}^{v} \geq \mathbf{0}$ indicates all elements of them are nonnegative.

The consensus coefficient matrix $\mathbf{Z}$ is obtained via the average value of $\mathbf{Z}^{v}$. On the contrary, some methods \cite{2019Multi,7837980} have been proposed to directly attain a common underlying matrix across views, shown as follows,
\begin{equation}\label{one_step_MFMVC}
\begin{aligned}
&\min _{\mathbf{H} \geq 0, \mathbf{Z} \geq 0} \sum_{v=1}^V\left\|\mathbf{X}^v-\mathbf{H}^{v} \mathbf{Z}\right\|_F^2+\lambda h(\mathbf{Z}), \\&\text { s.t. }  h(\mathbf{Z})=\operatorname{Tr}\left[\mathbf{Z}^{ \top}\left(\sum_{v=1}^V \alpha^v \mathbf{L}^v\right) \mathbf{Z}\right],
\end{aligned}
\end{equation}
where $\mathbf{L}^v$ denotes the Laplacian matrix of $v$-th view and $\alpha^v$ is its weight. After attaining the consensus coefficient matrix $\mathbf{Z}$, a subsequent standard $k$-means is performed to get the final cluster assignments.

Due to the satisfactory clustering performance and low time complexity, a group of MVC methods has been proposed based on NMF. Cai \textit{et al.} offer a new robust large-scale multi-view clustering method to integrate heterogeneous representations of large-scale data \cite{2013Multi}. Unlike the methods mentioned above, Ding \textit{et al.}  provide a 3-factor NMF algorithm and observes that orthogonality constraint leads to rigorous clustering interpretation \cite{2006Orthogonal}.

Although existing NMF-based MVC methods can achieve excellent clustering performance, most get trouble handling incomplete views, resulting in their limited application range.

\subsection{NMF-Based Incomplete Multi-View Clustering}
The recent work in \cite{Li2014PartialMC} has extended the existing NMF-based MVC to incomplete multi-view situations. Specifically, it utilizes the shared samples among views to obtain consistent information and the individual samples to attain specific knowledge, which is mathematically fulfilled as,
\begin{equation}\label{one_step_MFIMVC}
\begin{aligned}
&\min _{\left\{\mathbf {H}^{(v)}, \bar {\mathbf{Z}}^{(v)}\right\}_{v=1}^2}\left\|\left[
\mathbf{X}_c^{(1)},
\hat{\mathbf{X}}^{(1)}
\right]-\mathbf{H}^{(1)} \left[
\mathbf{Z}_c, \hat {\mathbf{Z}}^{(1)}
\right] \right\|_F^2\\&
+\lambda\left\|\bar{\mathbf{Z}}^{(1)}\right\|_1
+\left\|\left[
\mathbf{X}_c^{(2)}, \hat{\mathbf{X}}^{(2)}
\right]-\mathbf {H}^{(2)}\left[
\mathbf{Z}_c,
\hat{\mathbf{Z}}^{(2)}
\right] \right\|_F^2+\\&
\lambda\left\|\bar{\mathbf{Z}}^{(2)}\right\|_1\\
&\text { s.t. } \mathbf {H}^{(1)} \geq 0, \mathbf {H}^{(2)} \geq 0 \text {, }\bar{\mathbf{Z}}^{(1)} \geq 0, \bar{\mathbf{Z}}^{(2)} \geq 0,
\end{aligned}
\end{equation}
where $\bar {\mathbf{Z}}^{(1)}=\left[ \mathbf{Z}_c, \hat {\mathbf{Z}}^{(1)}\right]$ and $\bar {\mathbf{Z}}^{(2)}=\left[ \mathbf{Z}_c, \hat {\mathbf{Z}}^{(2)} \right]$, in which  $\mathbf{Z}_c$, $\hat {\mathbf{Z}}^{(1)}$, and $\hat {\mathbf{Z}}^{(2)}$ denote the shared samples, the specific samples in the first view and the individual ones in the second view, respectively. The final coefficient matrix is obtained via $\mathbf{Z}=\left[ \mathbf{Z}_c, \hat {\mathbf{Z}}^{(1)},\hat {\mathbf{Z}}^{(2)}\right]$. However, the shared instances among views become rare with the view number increasing. To solve it, Shao \textit{et al.} fill the incomplete views with the average features, then develop a weighted NMF \cite{Shao2015MultipleIV}, where the weight factor is evaluated by the missing ratio of each view. To further eliminate the influence of missing samples, Hu \textit{et al.} incorporate the global information among views with the help of regression\cite{ijcai2018p313}, which can be defined as,
\begin{equation}\label{MFIMVC}
\begin{aligned}
\mathcal{J}=& \sum_{i=1}^{V}\left\{\left\|\left(\mathbf{X}^{(i)}-\mathbf{H}^{(i)} \mathbf{Z}^T\right) \mathbf{W}^{(i)}\right\|_F^2\right.\\
&\left.+\alpha\left(\left\|\mathbf{B}^{(i)^T} \mathbf{H}^{(i)}-\mathbf{I}\right\|_F^2+\beta\left\|\mathbf{B}^{(i)}\right\|_{2,1}\right)\right\}, \\
\text { s.t. } & \mathbf{Z} \geq 0,
\end{aligned}
\end{equation}
where $\alpha$ and $\beta$ are balanced hyper-parameters, $\mathbf{B}^{(i)} \in \mathbb{R}^{d_i \times k}$ and $\mathbf{W}^{(i)}$ are used to denote the regression matrix and the weight matrix of $i$-th view, respectively.

Although researchers have developed many methods to handle incomplete views \cite{10.5555/3060832.3060956, Lian2021PartialMC, Hu2019OnePassIM}, most suffer from the following non-negligible drawbacks. First, the simple imputation strategies and unreasonable view missing assumptions. For instance, MIC \cite{10.1007/978-3-319-23528-8_20} fills the missing samples with the average values of complete data. However, it fails to stay stable on large missing ratios. PVC \cite{Li2014PartialMC} enforces some data entries observed in all views, which is impractical when the view number is sizable. Second is the overlook of real-time data. Under the assumption that all the views are accumulated in advance, state-of-the-art algorithms get into trouble handling situations where the number of views is collected over time. The naive way of existing incomplete MVC to take real-time views is to store and recalculate views repeatedly, wasting time and space resources. In the next section, we propose a novel method termed Fast Continual Multi-View Clustering with Incomplete Views (FCMVC-IV) to solve these problems.

\section{Fast Continual Multi-View Clustering with Incomplete Views}
In this section, we first provide the proposed formulation of FCMVC-IV. Then we develop an alternating three-step optimization method with proven convergence. Furthermore, the time and space complexity are discussed to illustrate the superiority of the proposed method.
\subsection{The Proposed Formulation}
As mentioned above, incomplete MVC has trouble dealing with the situation where the number of views increases over time. In some scenarios, views cannot be stored due to privacy policies or memory burdens. Considering that the views might be incomplete due to the missing samples, the new coming view could contain new samples that do not exist in the previous views, making the problem tougher. Suc? an incomplete continual data problem (ICDP) in MVC is tough to solve since incomplete information increases the difficulty of extracting consistent and complementary knowledge among views. To address this issue, we propose FCMVC-IV, which conducts continual MVC with incomplete views via matrix factorization.

Since the views are accumulated over time, and some samples might be absent in the previous views, it is arbitrary to learn them without any clues. Unlike existing incomplete MVC algorithms that impute incomplete views, FCMVC-IV optimizes and extends the consensus knowledge with the appearing samples. Specifically, we utilize a consensus coefficient matrix to preserve the learned knowledge and amend it with a fresh view. Suppose that the $t$-th view with $n_t$ data samples is collected and $\mathbf{X}^{t} \in \mathbb{R}^{d_{t}\times n_t}$ denotes its data matrix, while the consensus partition matrix $\mathbf{Z}^{t-1}\in \mathbb{R}^{k\times n_{t-1}^a}$ has been already acquired, where $n_{t-1}^a$ is the number of samples contained in the previous views. The variance of the sample number between the two matrices makes the clustering process intractable. To update the consensus coefficient matrix, we define two indicator matrices as,
\begin{equation}\label{define_M1}
\begin{aligned}
\mathbf M_{1}^{t}(i,i)= \begin{cases}1, & \text { if } i \text {-th sample is in the } t \text {-th view,} \\ 0, & \text { otherwise, }\end{cases}
\end{aligned}
\end{equation}
and
\begin{equation}\label{define_M2}
\begin{aligned}
\mathbf M_{2}^{t}(i,i)= \begin{cases}1, & \text { if } i \text {-th sample is in the previous views, } \\ 0, & \text { otherwise. }\end{cases}
\end{aligned}
\end{equation}

By two indicator matrices $\mathbf{M}_1^{t}\in \mathbb{R}^{n_t^a\times n_t}$, $\mathbf{M}_2^{t} \in \mathbb{R}^{n_t^a\times n_{t-1}^a}$, and the basic matrix $\mathbf{H}^{t}$, the consensus matrix $\mathbf{Z}^{t}$ can be mapped into the same dimension as $\mathbf{X}^{t}$ and $\mathbf{Z}^{t-1}$, separately. Thus, we can update the optimal consensus coefficient matrix $\mathbf{Z}^{t}$ by minimizing the objective formulation as,
\begin{equation}\label{FCMVC-IV}
\begin{aligned}
&\min _{\mathbf{Z}^t, \mathbf{W}^t, \mathbf{H}^t}   \frac{1}{2} \left\|\mathbf{X}^{t}-\mathbf{H}^{t} \mathbf{Z}^{t}\mathbf{M}_1^t\right\|_{F}^{2} -\operatorname{Tr}\left(\left({\mathbf{Z}^{t} \mathbf{M}_2^t}\right)^{\top} \mathbf{W}^t \mathbf{Z}^{t-1}\right),\\&\text { s.t. } {\mathbf{Z}^{t}} {\mathbf{Z}^{t}}^{\top}=\mathbf{I}_{k}, {\mathbf{H}^{t}}^{\top} {\mathbf{H}^{t}}=\mathbf{I}_{k}, {\mathbf{W}^t}^{\top} {\mathbf{W}^t}=\mathbf{I}_{k},
\end{aligned}
\end{equation}
where $\mathbf{W}^t$ is a rotation matrix to match $\mathbf{Z}^{t} \mathbf{M}_2^t$ and $\mathbf{Z}^{t-1}$. 

In this way, the fusion of previous knowledge and the incoming view guides the generation of the consensus coefficient matrix, and the latter induces the alignment of two pieces of information in turn. Although the model in Eq. (\ref{FCMVC-IV}) seems simple, it holds the following merits:
\begin{itemize}
  \item Flexibility: Instead of keeping and recalculating all views, FCMVC-IV stores a consensus coefficient matrix to update when a newly collected view is available, which is more flexible and can handle real-time data.
  \item Our proposed algorithm combines multi-view matrix factorization and continual learning into a framework. It can handle large-scale data owing to its linear time complexity concerning the sample number.
  \item FCMVC-IV is general and likewise suitable for complete views. By setting $\mathbf{M}_1^{t}$ and $\mathbf{M}_2^{t}$ as identity matrices, the algorithm evolves to handle complete views.
\end{itemize}
\subsection{Optimization}
There are three variables in Eq. \eqref{FCMVC-IV} to optimize. Therefore, updating them simultaneously is an arduous task. To solve the optimization problem, we design a three-step alternating optimization to optimize each variable while the others are fixed. The optimization process is reduced to singular value decomposition (SVD) at each step.

{\bf Solving $\mathbf{Z}^{t}$ with fixed $\mathbf{H}^{t}$ and $\mathbf{W}^t$}. With other variables fixed in Eq. \eqref{FCMVC-IV}, $\mathbf{Z}^{t}$ can be updated by the following formula:
\begin{equation}\label{opt_Z}
\begin{aligned}
\max_{\mathbf{Z}^{t}} \operatorname{Tr}\left({\mathbf{Z}^{t}} \mathbf{A}\right),
&\text { s.t. } {\mathbf{Z}^{t}} {\mathbf{Z}^{t}}^{\top}=\mathbf{I}_{k},
\end{aligned}
\end{equation}
where
\begin{equation}
\begin{aligned}
\mathbf{A}={\mathbf{M}_1^t}{\mathbf{X}^{t}}^{\top}{\mathbf{H}^{t}}+{\mathbf{M}_2^t}{\mathbf{Z}^{t-1}}^{\top}{\mathbf{W}^t}^{\top}.
\end{aligned}
\end{equation}

Supposing the SVD result of $\mathbf{A}=\mathbf{S} \boldsymbol{\Sigma} \mathbf{V}^{\mathrm{T}}$, according to \cite{ijcai2019-524}, the optimization problem can be solved as follows,
\begin{equation}\label{SVD_Z}
\begin{aligned}
\mathbf{Z}^{t}=\mathbf{V}\mathbf{S}  ^{\top}.
\end{aligned}
\end{equation}

The computational complexity is $\mathcal{O}\left(n_{t}^a k^{2}\right)$.

{\bf Solving $\mathbf{W}^t$ with fixed $\mathbf{Z}^{t}$ and $\mathbf{H}^{t}$}. By dropping the irrelevant variables involved in Eq. \eqref{FCMVC-IV}, the objective formulation concerning $\mathbf{W}^t$ can be rewritten as,
\begin{equation}\label{opt_W}
\begin{aligned}
\max_{\mathbf{W}^t} \operatorname{Tr}\left({\mathbf{W}^t}^{\top} \mathbf{B}\right),
&\text { s.t. } {\mathbf{W}^t}^{\top} {\mathbf{W}^t} = \mathbf{I}_{k},
\end{aligned}
\end{equation}
where
\begin{equation}
\begin{aligned}
\mathbf{B}={\mathbf{Z}^{t}}{\mathbf{M}_2^t}{\mathbf{Z}^{t-1}}^{\top}.
\end{aligned}
\end{equation}

Same as Eq. \eqref{opt_Z}, it can be efficiently solved via SVD with computational complexity $\mathcal{O}\left(k^{3}\right)$.

{\bf Solving $\mathbf{H}^{t}$ with fixed $\mathbf{Z}^{t}$ and $\mathbf{W}^t$}. Given $\mathbf{Z}^{t}$ and $\mathbf{W}^t$, the formulation respecting $\mathbf{H}^{t}$ is reduced to:
\begin{equation}\label{opt_H}
\begin{aligned}
\max_{\mathbf{H}^{t}} \operatorname{Tr}\left({\mathbf{H}^{t}}^{\top} \mathbf{C}\right),
&\text { s.t. } {\mathbf{H}^{t}}^{\top} {\mathbf{H}^{t}} = \mathbf{I}_{k},
\end{aligned}
\end{equation}
where
\begin{equation}
\begin{aligned}
\mathbf{C}={\mathbf{X}^{t}}{\mathbf{M}_1^t}^{\top}{\mathbf{Z}^{t}}^{\top}.
\end{aligned}
\end{equation}

Similar to Eq. (\ref{opt_Z}), Eq. (\ref{opt_H}) can be efficiently solved via SVD with computational complexity $\mathcal{O}\left(d_t k^{2}\right)$.

\begin{algorithm}
	\renewcommand{\algorithmicrequire}{\textbf{Input:}}
	\renewcommand{\algorithmicensure}{\textbf{Output:}}
	\caption{Fast Continual Multi-View Clustering with Incomplete Views}
	\label{alg:1}
	\begin{algorithmic}[1]
		\REQUIRE $\left\{\mathbf{X}^t\right\}_{t=1}^{m}$, $k$, and  $\varepsilon_0$.
		\ENSURE $\mathbf{Z}^m$.
		\STATE Initialize the consensus coefficient matrix $\mathbf{Z}^{1}$ via matrix factorization on $\mathbf{X}^{1}$. 
		\FOR{$t=2$ to $m$}
		\STATE Initialize $\mathbf{H}^t$, $\mathbf{W}^t$, $\mathbf{M}_1^t$, $\mathbf{M}_2^t$, and $i=1$.
		\WHILE {not converged}
		\STATE Update $\mathbf{Z}^t$ by solving Eq. (\ref{opt_Z}).
		\STATE Update $\mathbf{W}^t$ by solving Eq. (\ref{opt_W}).
		\STATE Update $\mathbf{H}^t$ by solving Eq. (\ref{opt_H}).
		\STATE $i\gets i+1$
		\ENDWHILE{ $\left(obj^{i-1}-obj^{i}\right)/obj^i\le\varepsilon_0$}
		\ENDFOR
	\end{algorithmic}  
	\label{algo_whole}
\end{algorithm}
The optimization process of the entire algorithm is summarized in Algorithm 1. It is worth noting that the views arrive in sequence in practical applications, and FCMVC-IV can efficiently handle them. The final clustering results will be obtained by conducting k-means on $\mathbf{Z}^m$, where $m$ is the number of total views.

\subsection{Convergence}
By Cauchy-Schwartz inequality, we have:

\begin{equation}\label{Cauchy-Schwartz_1}
\begin{aligned}
& \operatorname{Tr}\left(\left({\mathbf{Z}^{t} \mathbf{M}_2^t}\right)^{\top} \mathbf{W}^t \mathbf{Z}^{t-1}\right)\\
\leq& \|\mathbf{Z}^{t} \mathbf{M}_2^t \|_F \|\mathbf{W}^t \mathbf{Z}^{t-1}\|_F \\
\leq & \|\mathbf{Z}^{t}\|_F\| \mathbf{M}_2^t \|_F \|\mathbf{W}^t\|_F \|\mathbf{Z}^{t-1}\|_F\\
=&\sqrt{n_{t-1}^a} k^{3/2}.
\end{aligned}
\end{equation}

Then,
\begin{equation}\label{lower_bound}
\begin{aligned}
&\frac{1}{2} \left\|\mathbf{X}^{t}-\mathbf{H}^{t} \mathbf{Z}^{t}\mathbf{M}_1^t\right\|_{F}^{2} -\operatorname{Tr}\left(\left({\mathbf{Z}^{t} \mathbf{M}_2^t}\right)^{\top} \mathbf{W}^t \mathbf{Z}^{t-1}\right)\\
\geq &-\sqrt{n_{t-1}^a} k^{3/2}.
\end{aligned}
\end{equation}

Therefore, the objective function exists a lower bound. Then we will verify that the objective value of Eq. (\ref{FCMVC-IV}) monotonically decreases. 
For ease of expression, we simplify the objective in Eq. (\ref{FCMVC-IV}) as,
\begin{equation}
\begin{aligned}
\min_{\mathbf{Z}^t, \mathbf{W}^t, \mathbf{H}^t} \mathcal{J}\left(\mathbf{Z}^t, \mathbf{W}^t, \mathbf{H}^t\right).
\end{aligned}
\end{equation}

As demonstrated in Algorithm 1, the optimization process consists of three iterative parts in each iteration, i.e., $\mathbf{Z}^t$, $\mathbf{W}^t$, and $\mathbf{H}^t$ subproblems. It is worth mentioning that superscript $i$ denotes the optimization process at round $i$.

\subsubsection{\texorpdfstring{${\mathbf{Z}^t}$}- Subproblem} 
Given ${\mathbf{W}^t}^{\left(i\right)}$ and ${\mathbf{H}^t}^{\left(i\right)}$, ${\mathbf{Z}^t}^{\left(i+1\right)}$ can be obtained via optimizing Eq. (\ref{opt_Z}), leading to:
\begin{equation}\label{Z_ne}
\begin{aligned}
 \mathcal{J}\left({\mathbf{Z}^t}^{\left(i+1\right)}, {\mathbf{W}^t}^{\left(i\right)}, {\mathbf{H}^t}^{\left(i\right)}\right) \leq\\
 \mathcal{J}\left({\mathbf{Z}^t}^{\left(i\right)}, {\mathbf{W}^t}^{\left(i\right)}, {\mathbf{H}^t}^{\left(i\right)}\right).
\end{aligned}
\end{equation}

\subsubsection{\texorpdfstring{${\mathbf{W}^t}$}- Subproblem} Given ${\mathbf{Z}^t}^{\left(i+1\right)}$ and ${\mathbf{H}^t}^{\left(i\right)}$, $\mathbf{W}_{t}^{\left(i+1\right)}$ can be obtained via optimizing Eq. (\ref{opt_W}), leading to:
\begin{equation}\label{W_ne}
\begin{aligned}
 \mathcal{J}\left({\mathbf{Z}^t}^{\left(i+1\right)}, {\mathbf{W}^t}^{\left(i+1\right)}, {\mathbf{H}^t}^{\left(i\right)}\right) \leq\\
 \mathcal{J}\left({\mathbf{Z}^t}^{\left(i+1\right)}, {\mathbf{W}^t}^{\left(i\right)}, {\mathbf{H}^t}^{\left(i\right)}\right).
\end{aligned}
\end{equation}

\subsubsection{\texorpdfstring{${\mathbf{H}^t}$}- Subproblem} Given ${\mathbf{Z}^t}^{\left(i+1\right)}$ and ${\mathbf{W}^t}^{\left(i+1\right)}$, $\mathbf{H}_{t}^{\left(i+1\right)}$ can be obtained via optimizing Eq. (\ref{opt_H}), leading to:
\begin{equation}\label{H_ne}
\begin{aligned}
 \mathcal{J}\left({\mathbf{Z}^t}^{\left(i+1\right)}, {\mathbf{W}^t}^{\left(i+1\right)}, {\mathbf{H}^t}^{\left(i+1\right)}\right)\leq\\
 \mathcal{J}\left({\mathbf{Z}^t}^{\left(i+1\right)}, {\mathbf{W}^t}^{\left(i+1\right)}, {\mathbf{H}^t}^{\left(i\right)}\right).
\end{aligned}
\end{equation}

Together with Eq. (\ref{Z_ne}) to Eq. (\ref{H_ne}), we have:
\begin{equation}\label{final_ne}
\begin{aligned}
 \mathcal{J}\left({\mathbf{Z}^t}^{\left(i+1\right)}, {\mathbf{W}^t}^{\left(i+1\right)}, {\mathbf{H}^t}^{\left(i+1\right)}\right)\leq\\
 \mathcal{J}\left({\mathbf{Z}^t}^{\left(i\right)}, {\mathbf{W}^t}^{\left(i\right)}, {\mathbf{H}^t}^{\left(i\right)}\right),
\end{aligned}
\end{equation}
which demonstrates that the objective value monotonically decreases with iterations.

Based on Eq. (\ref{lower_bound}), we can conclude that Eq. (\ref{FCMVC-IV}) exists a lower bound. Therefore, the algorithm is theoretically convergent.

\subsection{Discussion}
Firstly, we analyze the time and space complexity of our proposed method. Then the initialization of FCMVC-IV is provided.

\subsubsection{Time Complexity}
According to the optimization process outlined in Algorithm 1, three subproblems comprise the optimization procedure. The time complexities are $\mathcal{O}\left(n_{t}^a k^{2}\right)$, $\mathcal{O}\left(k^{3}\right)$ and $\mathcal{O}\left(d_t k^{2}\right)$, separately. Therefore, the total time cost of the optimization process is $\mathcal{O}\left(n_{t}^a k^{2}+k^{3}+d_t k^{2}\right)$. Given that $n_{t}^a\leq n$ and the value of $n_{t}^a$ changes with $t$, we simplify the time complexity as  $\mathcal{O}\left(n k^{2}+k^{3}+d_t k^{2}\right)$. Let $T$ denote the maximum number of iterations and $m$ represent the number of views, the computational complexity of FCMVC-IV is $\mathcal{O}\left(T\left(mn k^{2}+mk^{3}+d k^{2}\right)\right)$ where $d=\sum_{t=1}^{m} d_t$, the time complexity is linear to the sample number. Compared with traditional incomplete MVC methods, most of them undergo a square or cubic time, which is unsuitable for large-scale data. In the meanwhile, they need to fuse all the views when a new view is available, so the computation complexity of traditional ones is $\mathcal{O}\left(m^{2}\right)$ concerning $m$, which is $\mathcal{O}\left(m\right)$ in FCMVC-IV.

\subsubsection{Space Complexity}
During the optimization process, the major storage resources are occupied by three matrices, i.e., $\mathbf{Z}^{t}\in \mathbf{R}^{k\times n_{t}^a}$, $\mathbf{H}^{t}\in \mathbf{R}^{d_t\times k}$ and $\mathbf{W}^{t}\in \mathbf{R}^{k\times k}$. Thus the space complexity of our proposed method is $\mathcal{O}\left(n_{t}^a k+d_t k+k^2\right)$, which is independent of the view number.

\subsubsection{Initialization of \texorpdfstring{$\mathbf{H}^t$}. and  \texorpdfstring{$\mathbf{W}^t$}.}
In our implementation, $\mathbf{H}^t$ is generated by setting its first k rows to the identity matrix and other elements as zero. Meanwhile, we initialize $\mathbf{W}^t$ as an identity matrix.

\section{Experiments}
Since our proposed method can handle both complete and incomplete views, in this section, we conduct experiments to compare FCMVC-IV with several complete/incomplete MVC methods on diverse representative datasets, respectively. After that, its running time and the sensitivity to view order are discussed.

\subsection{Experimental Settings}
  \begin{table}
 \begin{center}
 \caption{Datasets used in our experiments.} 
 \label{dataset}
\begin{tabular}{ cccc}
\toprule
 Dataset & Samples & Views & Clusters\\  \hline 
ORL & 400 & 3 & 40\\ 
proteinFold & 694 & 12 & 27\\ 
uci-digit & 2000 & 3 & 10\\ 
Wiki & 2866 & 2 & 10\\ 
Reuters & 18758 & 5 & 5\\ 
Caltech256 & 30607 & 4 & 257\\ 
VGGFace2& 36287 & 4 & 100\\ 
YouTubeFace10& 38654 & 4 & 10\\ \bottomrule


\end{tabular}
\end{center}
\end{table}

\begin{table*}[]
\centering
	\caption{Empirical evaluation and comparison of FCMVC-IV with nine baseline methods on eight complete benchmark datasets in terms of clustering accuracy (ACC), normalized mutual information (NMI), Purity, and F-score.}
	\label{comlete_result}
	
\begin{tabular}{ccccccccccc}
\toprule

\multicolumn{1}{c}{{\color[HTML]{000000} }} & \multicolumn{1}{c}{\begin{tabular}[c]{@{}c@{}}FMR\\ \cite{ijcai2019p404}\end{tabular}} & \multicolumn{1}{c}{\begin{tabular}[c]{@{}c@{}}PMSC\\ \cite{Kang2019PartitionLM}\end{tabular}} & \multicolumn{1}{c}{\begin{tabular}[c]{@{}c@{}}AMGL\\ \cite{10.5555/3060832.3060884}\end{tabular}} & \multicolumn{1}{c}{\begin{tabular}[c]{@{}c@{}}MNMF\\ \cite{liu2013multi}\end{tabular}} & \multicolumn{1}{c}{\begin{tabular}[c]{@{}c@{}}LMVSC\\ \cite{DBLP:conf/aaai/KangZZSHX20}\end{tabular}} & \multicolumn{1}{c}{\begin{tabular}[c]{@{}c@{}}OPMC\\ \cite{9710821}\end{tabular}} & \multicolumn{1}{c}{\begin{tabular}[c]{@{}c@{}}FMCNOF\\ \cite{9305974}\end{tabular}} & \multicolumn{1}{c}{\begin{tabular}[c]{@{}c@{}}FPMVS\\ \cite{9646486}\end{tabular}}
 & \multicolumn{1}{c}{\begin{tabular}[c]{@{}c@{}}CMVC\\ \cite{10.1145/3503161.3547864}\end{tabular}} & \begin{tabular}[c]{@{}c@{}}FCMVC\\ proposed\end{tabular}\\ \hline
\multicolumn{1}{c}{Param.   num.}           &  \multicolumn{1}{c}{2}           & \multicolumn{1}{c}{3}           & \multicolumn{1}{c}{0}              & \multicolumn{1}{c}{1}           & \multicolumn{1}{c}{1}              & \multicolumn{1}{c}{0}              & \multicolumn{1}{c}{1}      & \multicolumn{1}{c}{0}           & \multicolumn{1}{c}{1}     &0              \\ \hline
\multicolumn{11}{c}{ACC}                                                                                                                                                                                                                                                                                                                                                              \\ \hline
\multicolumn{1}{c}{ORL}                     & \multicolumn{1}{c}{65.79}       & \multicolumn{1}{c}{63.47}       & \multicolumn{1}{c}{71.15}          & \multicolumn{1}{c}{{\ul 73.25}} & \multicolumn{1}{c}{65.65}          & \multicolumn{1}{c}{60.25}          & \multicolumn{1}{c}{27.50}  & \multicolumn{1}{c}{66.75}       & \multicolumn{1}{c}{72.27} & \textbf{75.75} \\ 
\multicolumn{1}{c}{proteinFold}             &  \multicolumn{1}{c}{ 32.23} & \multicolumn{1}{c}{20.65}       & \multicolumn{1}{c}{10.74}          & \multicolumn{1}{c}{11.53}       & \multicolumn{1}{c}{28.77}          & \multicolumn{1}{c}{27.23}          & \multicolumn{1}{c}{21.33}  & \multicolumn{1}{c}{30.03}       & \multicolumn{1}{c}{{\ul 33.74}} &\textbf{34.87} \\ 
\multicolumn{1}{c}{uci-digit}               &  \multicolumn{1}{c}{72.47}       & \multicolumn{1}{c}{70.92}       & \multicolumn{1}{c}{77.76}          & \multicolumn{1}{c}{10.05}       & \multicolumn{1}{c}{{\ul 83.00}}    & \multicolumn{1}{c}{78.35}          & \multicolumn{1}{c}{54.85}  & \multicolumn{1}{c}{81.89}       & \multicolumn{1}{c}{82.29}     &\textbf{85.30} \\ 
\multicolumn{1}{c}{Wiki}                    & \multicolumn{1}{c}{{\ul 56.94}} & \multicolumn{1}{c}{56.76}       & \multicolumn{1}{c}{12.23}          & \multicolumn{1}{c}{50.35}       & \multicolumn{1}{c}{56.79}          & \multicolumn{1}{c}{19.50}          & \multicolumn{1}{c}{29.62}  & \multicolumn{1}{c}{51.18}       & \multicolumn{1}{c}{56.35}     & \textbf{57.75} \\ 
\multicolumn{1}{c}{Reuters}                 & \multicolumn{1}{c}{-}           & \multicolumn{1}{c}{-}           & \multicolumn{1}{c}{16.72}          & \multicolumn{1}{c}{27.20}       & \multicolumn{1}{c}{49.05}          & \multicolumn{1}{c}{ 55.11}    & \multicolumn{1}{c}{36.83}  & \multicolumn{1}{c}{44.58}       & \multicolumn{1}{c}{{\ul 55.39}}     & \textbf{57.32} \\ 
\multicolumn{1}{c}{Caltech256}              &  \multicolumn{1}{c}{-}           & \multicolumn{1}{c}{-}           & \multicolumn{1}{c}{-}              & \multicolumn{1}{c}{2.71}        & \multicolumn{1}{c}{9.57}           & \multicolumn{1}{c}{{\ul 10.13}}    & \multicolumn{1}{c}{2.70}   & \multicolumn{1}{c}{8.78}        & \multicolumn{1}{c}{10.07}     &\textbf{10.24} \\ 
\multicolumn{1}{c}{VGGFace2}                &  \multicolumn{1}{c}{-}           & \multicolumn{1}{c}{-}           & \multicolumn{1}{c}{-}              & \multicolumn{1}{c}{1.66}        & \multicolumn{1}{c}{6.09}    & \multicolumn{1}{c}{5.57}          & \multicolumn{1}{c}{3.47}   & \multicolumn{1}{c}{6.02}       & \multicolumn{1}{c}{{\ul 6.42}}     &\textbf{7.84} \\ 
\multicolumn{1}{c}{YouTubeFace10}           &  \multicolumn{1}{c}{-}           & \multicolumn{1}{c}{-}           & \multicolumn{1}{c}{-}              & \multicolumn{1}{c}{15.71}       & \multicolumn{1}{c}{74.48}          & \multicolumn{1}{c}{{\ul 76.13}}    & \multicolumn{1}{c}{43.42}  & \multicolumn{1}{c}{67.09}       &\multicolumn{1}{c}{73.27}     & \textbf{93.40} \\ \hline

\multicolumn{11}{c}{NMI}                                                                                                                                                                                                                                                                                                                                                              \\ \hline
\multicolumn{1}{c}{ORL}                     &  \multicolumn{1}{c}{81.20}       & \multicolumn{1}{c}{80.93}       & \multicolumn{1}{c}{\textbf{87.64}} & \multicolumn{1}{c}{86.79}       & \multicolumn{1}{c}{83.35}          & \multicolumn{1}{c}{77.84}          & \multicolumn{1}{c}{49.23}  & \multicolumn{1}{c}{86.26}       &\multicolumn{1}{c}{85.24} & {\ul 87.20}    \\ 
\multicolumn{1}{c}{proteinFold}             &  \multicolumn{1}{c}{ 41.75} & \multicolumn{1}{c}{26.53}       & \multicolumn{1}{c}{19.69}          & \multicolumn{1}{c}{8.40}        & \multicolumn{1}{c}{39.16}          & \multicolumn{1}{c}{29.70}          & \multicolumn{1}{c}{25.86}  & \multicolumn{1}{c}{37.75}       & \multicolumn{1}{c}{{\ul 42.28}} &\textbf{43.48} \\ 
\multicolumn{1}{c}{uci-digit}               &  \multicolumn{1}{c}{70.33}       & \multicolumn{1}{c}{72.40}       & \multicolumn{1}{c}{\textbf{83.88}} & \multicolumn{1}{c}{0.89}        & \multicolumn{1}{c}{ 81.55}    & \multicolumn{1}{c}{78.31}          & \multicolumn{1}{c}{57.17}  & \multicolumn{1}{c}{78.76}       & \multicolumn{1}{c}{{\ul 82.72}}     &75.44          \\ 
\multicolumn{1}{c}{Wiki}                    &  \multicolumn{1}{c}{50.17}       & \multicolumn{1}{c}{{\ul 52.42}} & \multicolumn{1}{c}{0.83}           & \multicolumn{1}{c}{40.27}       & \multicolumn{1}{c}{49.89}          & \multicolumn{1}{c}{6.14}           & \multicolumn{1}{c}{18.96}  & \multicolumn{1}{c}{49.34}       & \multicolumn{1}{c}{51.72}     &\textbf{53.17} \\ 
\multicolumn{1}{c}{Reuters}                 &  \multicolumn{1}{c}{-}           & \multicolumn{1}{c}{-}           & \multicolumn{1}{c}{0.19}           & \multicolumn{1}{c}{0.07}        & \multicolumn{1}{c}{\textbf{32.13}} & \multicolumn{1}{c}{28.80}          & \multicolumn{1}{c}{17.06}  & \multicolumn{1}{c}{28.50}       &\multicolumn{1}{c}{30.01}     & {\ul 30.26}    \\ 
\multicolumn{1}{c}{Caltech256}              &  \multicolumn{1}{c}{-}           & \multicolumn{1}{c}{-}           & \multicolumn{1}{c}{-}              & \multicolumn{1}{c}{1.96}        & \multicolumn{1}{c}{\textbf{31.96}} & \multicolumn{1}{c}{30.70}          & \multicolumn{1}{c}{0.00}   & \multicolumn{1}{c}{22.97}       & \multicolumn{1}{c}{30.76}     &{\ul 31.38}    \\ 
\multicolumn{1}{c}{VGGFace2}                &  \multicolumn{1}{c}{-}           & \multicolumn{1}{c}{-}           & \multicolumn{1}{c}{-}              & \multicolumn{1}{c}{0.58}        & \multicolumn{1}{c}{11.92}    & \multicolumn{1}{c}{11.68}          & \multicolumn{1}{c}{5.81}   & \multicolumn{1}{c}{ 12.33}       &\multicolumn{1}{c}{{\ul 13.52}}     & \textbf{14.84} \\ 
\multicolumn{1}{c}{YouTubeFace10}           &  \multicolumn{1}{c}{-}           & \multicolumn{1}{c}{-}           & \multicolumn{1}{c}{-}              & \multicolumn{1}{c}{0.05}        & \multicolumn{1}{c}{77.74}          & \multicolumn{1}{c}{ 80.03}    & \multicolumn{1}{c}{39.15}  & \multicolumn{1}{c}{76.11}       & \multicolumn{1}{c}{{\ul 82.35}}     &\textbf{88.45} \\ \hline

\multicolumn{11}{c}{Purity}                                                                                                                                                                                                                                                                                                                                                           \\ \hline
\multicolumn{1}{c}{ORL}                     &  \multicolumn{1}{c}{69.10}       & \multicolumn{1}{c}{67.21}       & \multicolumn{1}{c}{{\ul 76.47}}    & \multicolumn{1}{c}{76.00}       & \multicolumn{1}{c}{69.18}          & \multicolumn{1}{c}{64.25}          & \multicolumn{1}{c}{28.25}  & \multicolumn{1}{c}{73.75}       & \multicolumn{1}{c}{73.92} &\textbf{78.50} \\ 
\multicolumn{1}{c}{proteinFold}             &  \multicolumn{1}{c}{38.94} & \multicolumn{1}{c}{24.79}       & \multicolumn{1}{c}{11.50}          & \multicolumn{1}{c}{14.84}       & \multicolumn{1}{c}{36.62}          & \multicolumn{1}{c}{31.70}          & \multicolumn{1}{c}{24.50}  & \multicolumn{1}{c}{34.95}       & \textbf{41.22} &\multicolumn{1}{c}{{\ul 40.63}} \\ 
\multicolumn{1}{c}{uci-digit}               &  \multicolumn{1}{c}{75.30}       & \multicolumn{1}{c}{73.97}       & \multicolumn{1}{c}{80.71}          & \multicolumn{1}{c}{10.45}       & \multicolumn{1}{c}{{\ul 84.47}}    & \multicolumn{1}{c}{83.05}          & \multicolumn{1}{c}{56.25}  & \multicolumn{1}{c}{82.08}       & \multicolumn{1}{c}{83.61}     &\textbf{85.30} \\ 
\multicolumn{1}{c}{Wiki}                    &  \multicolumn{1}{c}{59.47}       & \multicolumn{1}{c}{ 61.31} & \multicolumn{1}{c}{12.51}          & \multicolumn{1}{c}{55.51}       & \multicolumn{1}{c}{60.18}          & \multicolumn{1}{c}{22.47}          & \multicolumn{1}{c}{29.90}  & \multicolumn{1}{c}{55.97}       & \multicolumn{1}{c}{{\ul61.39}}     &\textbf{61.76} \\ 
\multicolumn{1}{c}{Reuters}                 &  \multicolumn{1}{c}{-}           & \multicolumn{1}{c}{-}           & \multicolumn{1}{c}{16.76}          & \multicolumn{1}{c}{27.23}       & \multicolumn{1}{c}{58.00}          & \multicolumn{1}{c}{{\ul 58.74}}    & \multicolumn{1}{c}{40.42}  & \multicolumn{1}{c}{53.58}       & \multicolumn{1}{c}{57.76}     &\textbf{59.24} \\ 
\multicolumn{1}{c}{Caltech256}              &  \multicolumn{1}{c}{-}           & \multicolumn{1}{c}{-}           & \multicolumn{1}{c}{-}              & \multicolumn{1}{c}{3.54}        & \multicolumn{1}{c}{{\ul 16.25}}    & \multicolumn{1}{c}{\textbf{16.57}} & \multicolumn{1}{c}{2.70}   & \multicolumn{1}{c}{11.88}  &     \multicolumn{1}{c}{15.38}     & 15.76          \\ 
\multicolumn{1}{c}{VGGFace2}                &  \multicolumn{1}{c}{-}           & \multicolumn{1}{c}{-}           & \multicolumn{1}{c}{-}              & \multicolumn{1}{c}{1.93}        & \multicolumn{1}{c}{7.02}     & \multicolumn{1}{c}{6.31}           & \multicolumn{1}{c}{3.50}   & \multicolumn{1}{c}{6.33}        & \multicolumn{1}{c}{{\ul 7.19}}     &\textbf{8.68}  \\ 
\multicolumn{1}{c}{YouTubeFace10}           &  \multicolumn{1}{c}{-}           & \multicolumn{1}{c}{-}           & \multicolumn{1}{c}{-}              & \multicolumn{1}{c}{15.73}       & \multicolumn{1}{c}{78.39}          & \multicolumn{1}{c}{{\ul 78.42}}    & \multicolumn{1}{c}{46.53}  & \multicolumn{1}{c}{69.43}       & \multicolumn{1}{c}{76.28}     &\textbf{93.40} \\ \hline

\multicolumn{11}{c}{F-score}                                                                                                                                                                                                                                                                                                                                                           \\ \hline
\multicolumn{1}{c}{ORL}                     &  \multicolumn{1}{c}{54.00}       & \multicolumn{1}{c}{52.57}       & \multicolumn{1}{c}{53.73}          & \multicolumn{1}{c}{{\ul 65.21}} & \multicolumn{1}{c}{56.52}          & \multicolumn{1}{c}{47.74}          & \multicolumn{1}{c}{13.80}  & \multicolumn{1}{c}{58.73}       & \multicolumn{1}{c}{63.22} &\textbf{66.97} \\ 
\multicolumn{1}{c}{proteinFold}             &  \multicolumn{1}{c}{18.70} & \multicolumn{1}{c}{13.42}       & \multicolumn{1}{c}{7.72}           & \multicolumn{1}{c}{10.02}       & \multicolumn{1}{c}{16.94}          & \multicolumn{1}{c}{7.22}           & \multicolumn{1}{c}{13.52}  & \multicolumn{1}{c}{17.09}       & \multicolumn{1}{c}{{{\ul 20.13}}}&\textbf{21.87} \\ 
\multicolumn{1}{c}{uci-digit}               & \multicolumn{1}{c}{64.55}       & \multicolumn{1}{c}{64.17}       & \multicolumn{1}{c}{75.03}          & \multicolumn{1}{c}{18.10}       & \multicolumn{1}{c}{\textbf{77.41}} & \multicolumn{1}{c}{70.53}          & \multicolumn{1}{c}{50.40}  & \multicolumn{1}{c}{{\ul 75.25}} &\multicolumn{1}{c}{74.38}     &  73.40          \\ 
\multicolumn{1}{c}{Wiki}                    &  \multicolumn{1}{c}{{\ul 48.39}} & \multicolumn{1}{c}{47.90}       & \multicolumn{1}{c}{12.52}          & \multicolumn{1}{c}{40.70}       & \multicolumn{1}{c}{47.85}          & \multicolumn{1}{c}{2.99}           & \multicolumn{1}{c}{23.61}  & \multicolumn{1}{c}{44.91}       & \multicolumn{1}{c}{47.09}     &\textbf{51.50} \\ 
\multicolumn{1}{c}{Reuters}                 &  \multicolumn{1}{c}{-}           & \multicolumn{1}{c}{-}           & \multicolumn{1}{c}{28.55}          & \multicolumn{1}{c}{35.26}       & \multicolumn{1}{c}{41.40}    & \multicolumn{1}{c}{27.40}          & \multicolumn{1}{c}{35.21}  & \multicolumn{1}{c}{39.06}       & \multicolumn{1}{c}{{\ul 43.61}}    &\textbf{45.37} \\ 
\multicolumn{1}{c}{Caltech256}              &  \multicolumn{1}{c}{-}           & \multicolumn{1}{c}{-}           & \multicolumn{1}{c}{-}              & \multicolumn{1}{c}{1.18}        & \multicolumn{1}{c}{6.00}           & \multicolumn{1}{c}{{\ul 7.25}}     & \multicolumn{1}{c}{1.17}   & \multicolumn{1}{c}{3.22}        & \multicolumn{1}{c}{7.13}     &\textbf{8.93}  \\ 
\multicolumn{1}{c}{VGGFace2}                &  \multicolumn{1}{c}{-}           & \multicolumn{1}{c}{-}           & \multicolumn{1}{c}{-}              & \multicolumn{1}{c}{2.12}        & \multicolumn{1}{c}{2.47}           & \multicolumn{1}{c}{1.15}           & \multicolumn{1}{c}{2.39}   & \multicolumn{1}{c}{{\ul 3.14}}  &\multicolumn{1}{c}{3.07}     & \textbf{3.14}  \\ 
\multicolumn{1}{c}{YouTubeFace10}           & \multicolumn{1}{c}{-}           & \multicolumn{1}{c}{-}           & \multicolumn{1}{c}{-}              & \multicolumn{1}{c}{19.69}       & \multicolumn{1}{c}{68.93}   & \multicolumn{1}{c}{68.57}          & \multicolumn{1}{c}{32.88}  & \multicolumn{1}{c}{66.10}       & \multicolumn{1}{c}{{\ul72.75}}     & \textbf{89.09} \\ 
\bottomrule
\end{tabular}
\end{table*}

\begin{figure*}[htbp]
	\centering
	\vspace{-0.2cm}
	\subfigure{
		\includegraphics[width=0.21\textwidth]{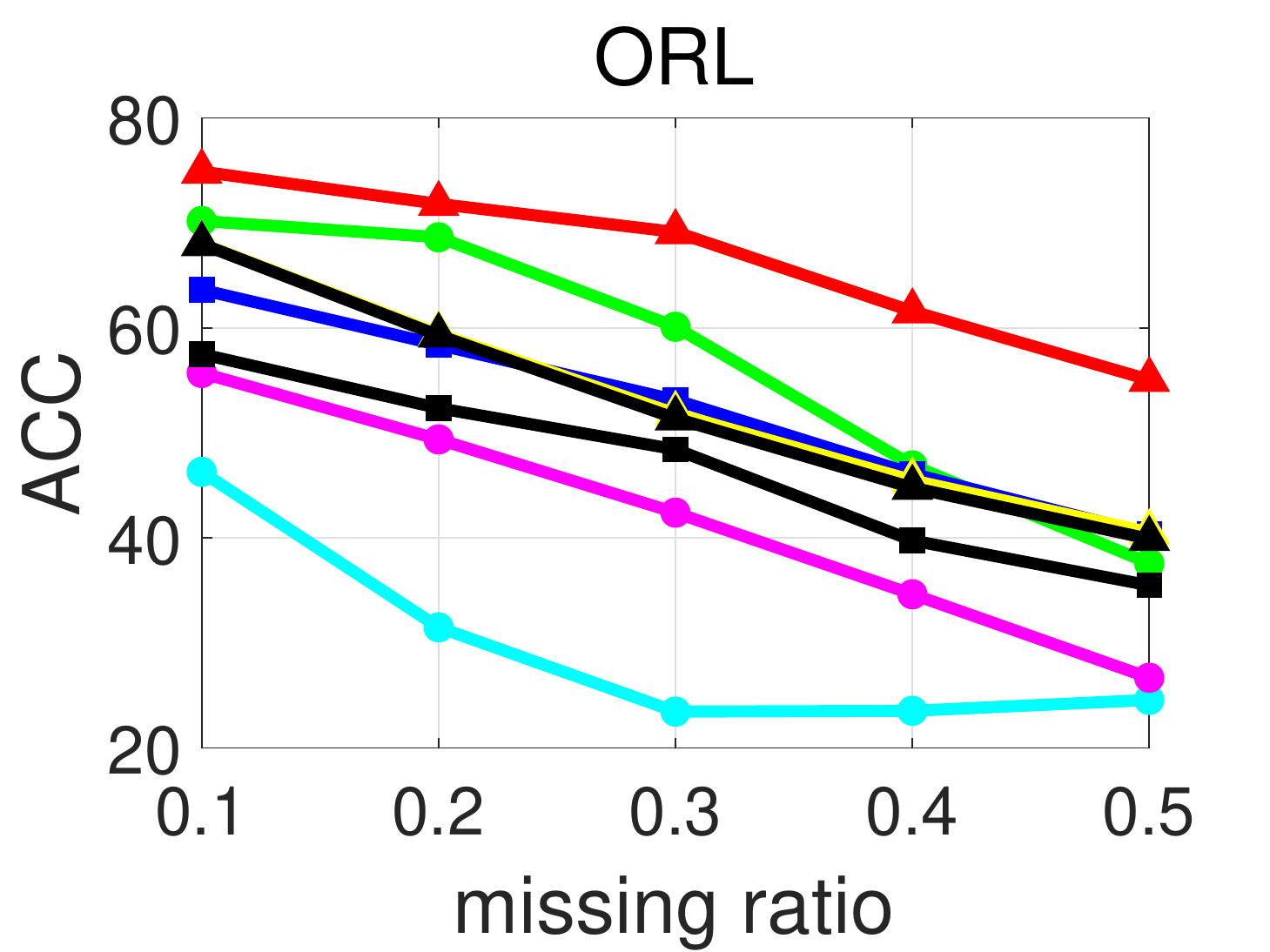}}
	\hspace{-0.2cm}
	\subfigure{
		\includegraphics[width=0.21\textwidth]{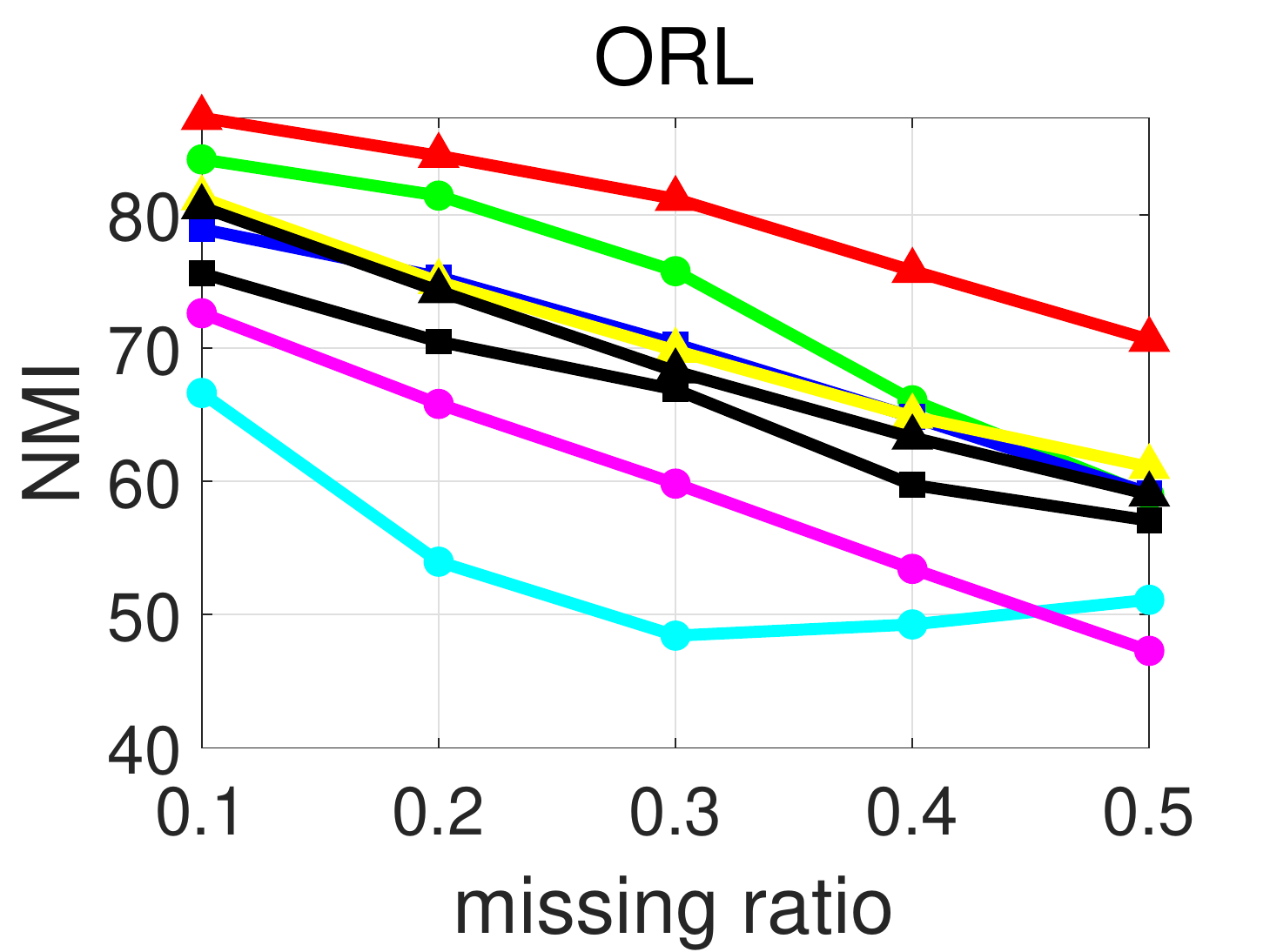}}
	\hspace{-0.2cm}
	\subfigure{
		\includegraphics[width=0.21\textwidth]{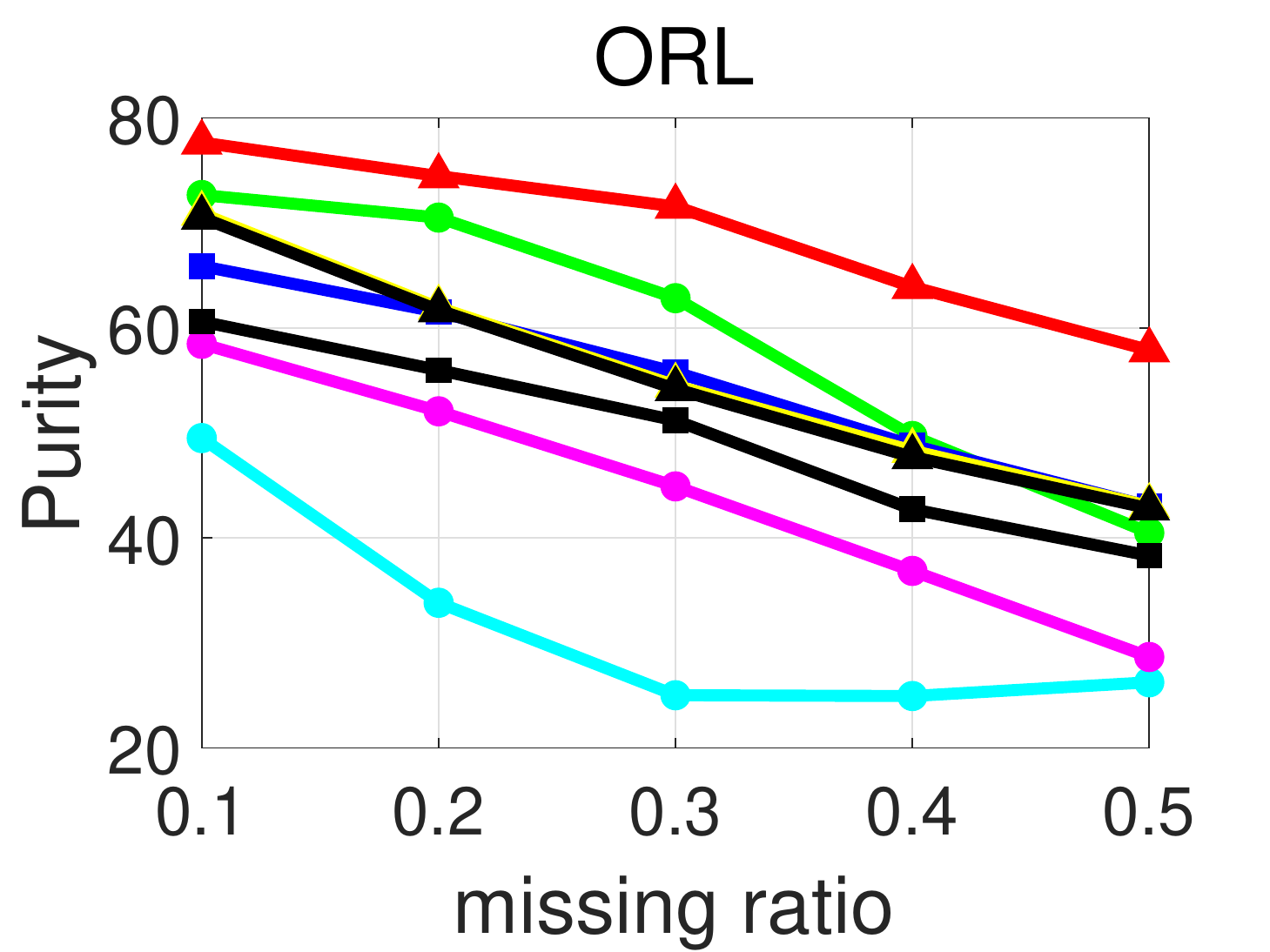}}
	\hspace{-0.2cm}
	\subfigure{
		\includegraphics[width=0.312\textwidth]{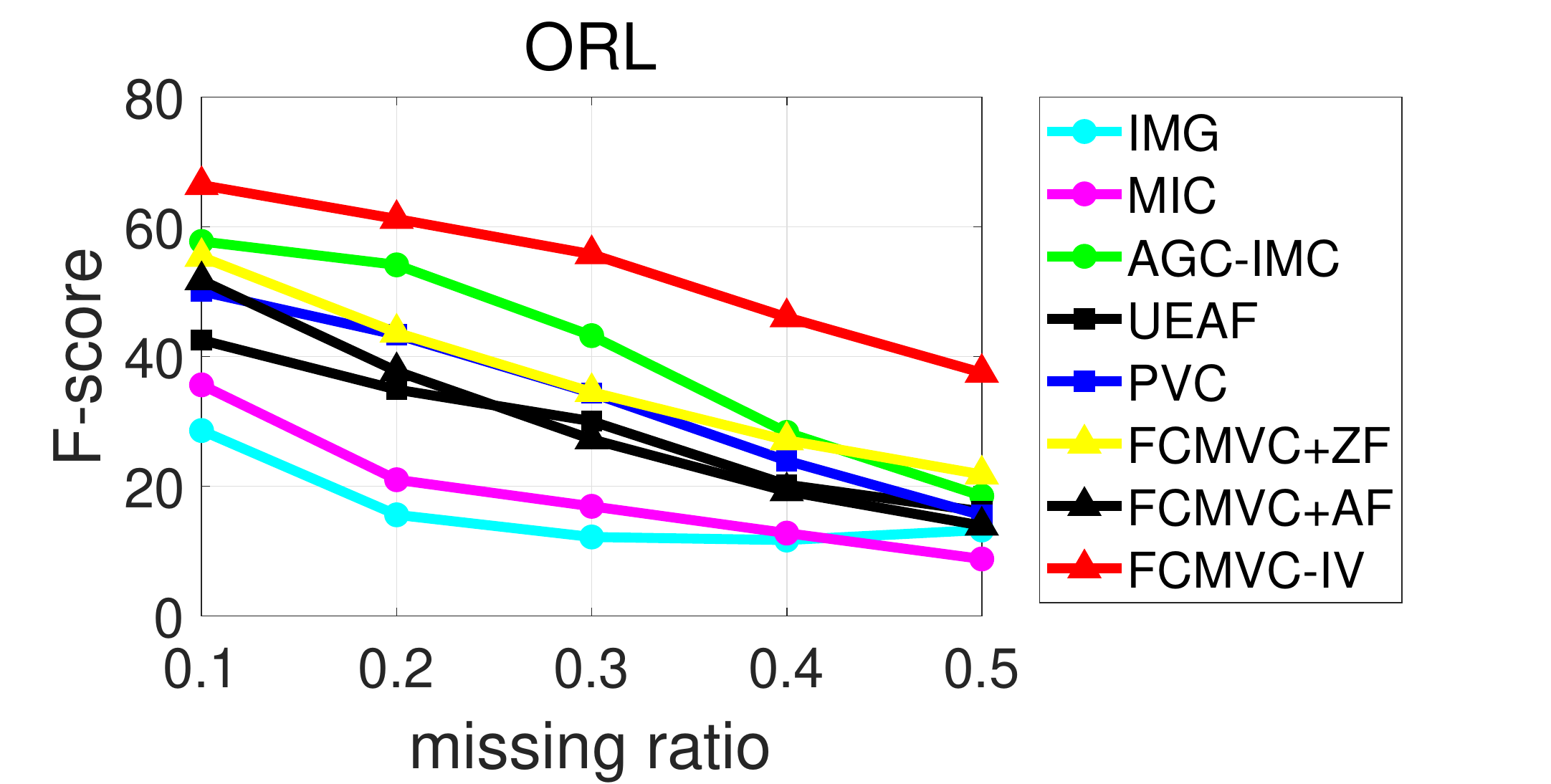}}
	\hspace{-0.2cm}
	\subfigure{
		\includegraphics[width=0.21\textwidth]{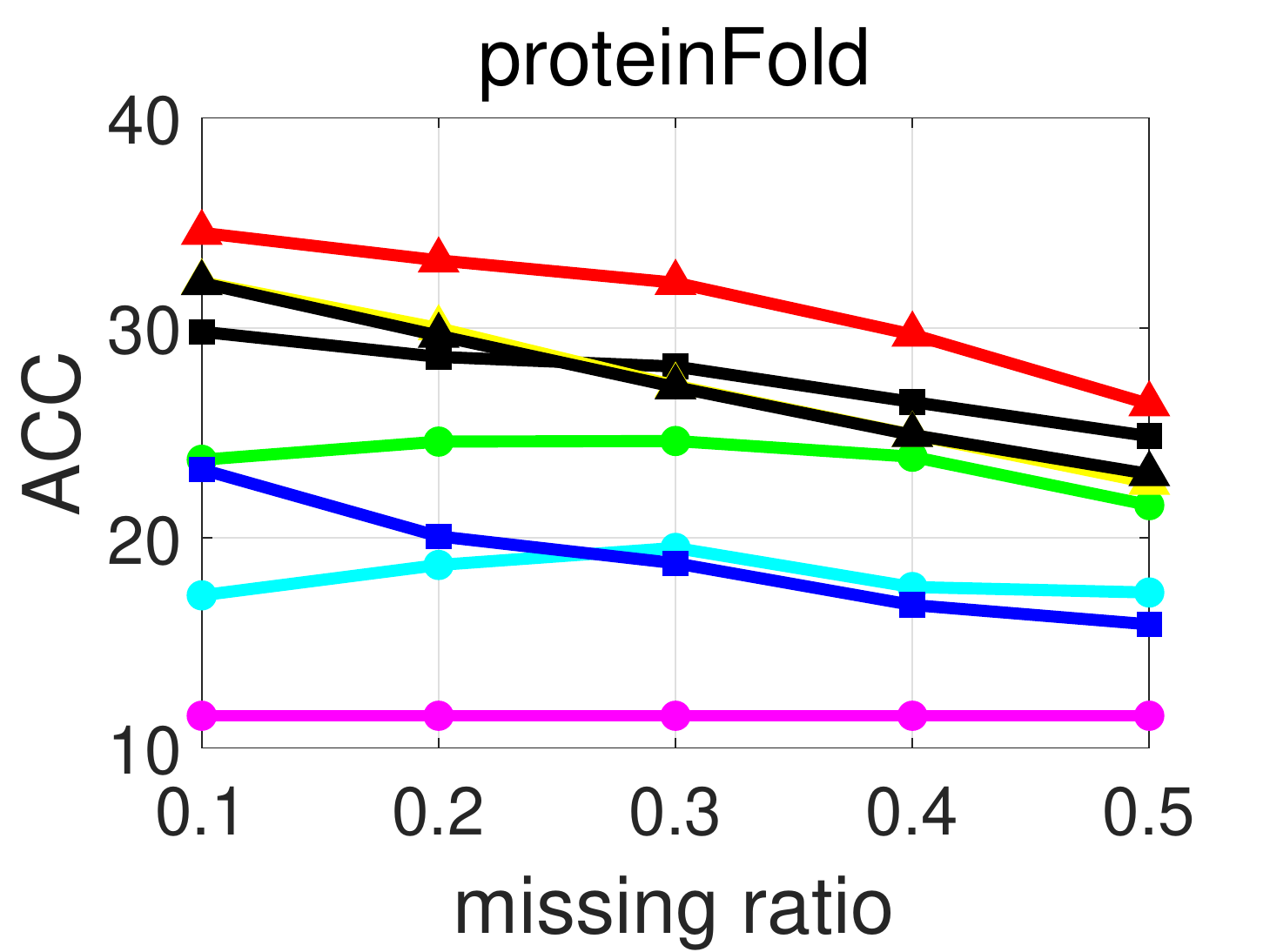}}
	\hspace{-0.2cm}
	\subfigure{
		\includegraphics[width=0.21\textwidth]{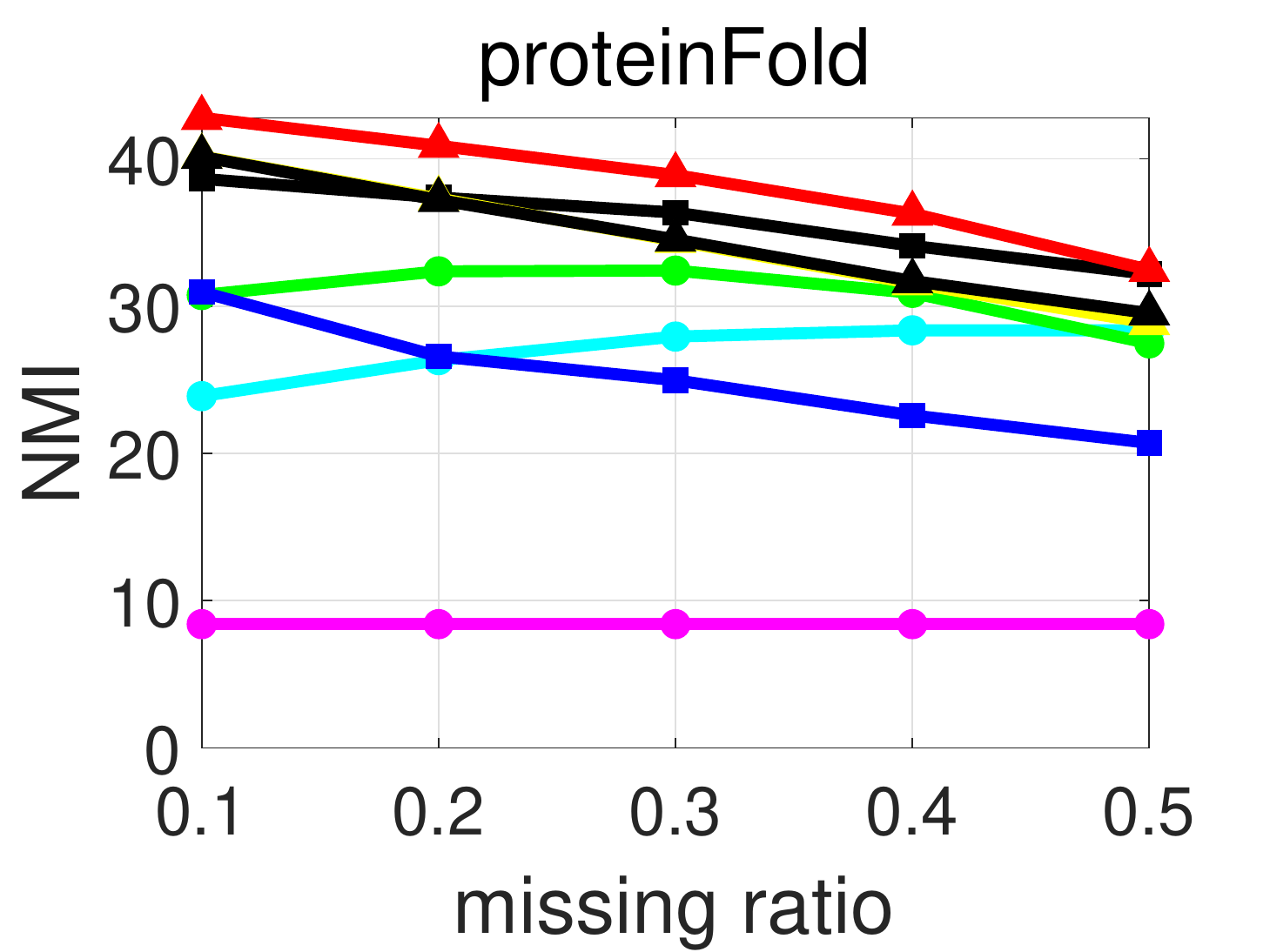}}
	\hspace{-0.2cm}
	\subfigure{
		\includegraphics[width=0.21\textwidth]{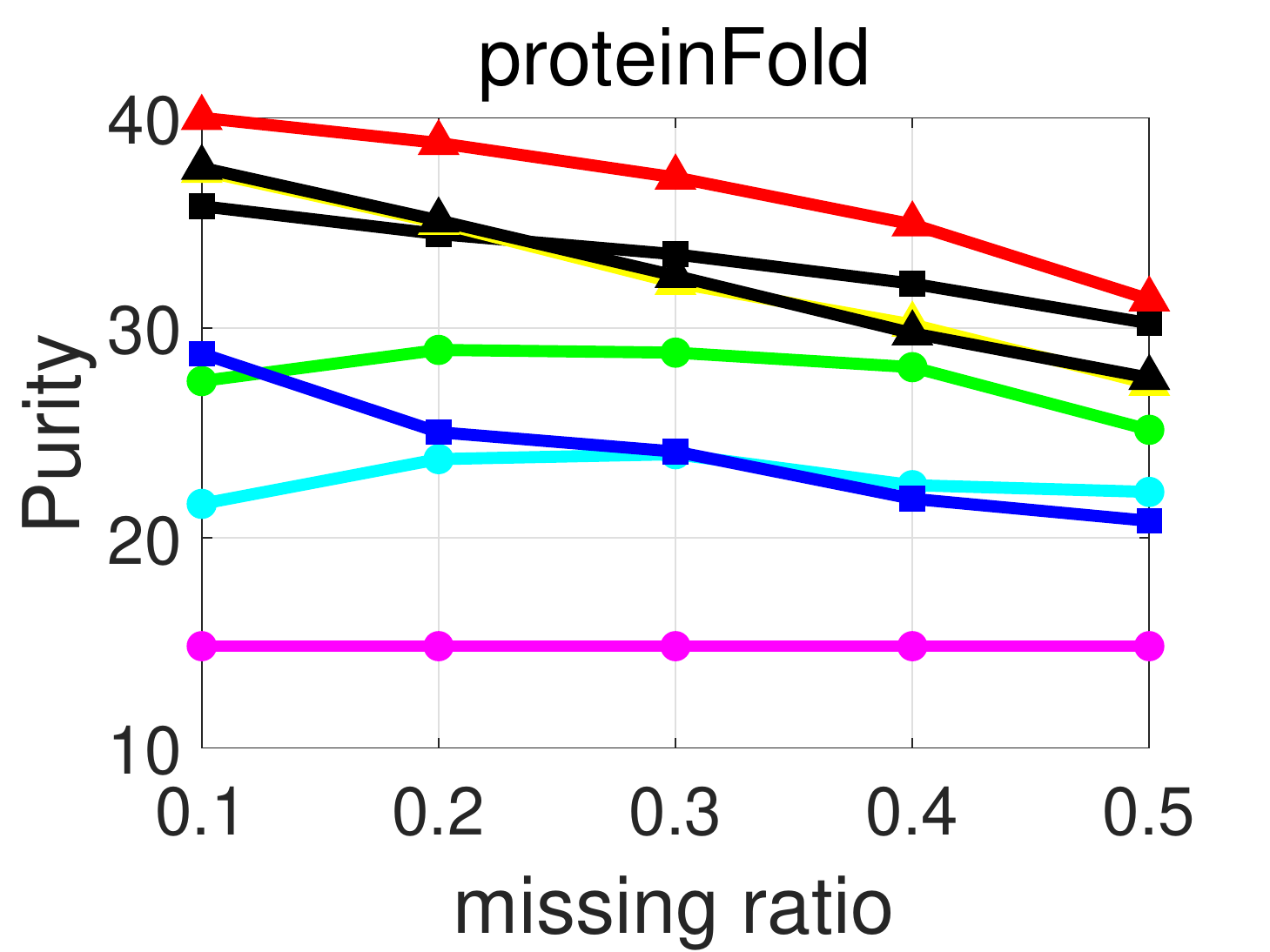}}
	\hspace{-0.2cm}
	\subfigure{
		\includegraphics[width=0.312\textwidth]{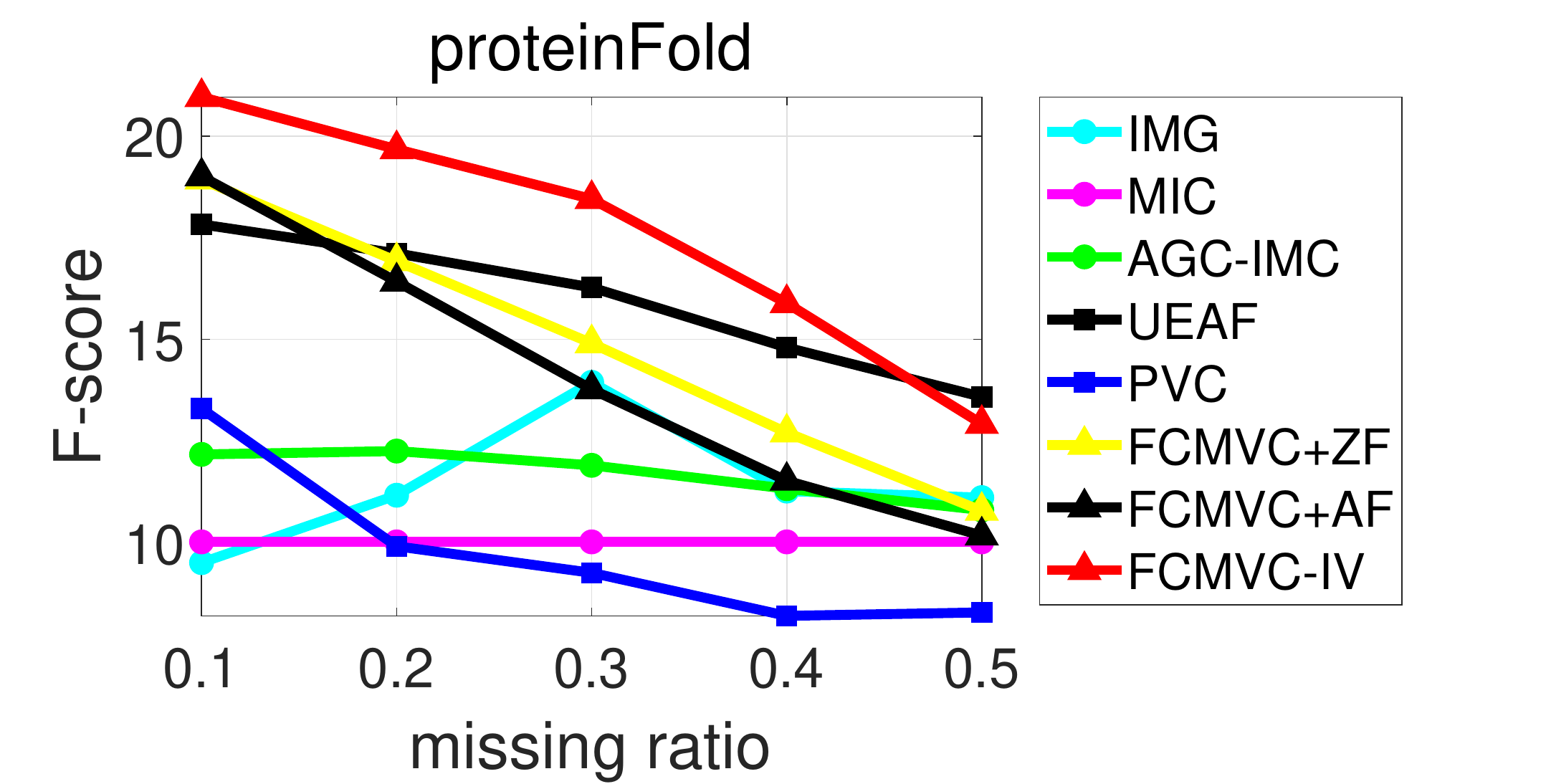}}
	\hspace{-0.2cm}
	\subfigure{
		\includegraphics[width=0.21\textwidth]{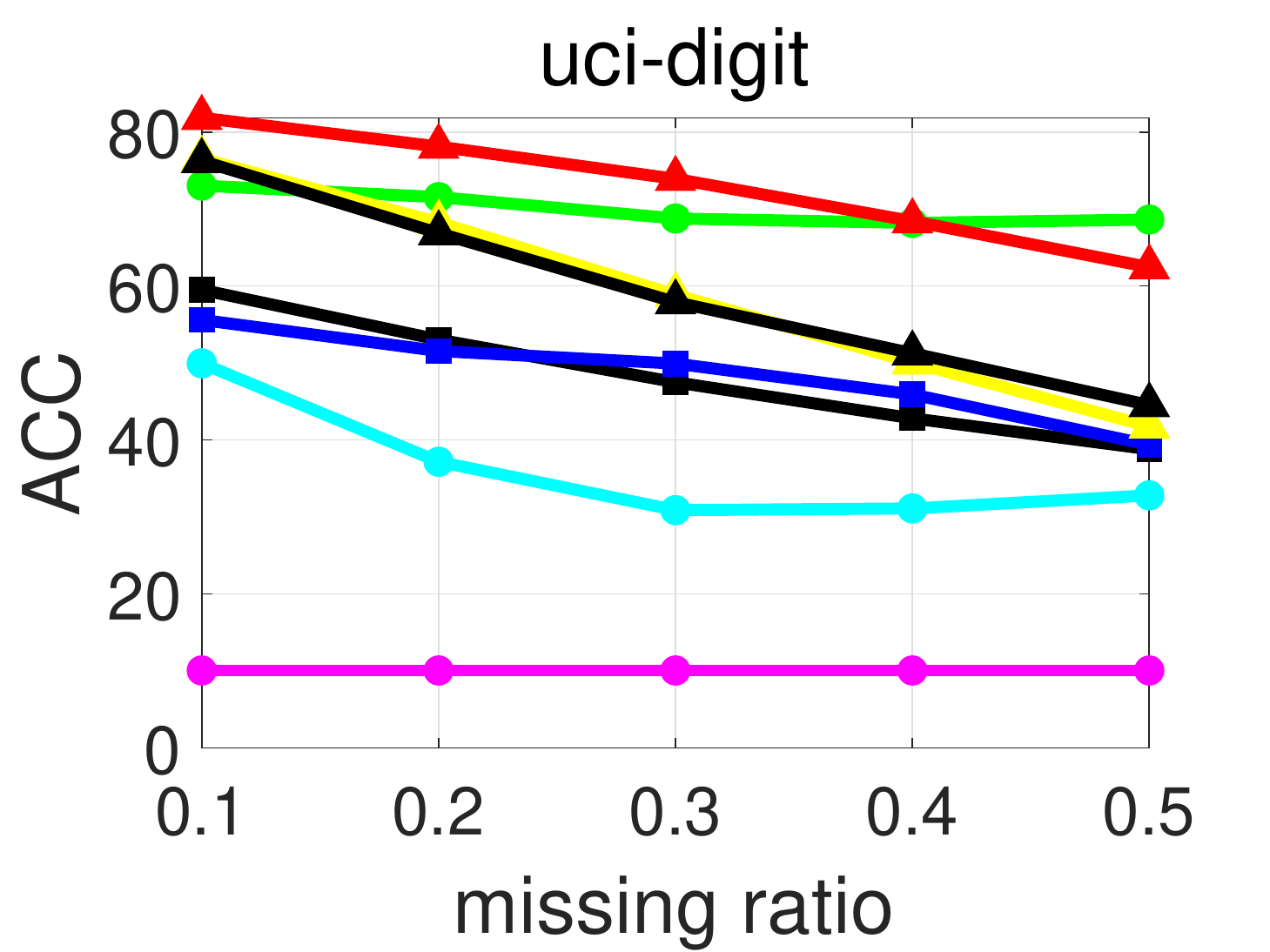}}
	\hspace{-0.2cm}
	\subfigure{
		\includegraphics[width=0.21\textwidth]{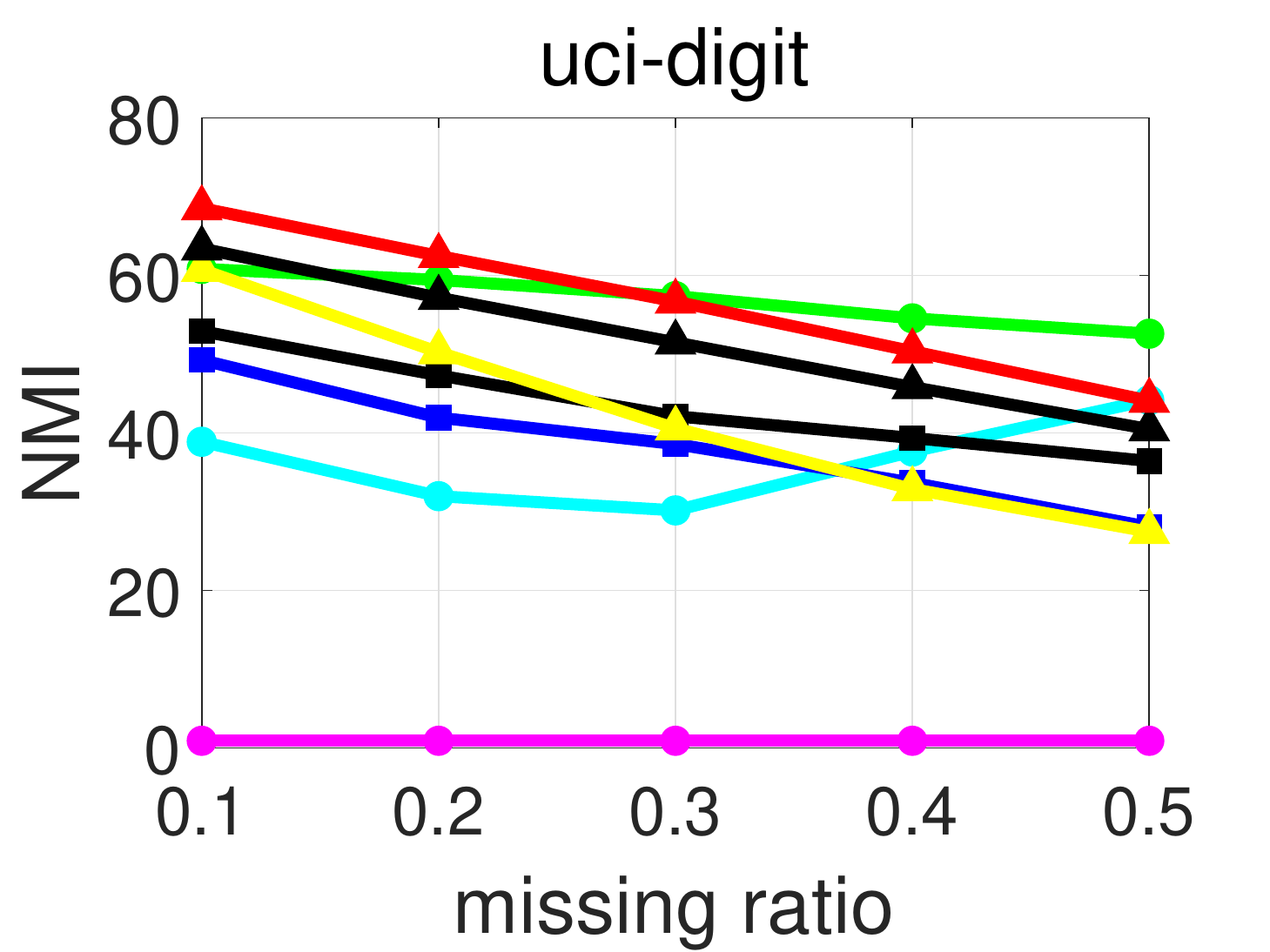}}
	\hspace{-0.2cm}
	\subfigure{
		\includegraphics[width=0.21\textwidth]{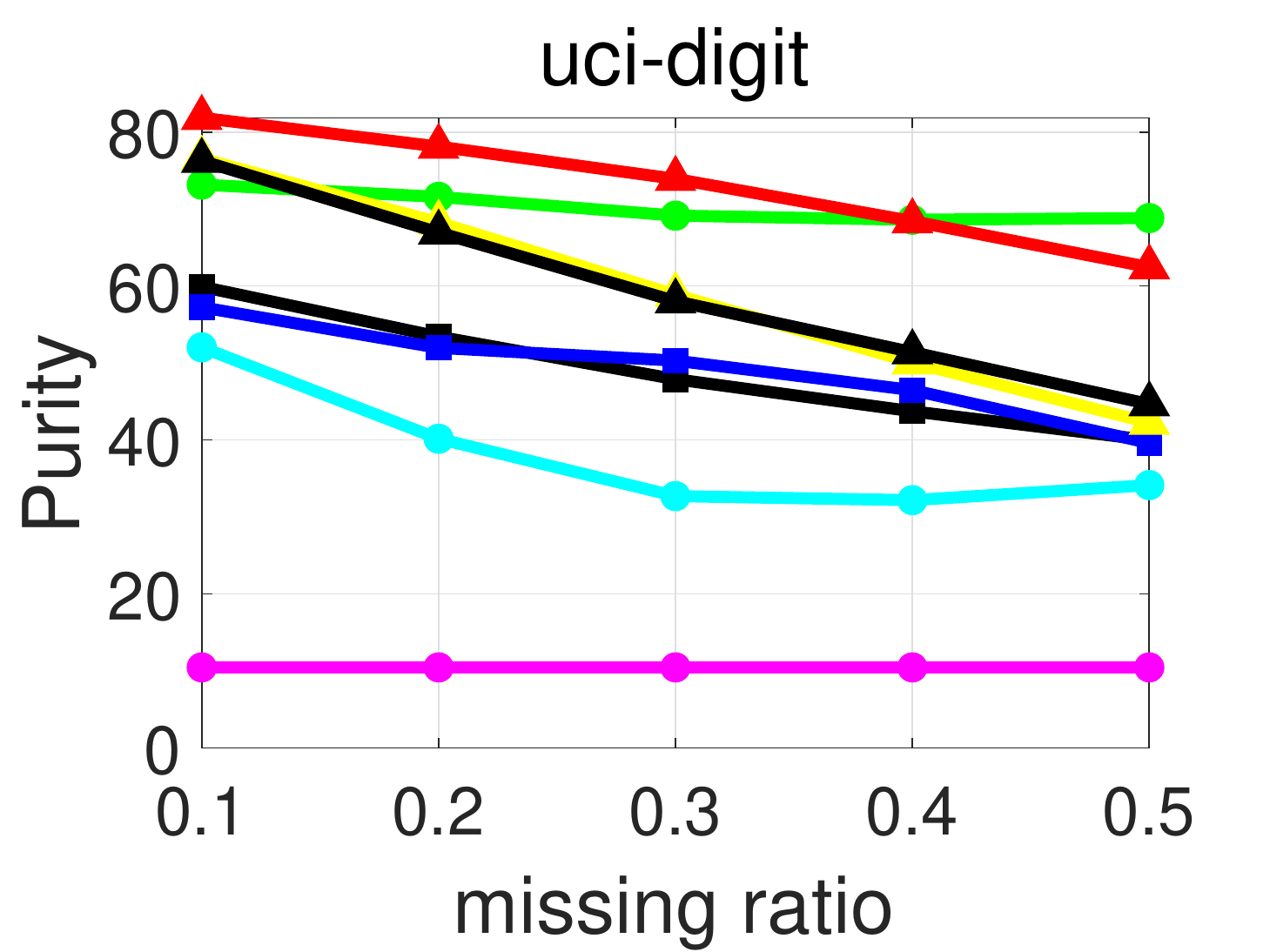}}
	\hspace{-0.2cm}
	\subfigure{
		\includegraphics[width=0.312\textwidth]{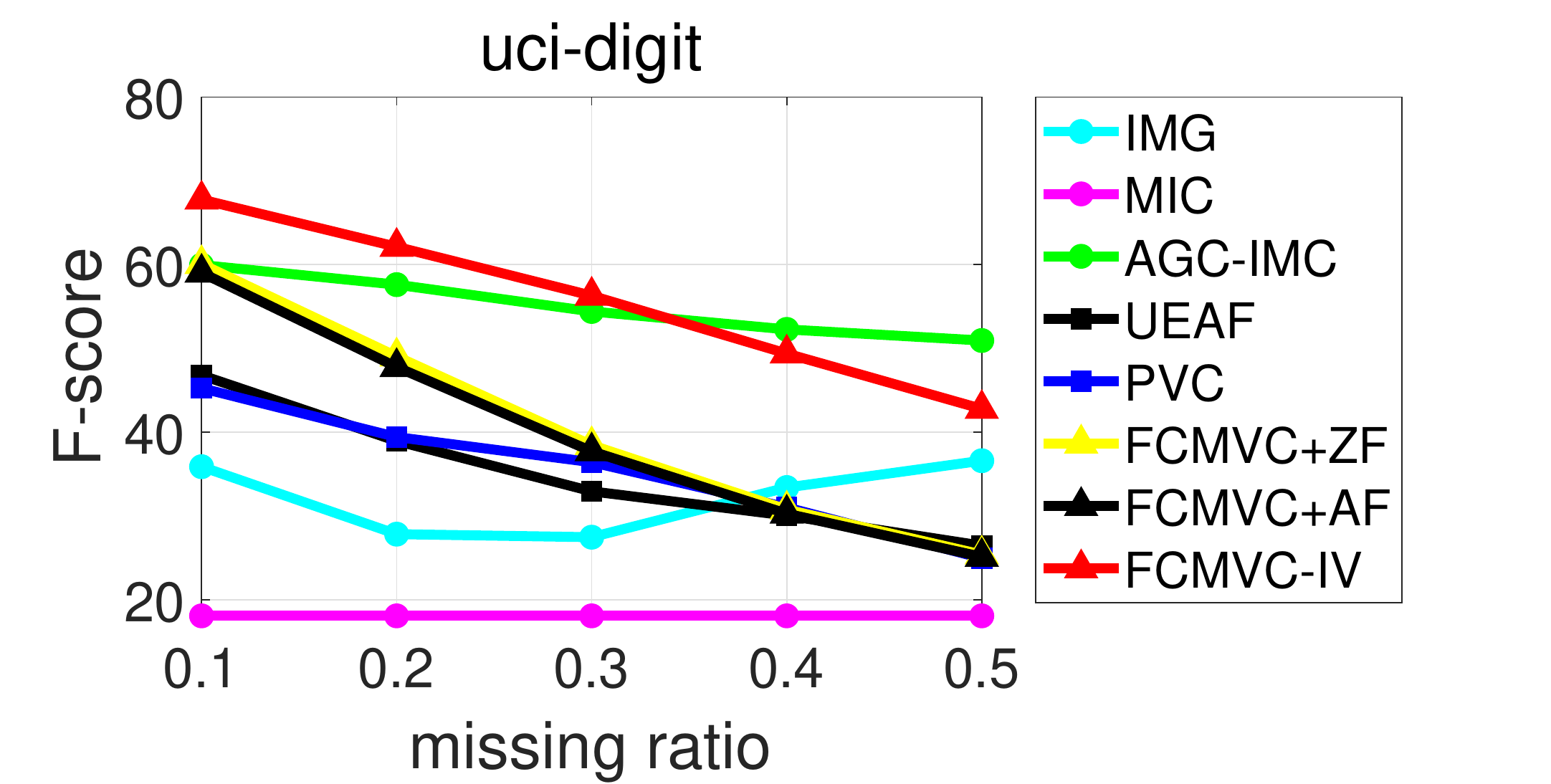}}
	\hspace{-0.2cm}
	\subfigure{
		\includegraphics[width=0.21\textwidth]{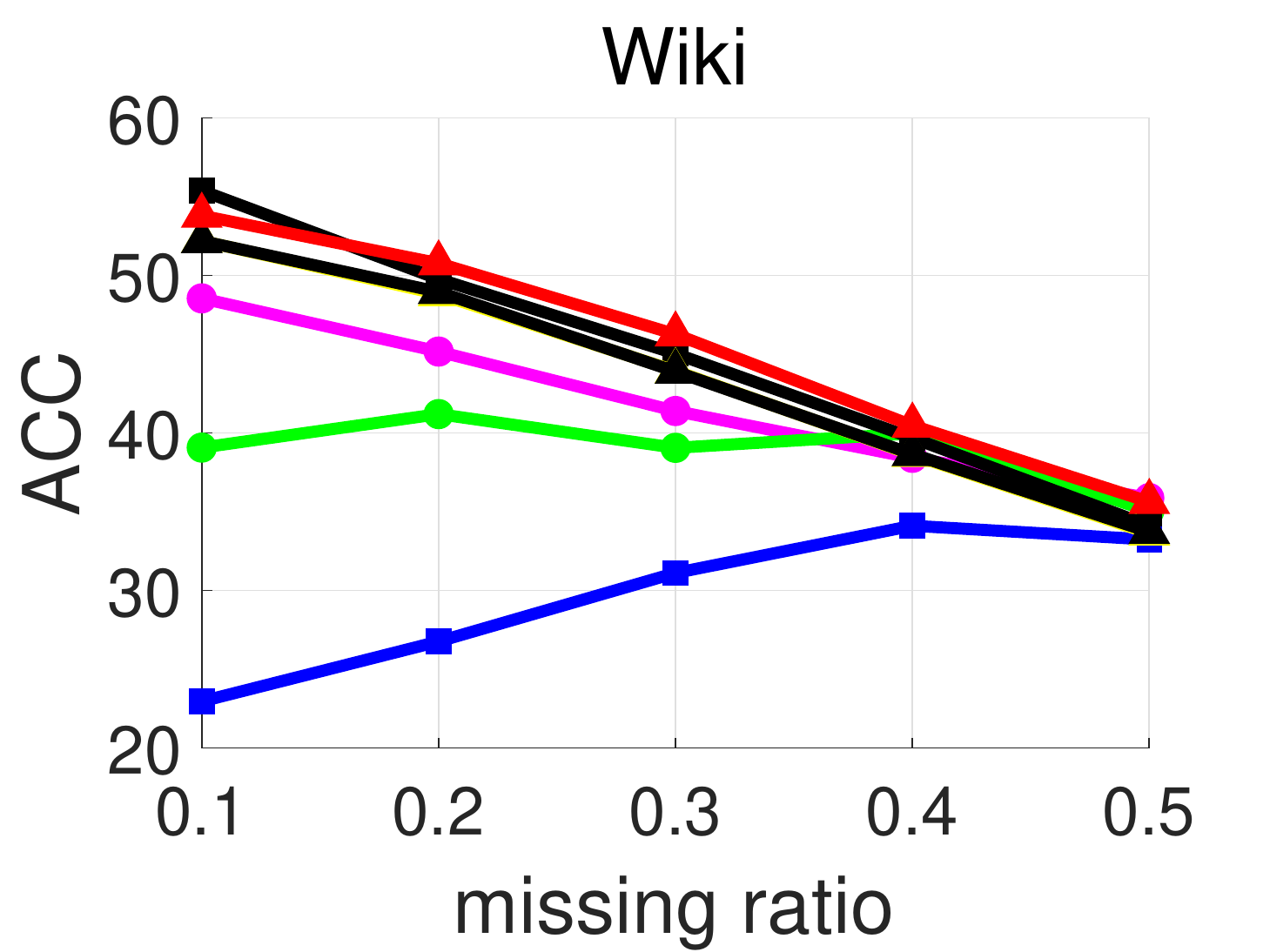}}
	\hspace{-0.2cm}
	\subfigure{
		\includegraphics[width=0.21\textwidth]{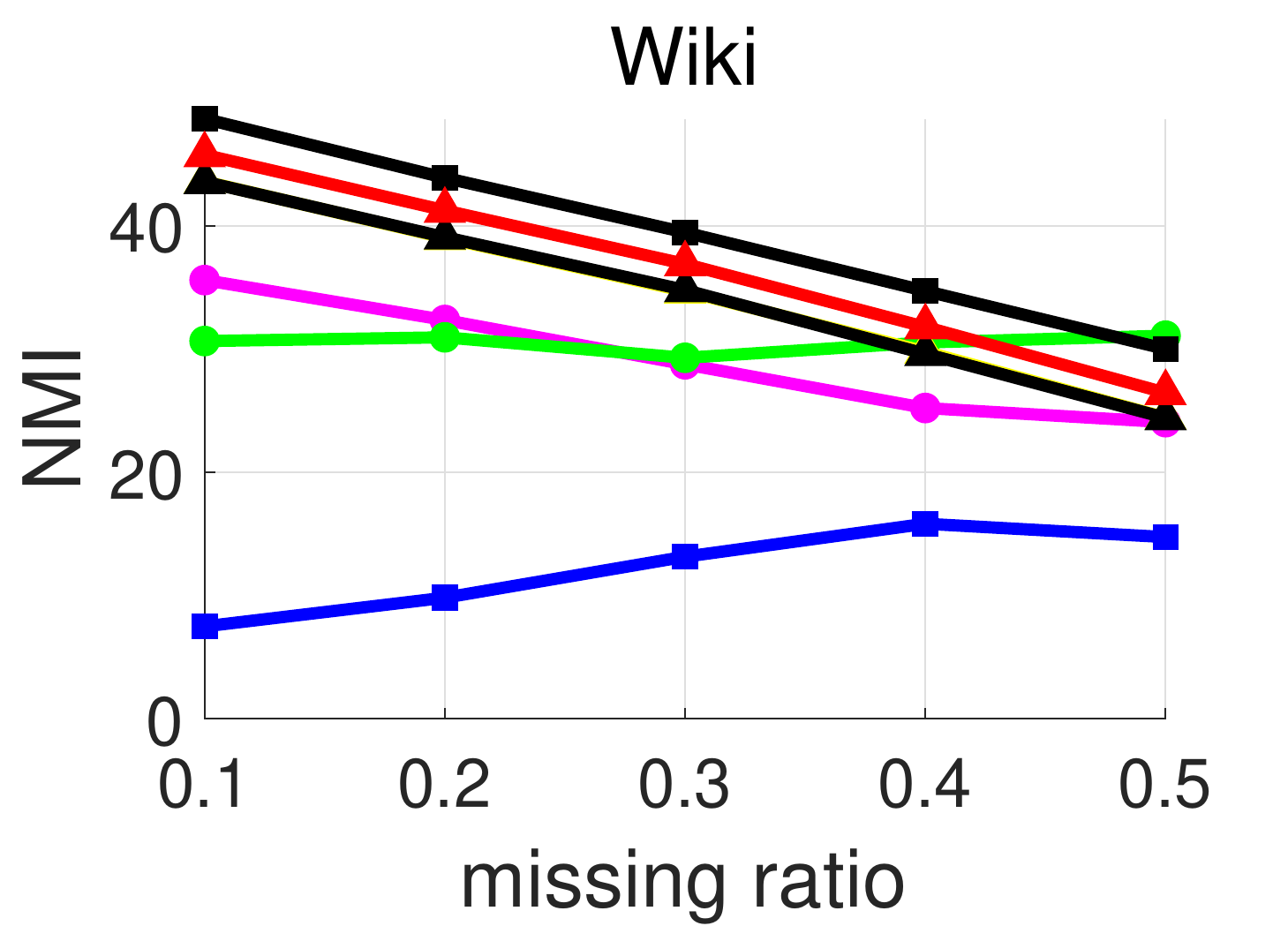}}
	\hspace{-0.2cm}
	\subfigure{
		\includegraphics[width=0.21\textwidth]{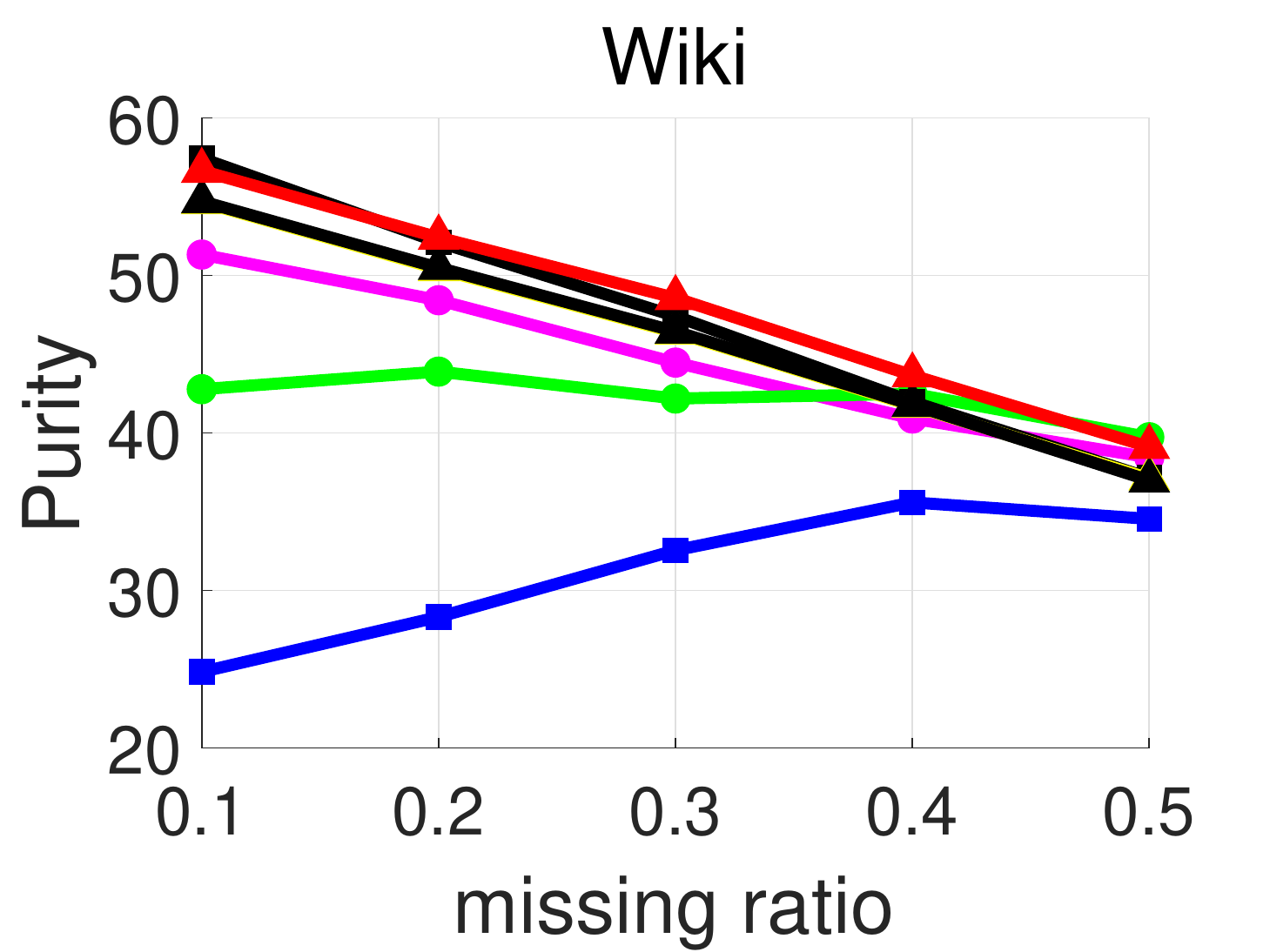}}
	\hspace{-0.2cm}
	\subfigure{
		\includegraphics[width=0.312\textwidth]{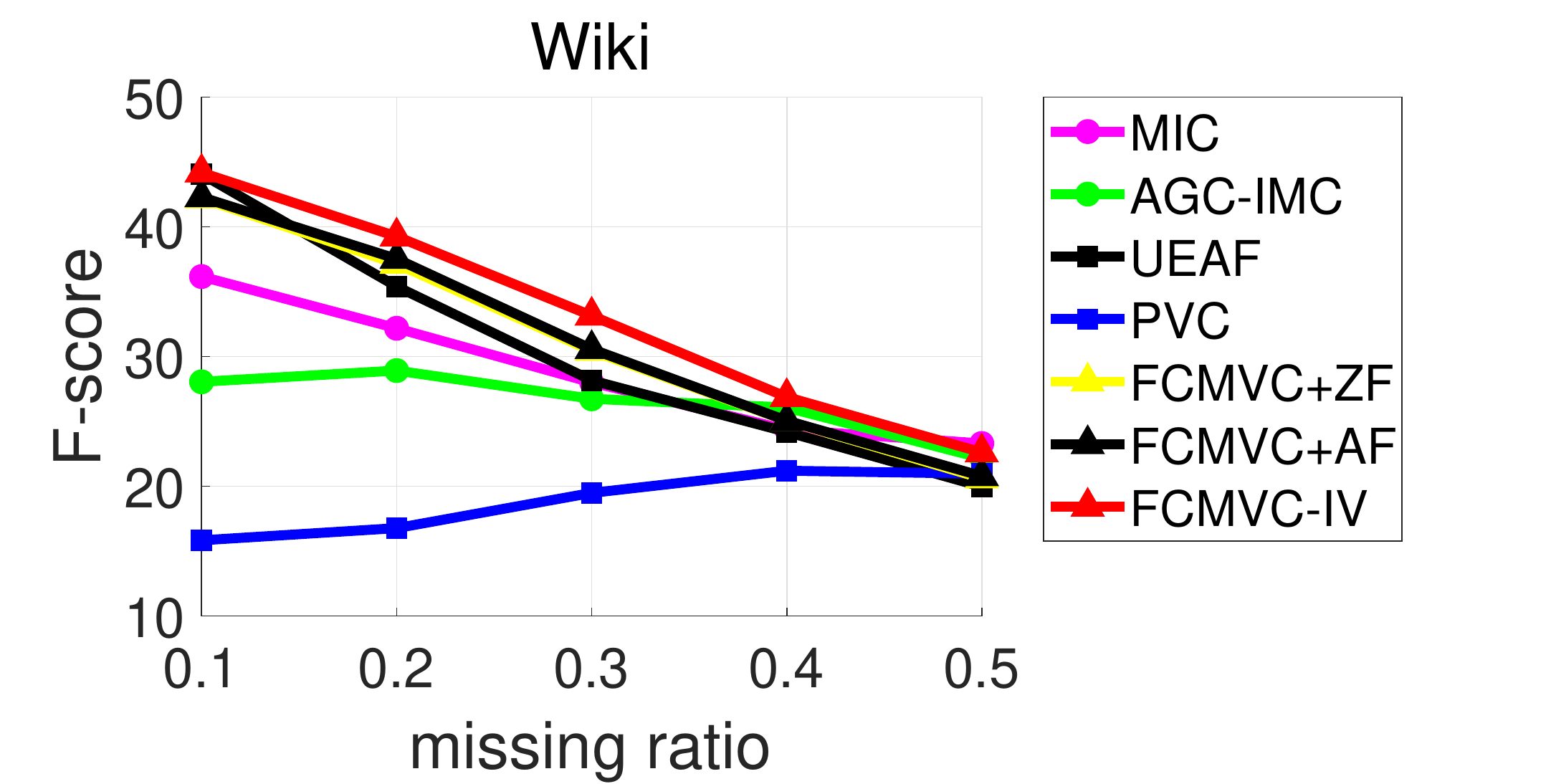}}
	\hspace{-0.2cm}
	\subfigure{
		\includegraphics[width=0.21\textwidth]{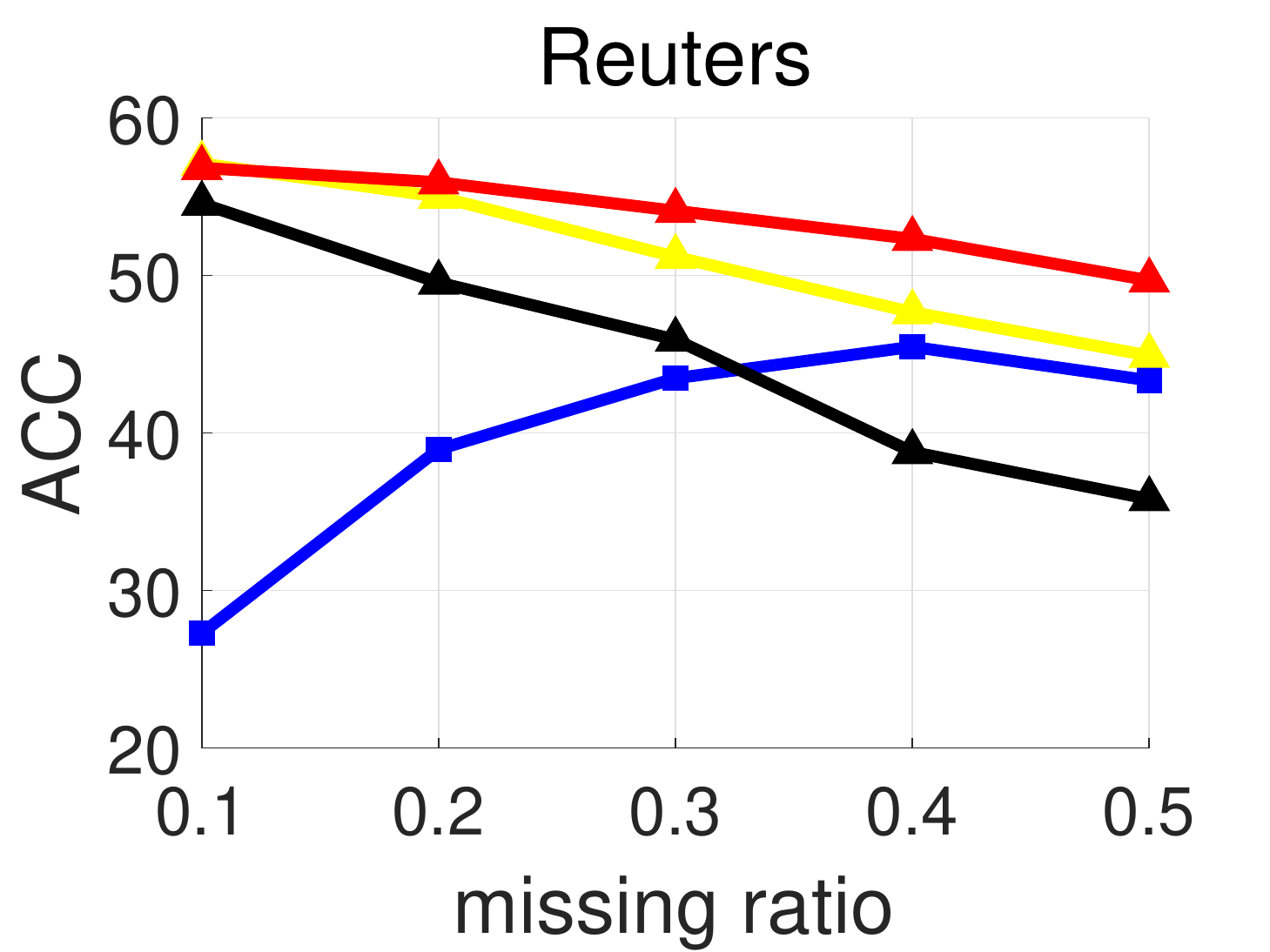}}
	\hspace{-0.2cm}
	\subfigure{
		\includegraphics[width=0.21\textwidth]{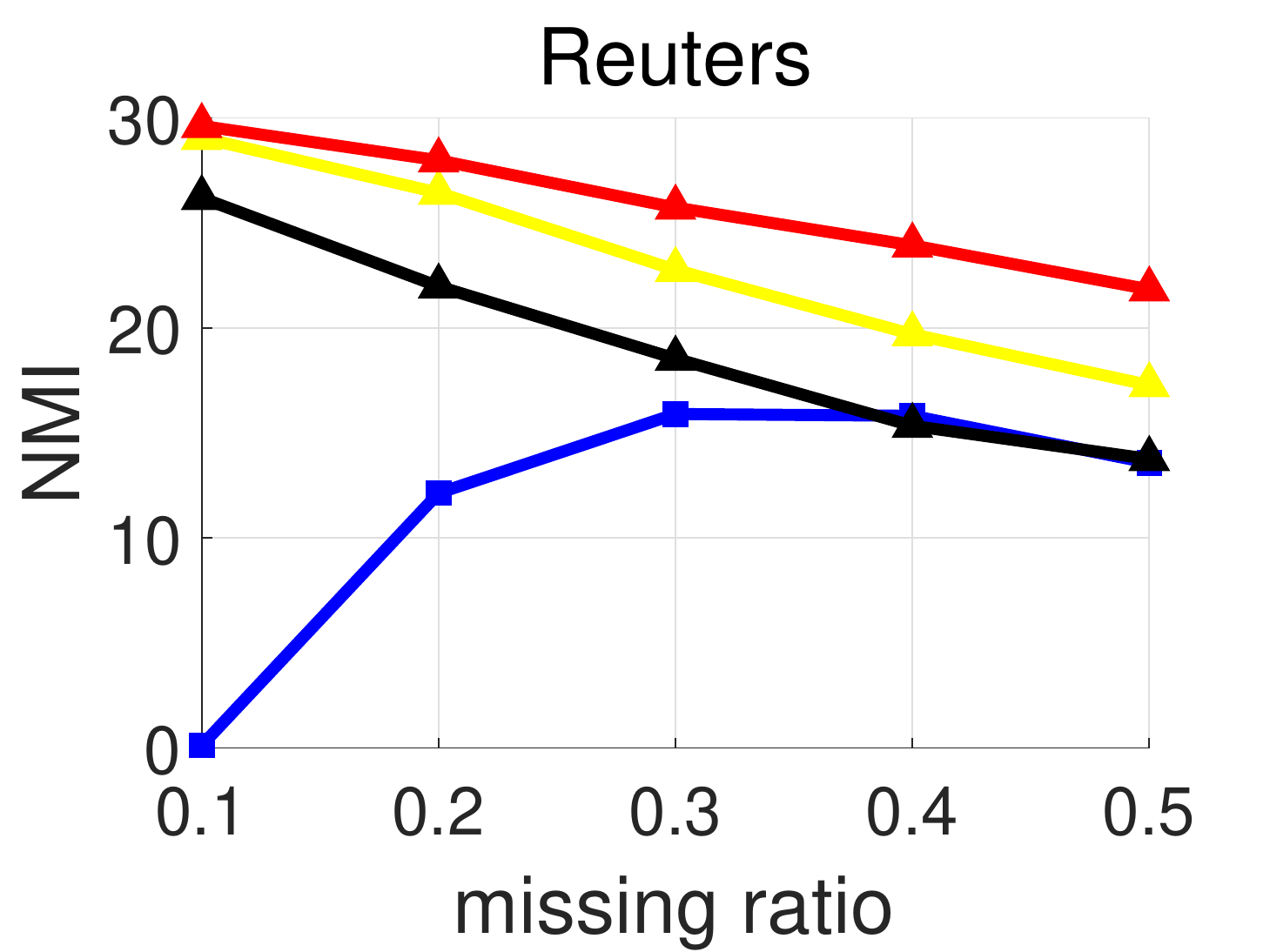}}
	\hspace{-0.2cm}
	\subfigure{
		\includegraphics[width=0.21\textwidth]{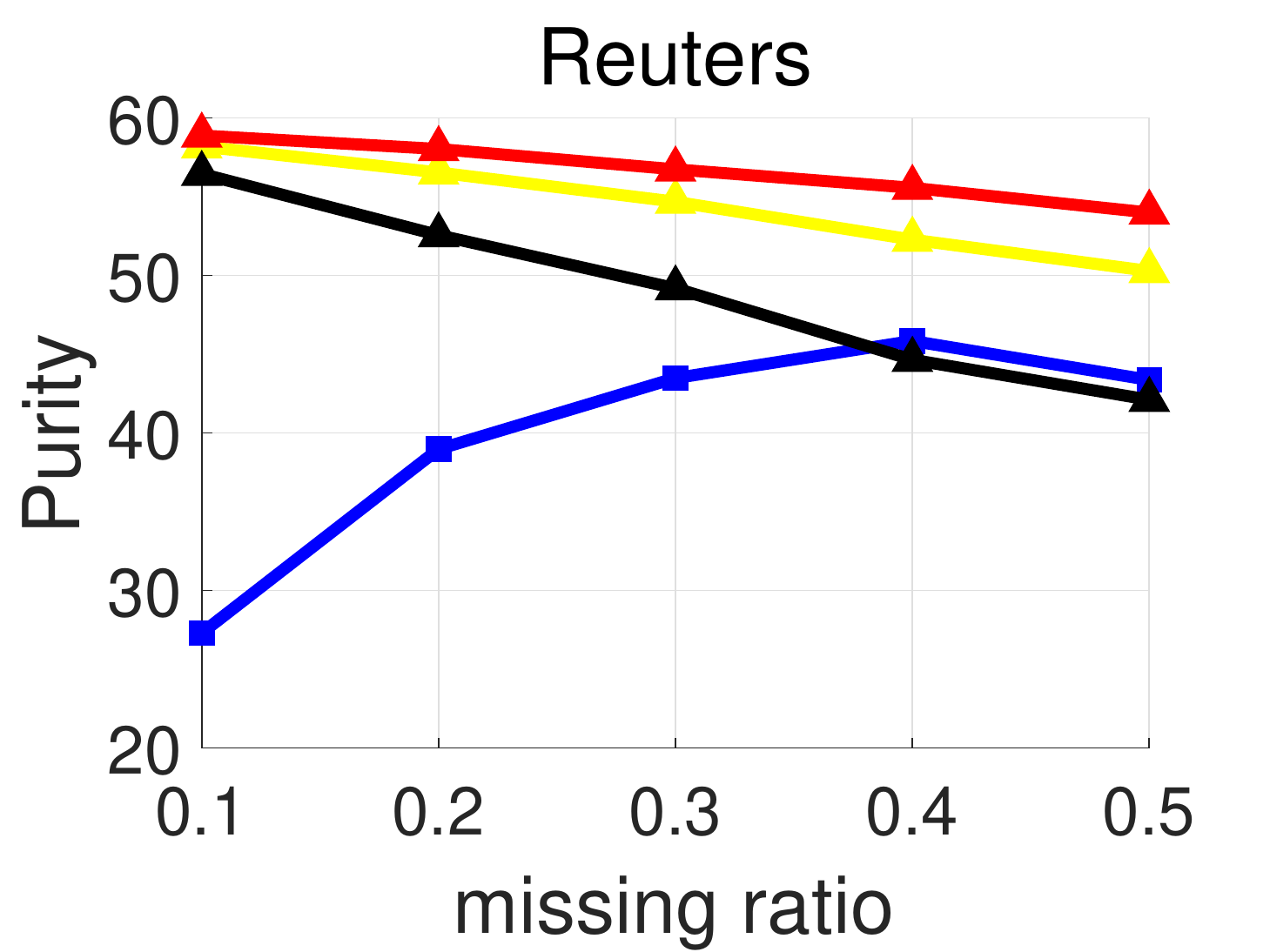}}
	\hspace{-0.2cm}
	\subfigure{
		\includegraphics[width=0.312\textwidth]{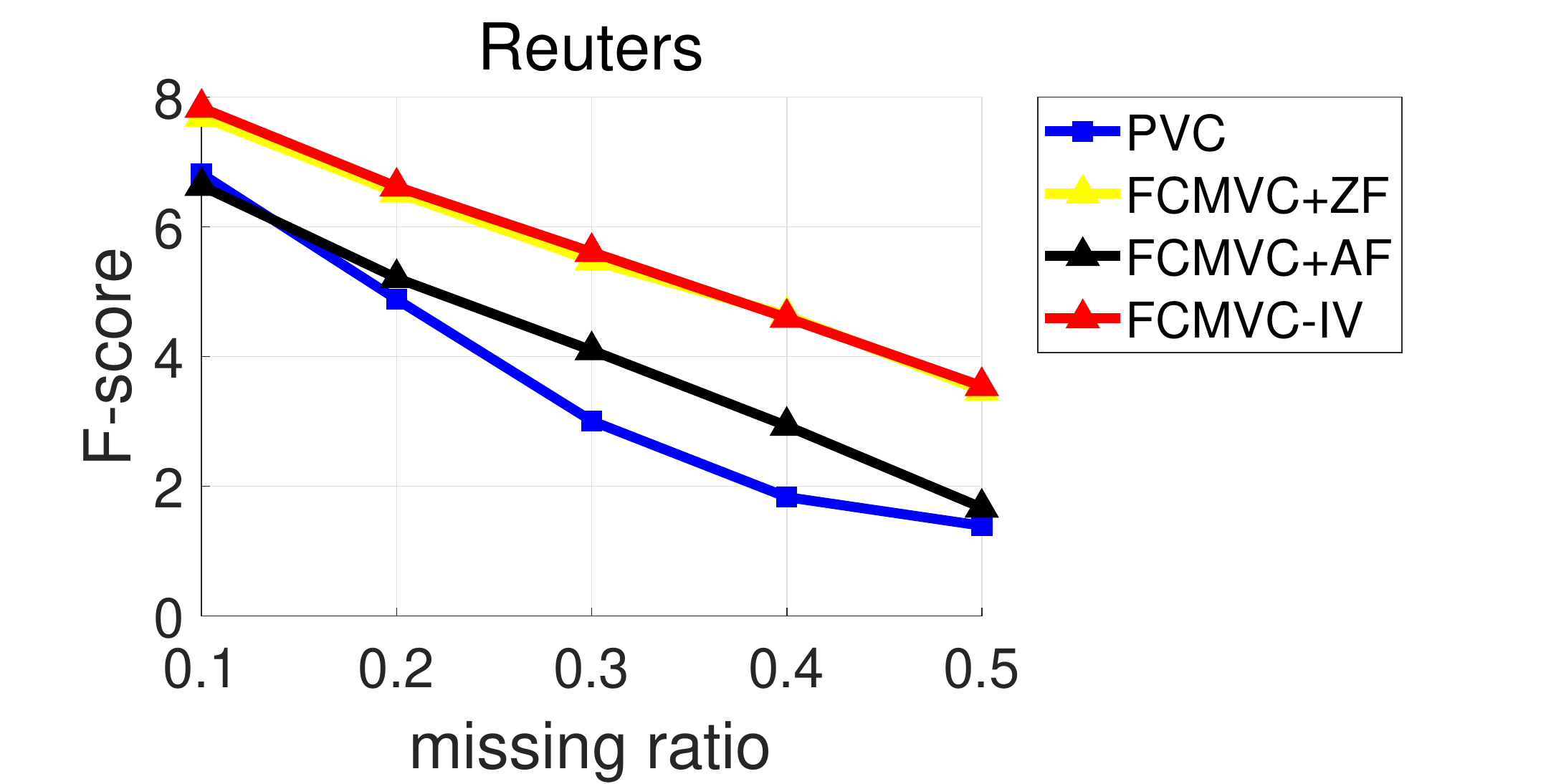}}
		\hspace{-0.2cm}
	
	\caption{The ACC, NMI, Purity, and F-score of different algorithms with the variation of missing ratios on five datasets. For each missing percentage, we randomly generate the “incomplete” patterns ten times and report the statistical results.}
	\label{fig_incom_res1}
\end{figure*}
\begin{figure*}[htbp]
	\centering
	\vspace{-0.2cm}
\subfigure{
		\includegraphics[width=0.21\textwidth]{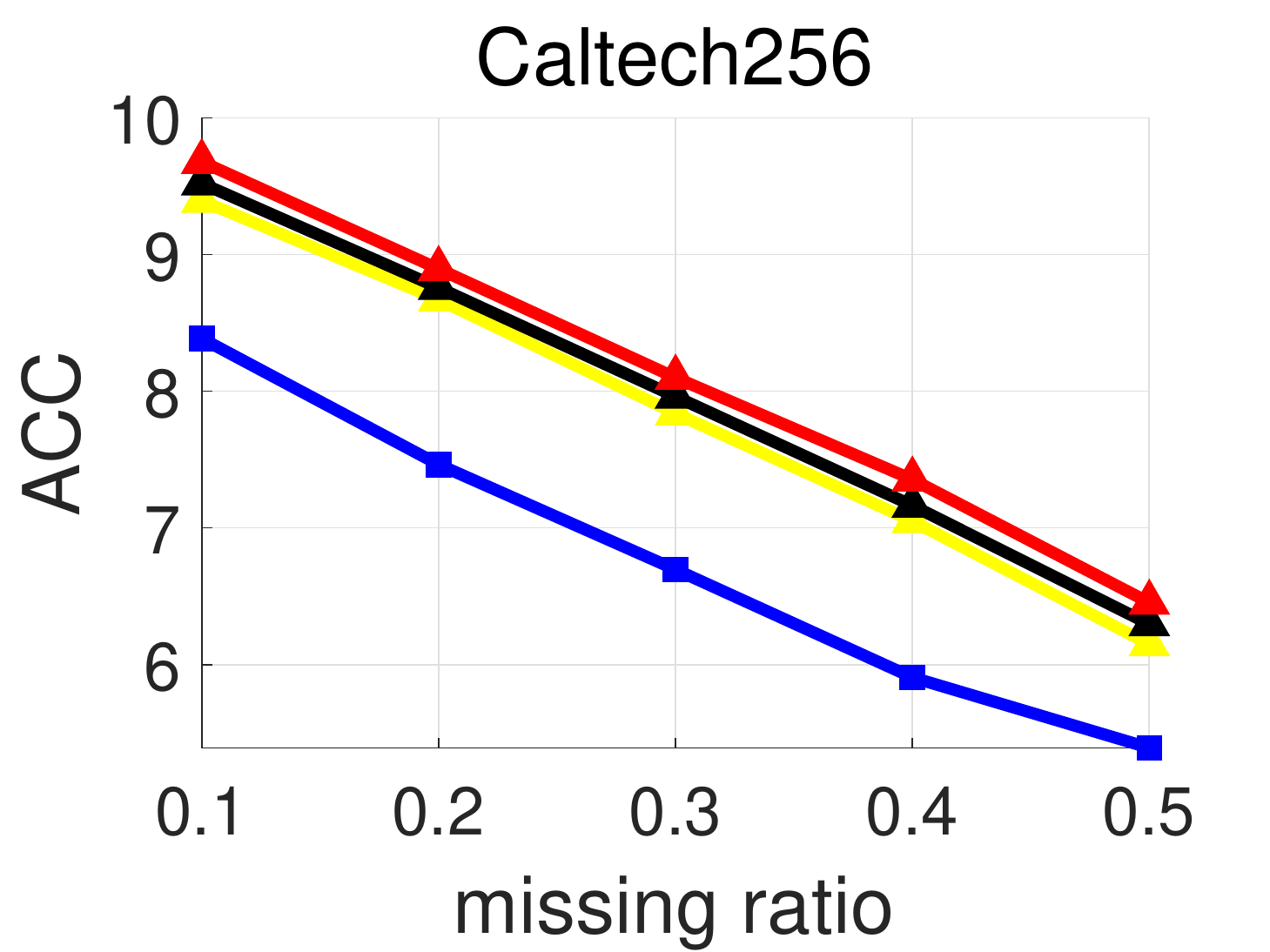}}
	\hspace{-0.2cm}
	\subfigure{
		\includegraphics[width=0.21\textwidth]{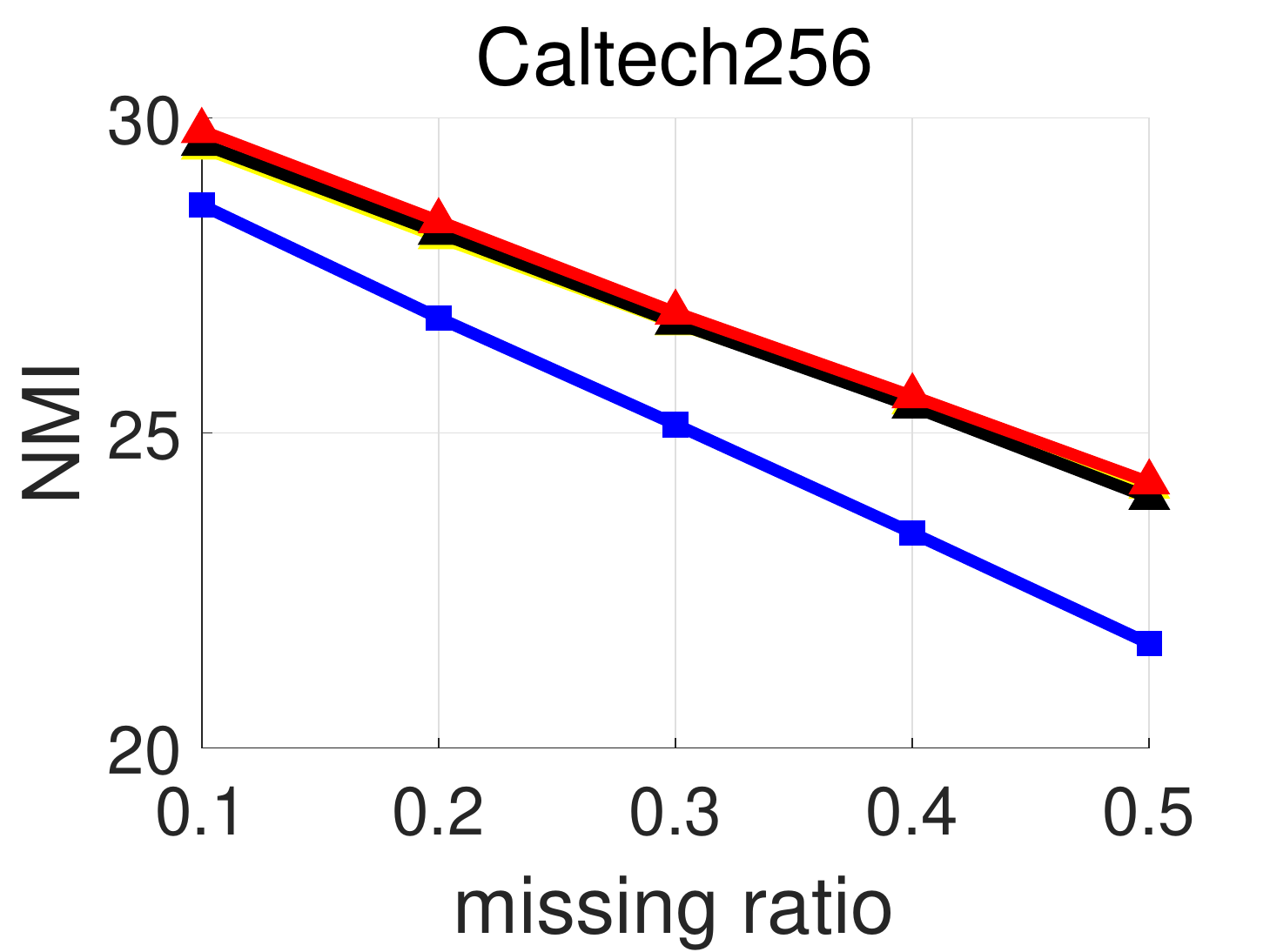}}
	\hspace{-0.2cm}
	\subfigure{
		\includegraphics[width=0.21\textwidth]{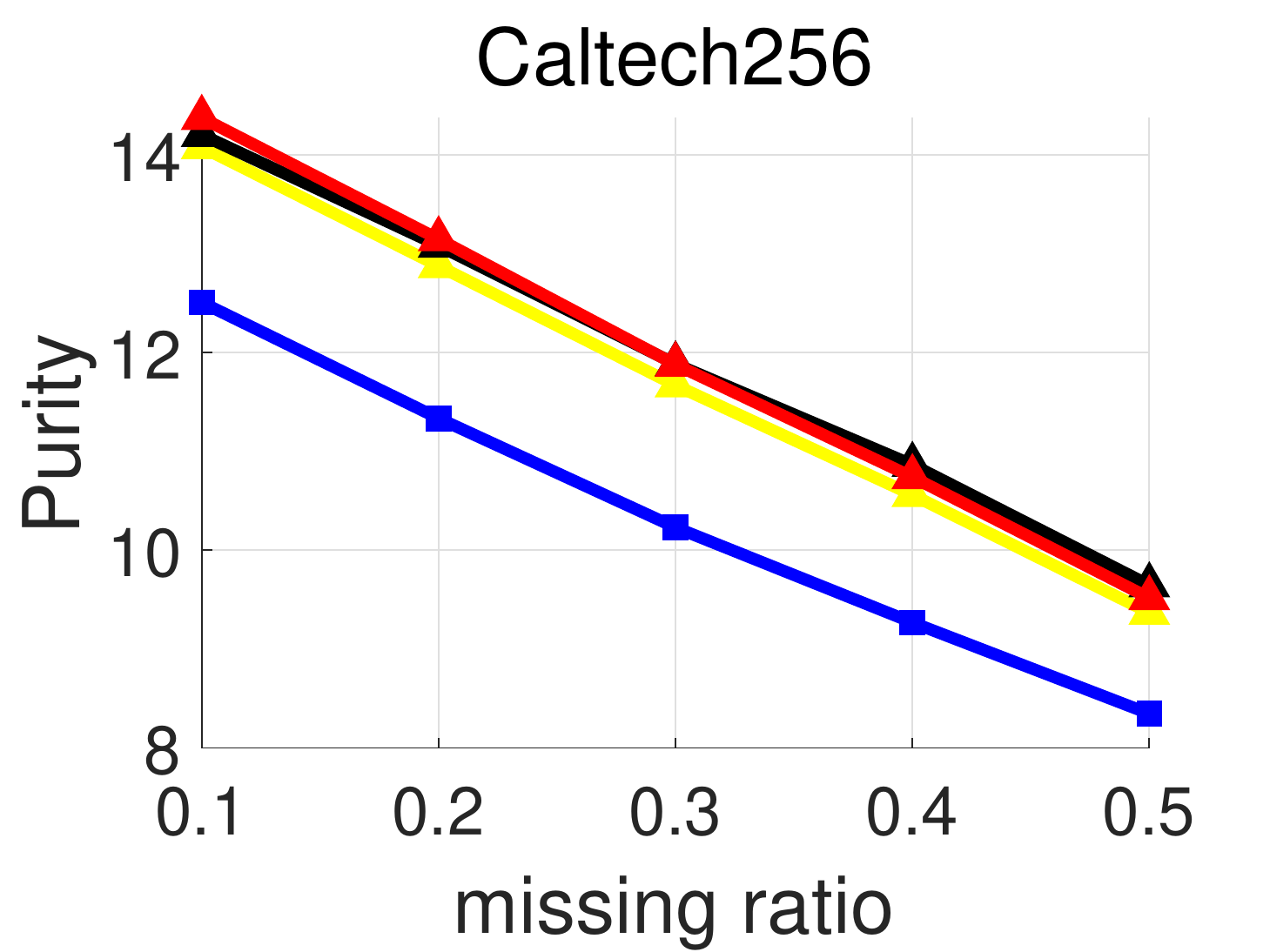}}
	\hspace{-0.2cm}
	\subfigure{
		\includegraphics[width=0.312\textwidth]{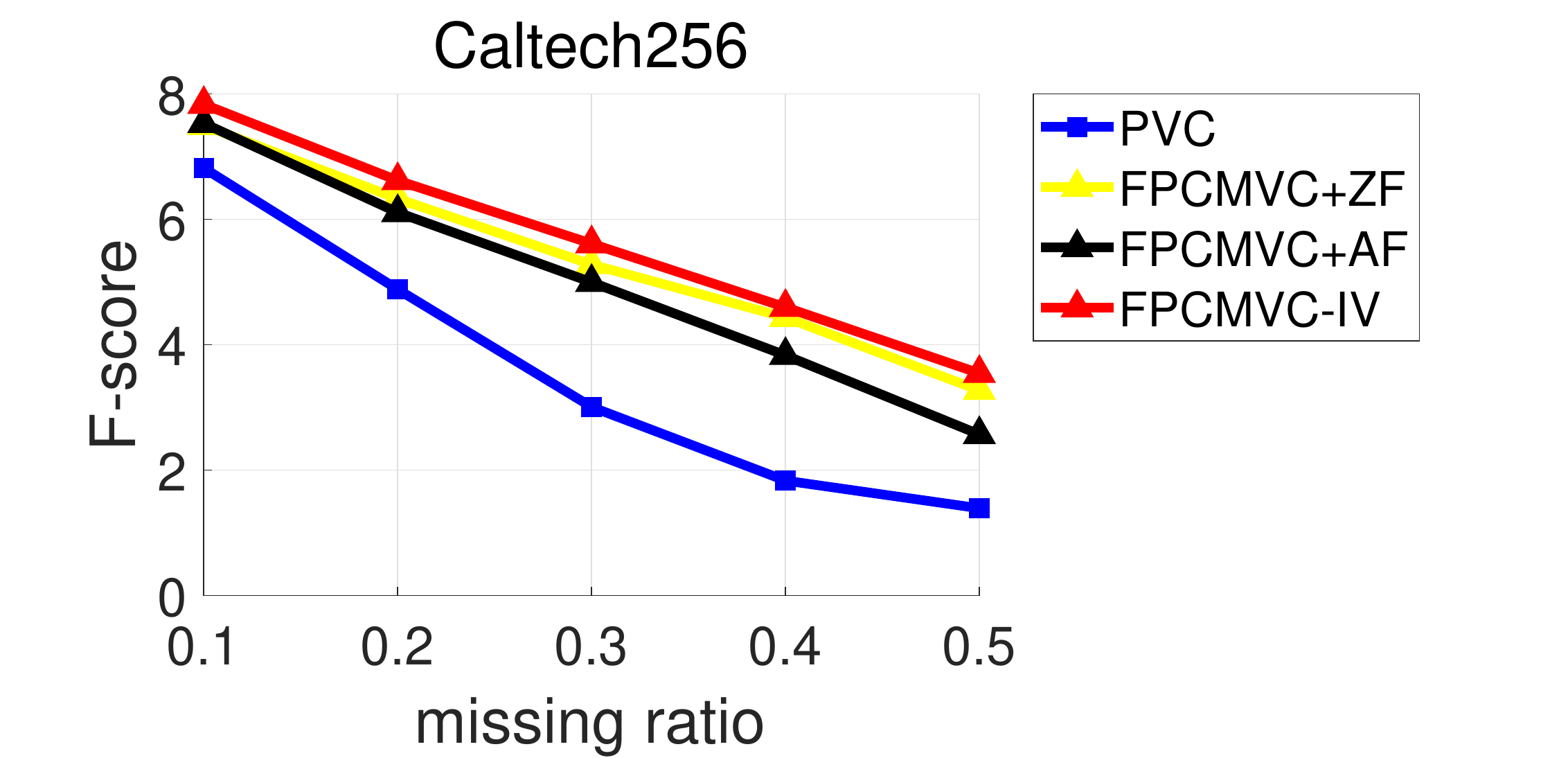}}
	\hspace{-0.2cm}
	\subfigure{
		\includegraphics[width=0.21\textwidth]{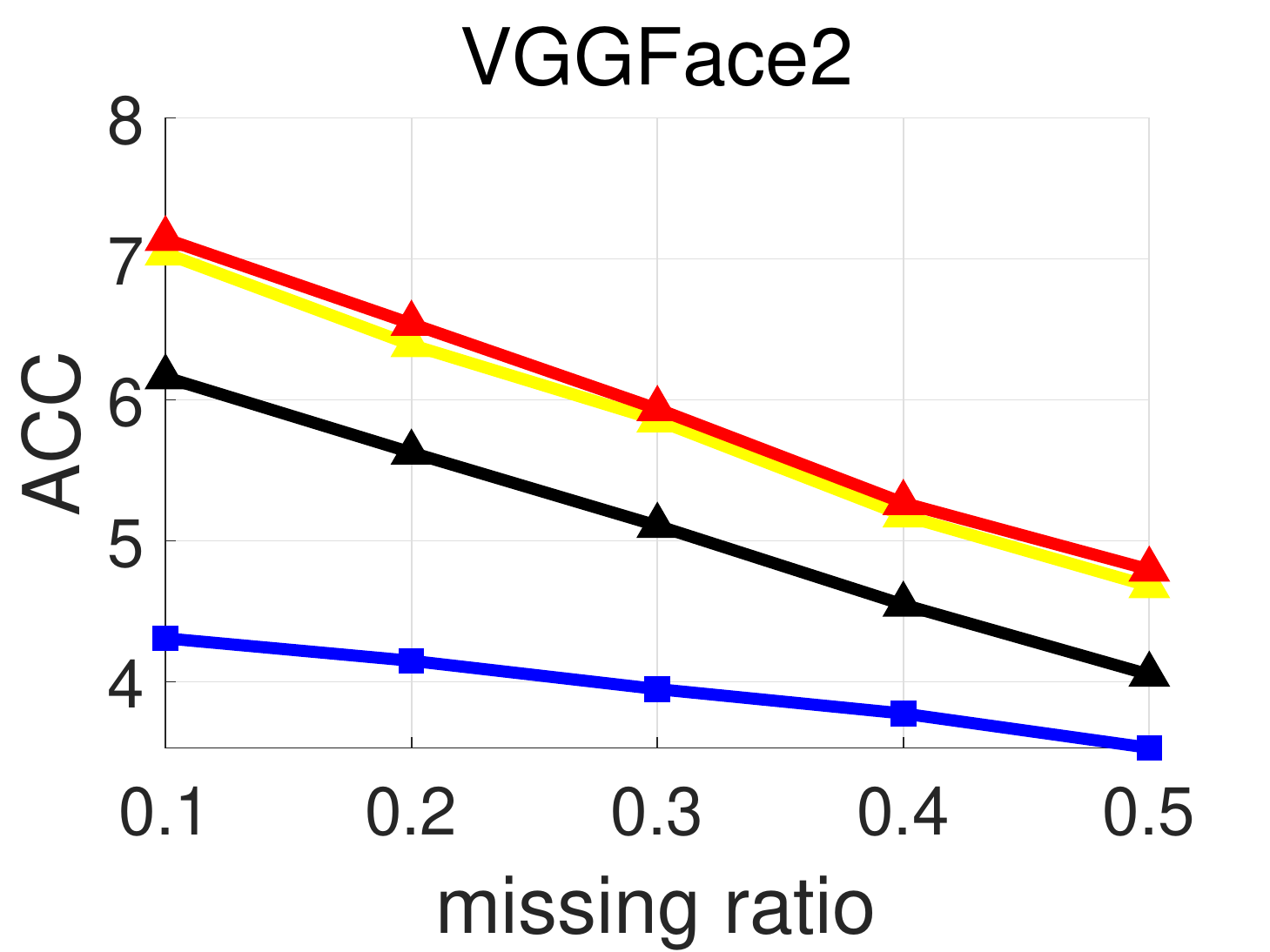}}
	\hspace{-0.2cm}
	\subfigure{
		\includegraphics[width=0.21\textwidth]{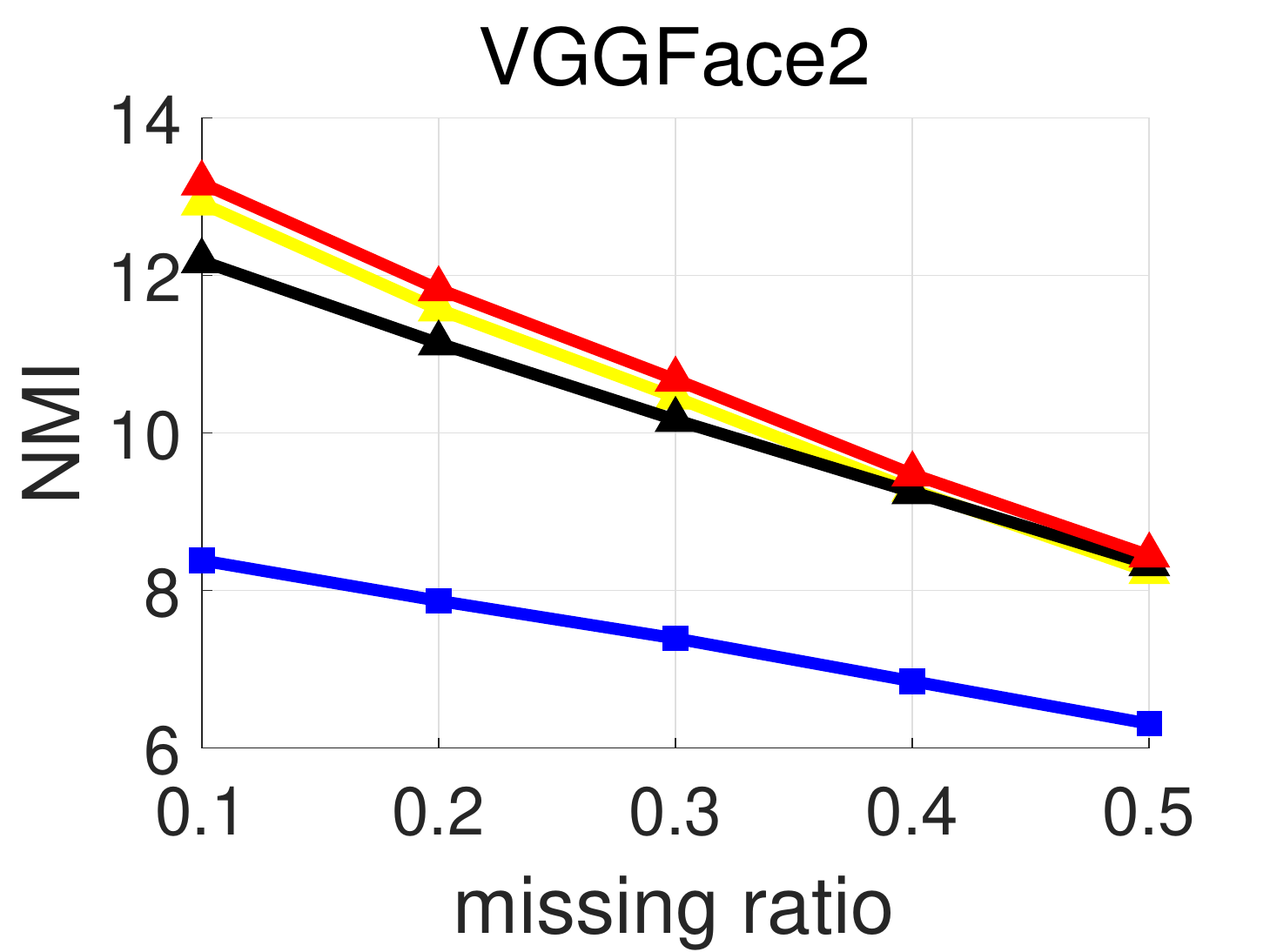}}
	\hspace{-0.2cm}
	\subfigure{
		\includegraphics[width=0.21\textwidth]{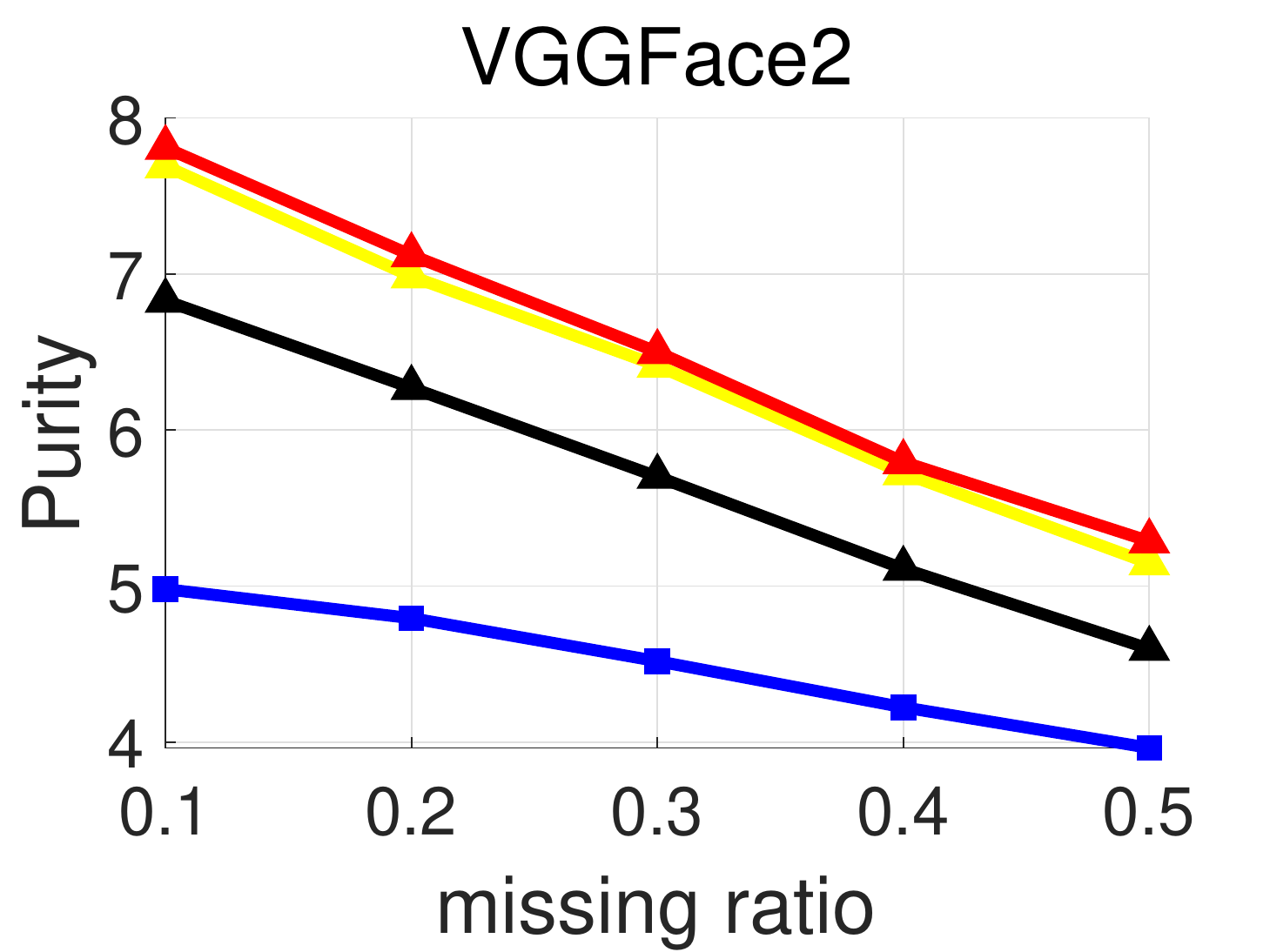}}
	\hspace{-0.2cm}
	\subfigure{
		\includegraphics[width=0.312\textwidth]{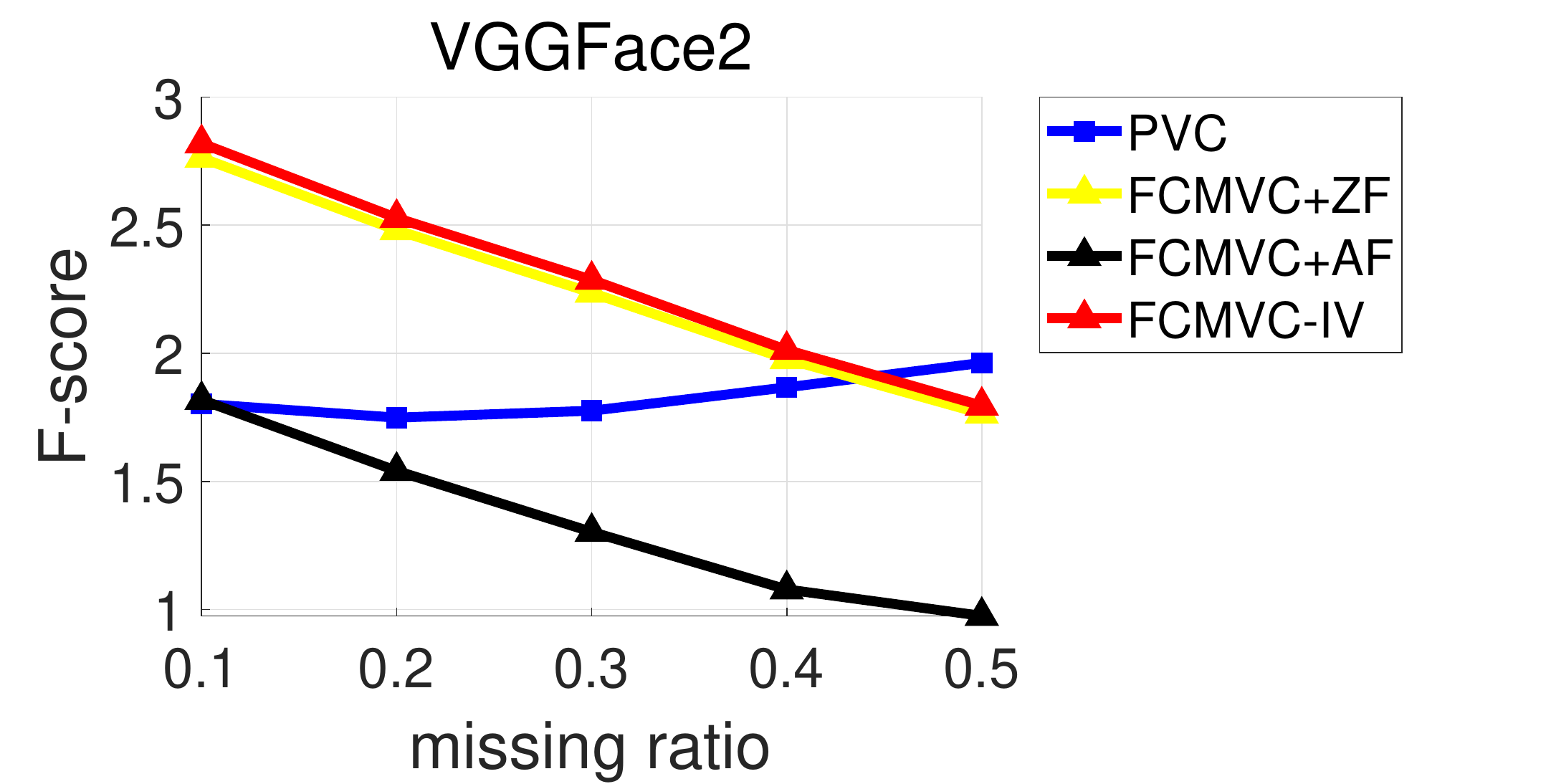}}
	\hspace{-0.2cm}
	\subfigure{
		\includegraphics[width=0.21\textwidth]{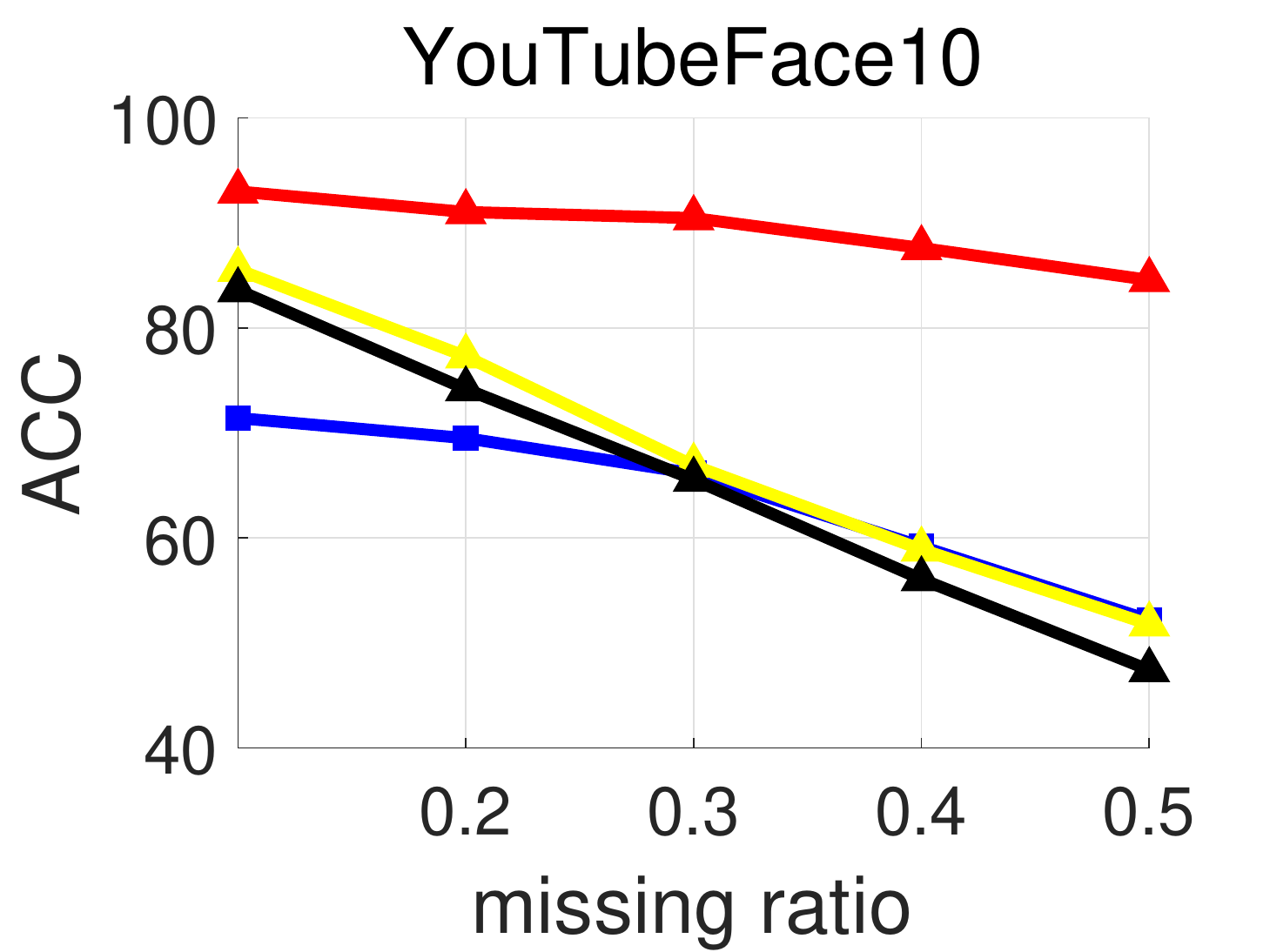}}
	\hspace{-0.2cm}
	\subfigure{
		\includegraphics[width=0.21\textwidth]{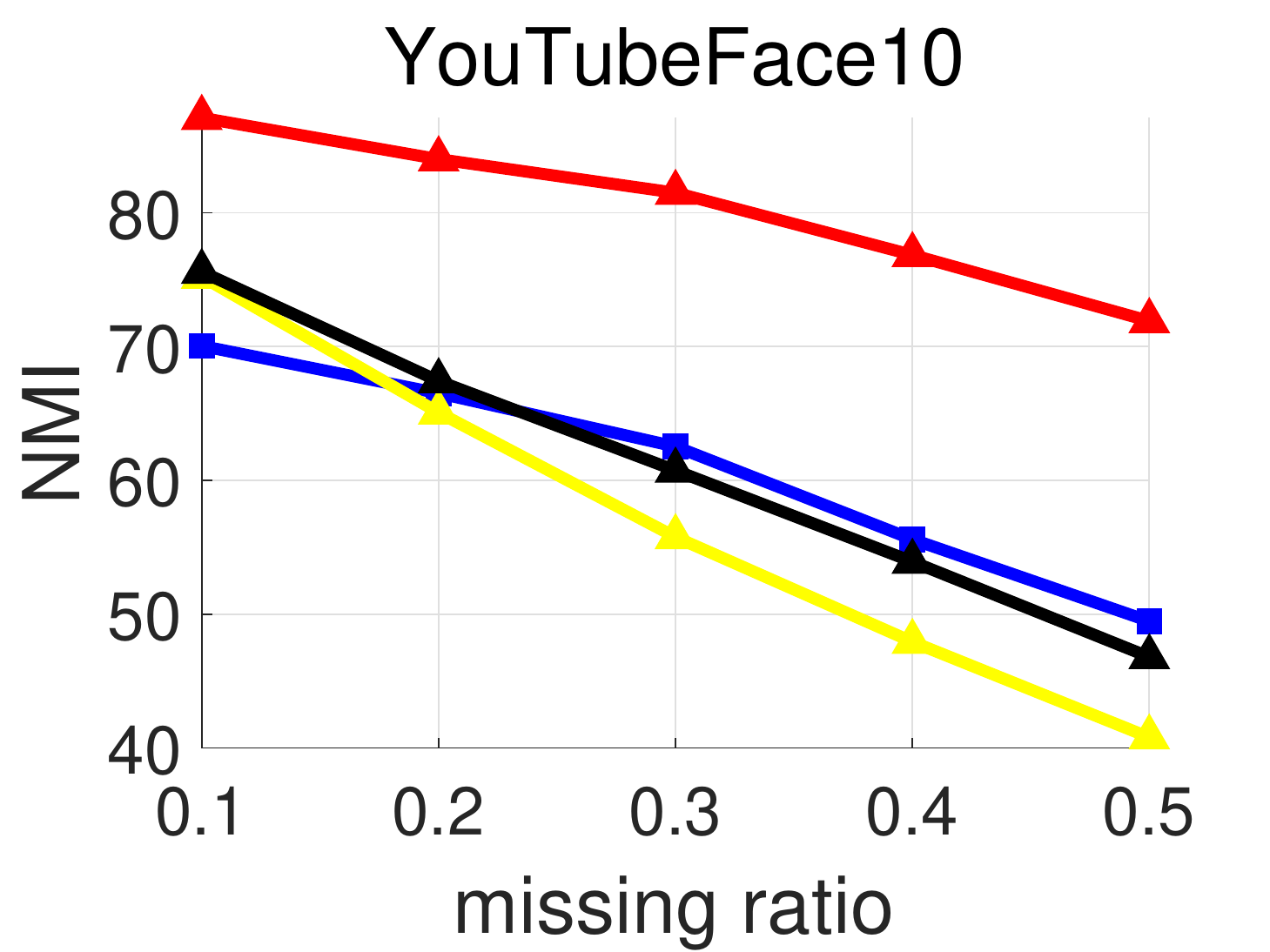}}
	\hspace{-0.2cm}
	\subfigure{
		\includegraphics[width=0.21\textwidth]{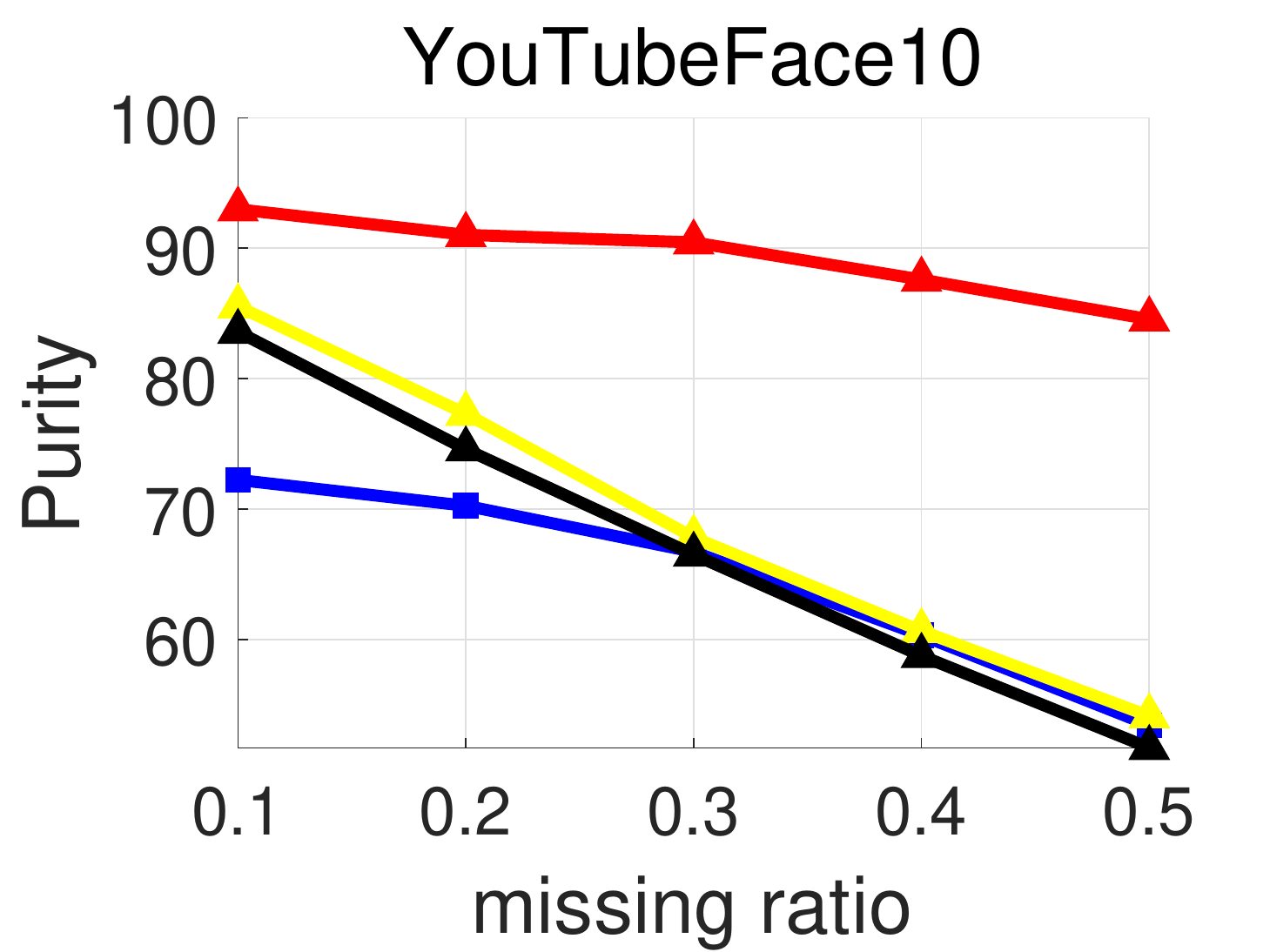}}
	\hspace{-0.2cm}
	\subfigure{
		\includegraphics[width=0.312\textwidth]{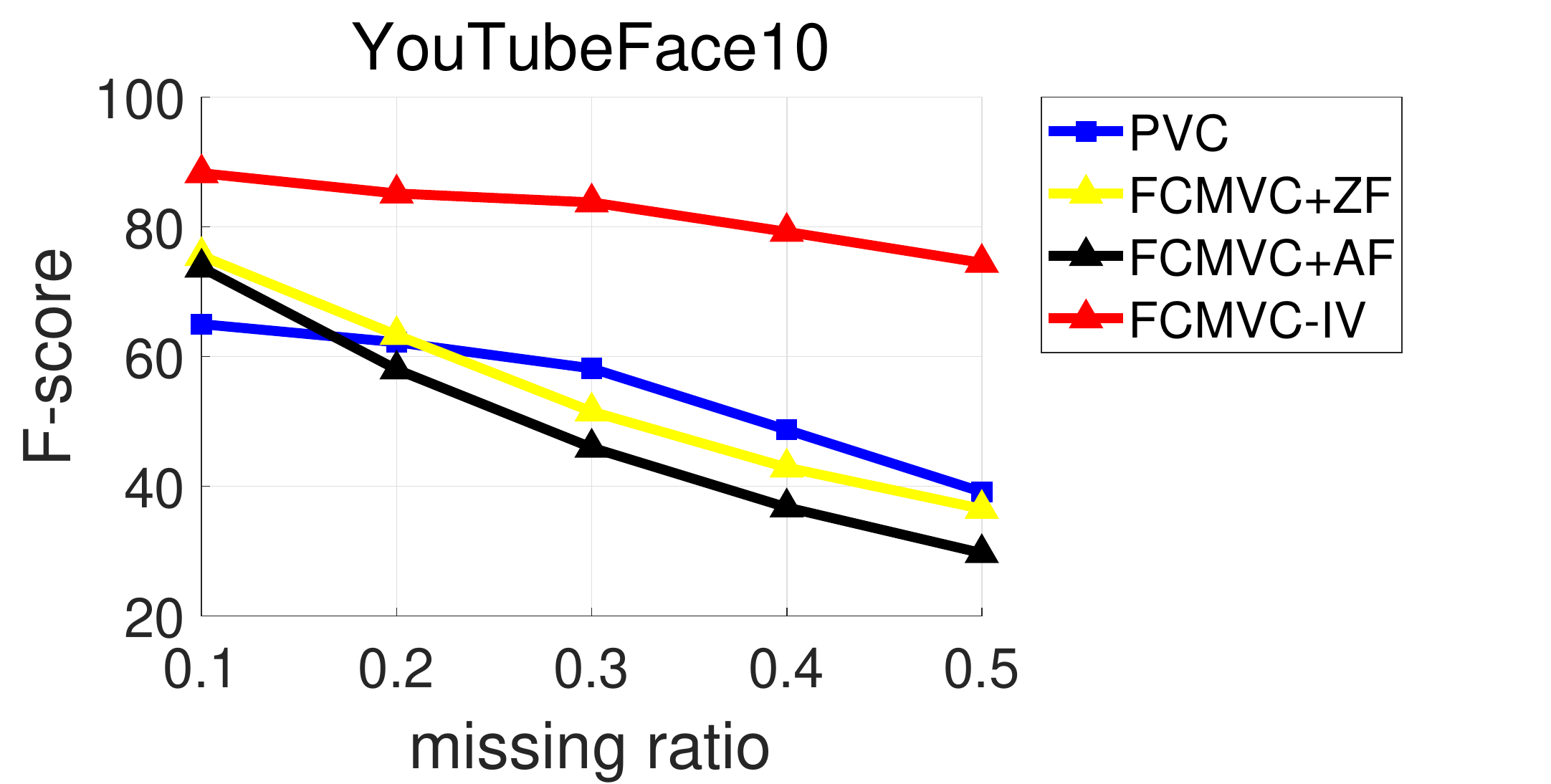}}
	\hspace{-0.2cm}
	
	\caption{The ACC, NMI, Purity, and F-score of different algorithms with the variation of missing ratios on three datasets. For each missing percentage, we randomly generate the “incomplete” patterns ten times and report the statistical results.}
	\label{fig_incom_res2}
\end{figure*}
 Eight benchmark datasets are used to testify the effectiveness of FCMVC-IV, including ORL\footnote{\url{https://cam-orl.co.uk/facedatabase.html}}, proteinFold\footnote{\url{http://mkl.ucsd.edu/dataset/protein-fold-prediction}}, uci-digit\footnote{\url{https://archive.ics.uci.edu/ml/index.php}}, Wiki\footnote{\url{http://www.svcl.ucsd.edu/projects/crossmodal/}}, 
 Reuters\footnote{\url{http://kdd.ics.uci.edu/databases/reuters21578/}}, Caltech256\footnote{\url{ https://www.kaggle.com/datasets/jessicali9530/caltech256/}}, VGGFace2\footnote{\url{http://www.robots.ox.ac.uk/~vgg/data/vgg_face2/}} and YouTubeFace10\footnote{\url{http://archive.ics.uci.edu/ml/datasets/YouTube+Multiview+Video+Games+Dataset}}. The detailed information of each dataset is summarized in Table \ref{dataset}.
 


As mentioned before, we compare algorithms with existing complete/incomplete MVC methods. We introduce them as follows.
\subsubsection{Complete MVC algorithms}
 \begin{enumerate}
 \item \textbf{Flexible Multi-View Representation Learning for Subspace Clustering (FMR)} \cite{ijcai2019p404}. It conducts subspace clustering based on flexible multi-view representation (FMR) learning.
 \item \textbf{Partition Level Multiview Subspace Clustering (PMSC)} \cite{Kang2019PartitionLM}. The algorithm proposes a unified multi-view subspace clustering model and incorporates the graph learning of each view.
\item \textbf{Parameter-free auto-weighted multiple graph learning (AMGL)} \cite{10.5555/3060832.3060884}. AMGL reformulates the standard spectral learning model and enables it to deal with multi-view tasks.
\item \textbf{Multi-view clustering via joint nonnegative matrix factorization (MNMF)} \cite{liu2013multi}. MNMF formulates a joint matrix factorization process by connecting NMF and PLSA.
\item \textbf{Large-scale Multi-view Subspace Clustering in Linear Time (LMVSC)} \cite{DBLP:conf/aaai/KangZZSHX20}. The algorithm integrates anchor graphs learned by each view, then implements spectral clustering on a smaller graph.
\item \textbf{One-pass Multi-view Clustering for Large-scale Data (OPMC)} \cite{9710821}. This work proposes a one-pass multi-view clustering algorithm and removes the non-negativity constraint of NMF.
\item \textbf{Fast Multi-View Clustering via Nonnegative and Orthogonal Factorization (FMCNOF)} \cite{9305974}. FMCNOF constrains a factor matrix of NMF with orthogonal constraints and obtains a direct cluster indicator matrix.
\item \textbf{Fast Parameter-Free Multi-View Subspace Clustering With Consensus Anchor Guidance (FPMVS)} \cite{9646486}. It combines anchor selection and subspace graph construction into a unified optimization formulation.
\item \textbf{Continual Multi-view Clustering (CMVC)} \cite{10.1145/3503161.3547864}. It combines continual learning and late fusion into a unified framework.
 \end{enumerate}
 
\subsubsection{Incomplete MVC algorithms}
\begin{enumerate}
 \item \textbf{Incomplete Multi-Modal Visual Data Grouping (IMG)} \cite{Zhao2016IncompleteMV}. IMG maps the original and incomplete data into a new and complete representation in a latent space.
 \item \textbf{Multiple Incomplete Views Clustering via Weighted Nonnegative Matrix Factorization with L2, 1 Regularization (MIC)} \cite{10.1007/978-3-319-23528-8_20}. This work generates a consensus matrix and minimizes the consensus among views.
 \item \textbf{Adaptive Graph Completion Based Incomplete Multi-View Clustering (AGC-IMC)} \cite{9154578}. The algorithm develops a joint framework for graph completion and consensus representation learning.
 \item \textbf{Unified embedding alignment with missing views inferring for incomplete multi-view clustering (UEAF)} \cite{Wen2019UnifiedEA}. The algorithm proposes a Unified Embedding Alignment Framework for robust incomplete multi-view clustering.
 \item \textbf{Partial Multi-View Clustering (PVC)} \cite{Li2014PartialMC}. PVC factorizes data matrices of each view to learn the latent subspaces individually, but it requires some samples shared among views.
 \item \textbf{FCMVC+ZF}. The algorithm conducts FCMVC with the zero-filling method.
 \item \textbf{FCMVC+AF}. The algorithm conducts FCMVC with the average-filling method.
\end{enumerate}

The implementations of the above algorithms are publicly available online, and we directly use these codes without any changes. As for algorithms with hyper-parameters, we adopt the same strategies as these papers to tune the hyper-parameters and choose the best results to report. We evaluate the performance of algorithms through four widely used metrics, including accuracy (ACC), normalized mutual information (NMI), purity, and F-score. 

We assume that the number of clusters $k$ is provided for all datasets and set it as the actual number of classes. To generate the incomplete views, we randomly remove $n\times r$ samples of each view except the last one, where $r$ denotes the missing ratio. To prevent instances from missing in all views, we remove the same number from the items existing in the previous views at random for the final view. Five missing ratios ranging from 0.1 to 0.5 are considered in our experiments. For each missing percentage, we randomly generate the incomplete patterns ten times and report the statistical results.

To reduce the impact of the randomness of k-means, each algorithm performs k-means 50 times and reports the average. Our experiments are implemented on a desktop computer with an Intel(R) Core(TM) i9-10850K CPU and 96 GB RAM, MATLAB 2020b (64-bit).

 \subsection{Experimental Results} 

\begin{table*}[]
\centering
	\caption{Aggregated ACC, NMI, Purity and F-score comparison (mean±std) of different clustering algorithms on all incomplete benchmark datasets.}
	\label{incomlete_result}
	
\begin{tabular}{ccccccccc}
\toprule
\multicolumn{1}{c}{{\color[HTML]{000000} }} & \multicolumn{1}{c}{\begin{tabular}[c]{@{}c@{}}IMG\\ \cite{Zhao2016IncompleteMV}\end{tabular}} & \multicolumn{1}{c}{\begin{tabular}[c]{@{}c@{}}MIC\\ \cite{10.1007/978-3-319-23528-8_20}\end{tabular}} & \multicolumn{1}{c}{\begin{tabular}[c]{@{}c@{}}AGC-IMC\\ \cite{9154578}\end{tabular}} & \multicolumn{1}{c}{\begin{tabular}[c]{@{}c@{}}UEAF\\ \cite{Wen2019UnifiedEA}\end{tabular}} & \multicolumn{1}{c}{\begin{tabular}[c]{@{}c@{}}PVC\\ \cite{Li2014PartialMC}\end{tabular}} & \multicolumn{1}{c}{\begin{tabular}[c]{@{}c@{}}FCMVC+ZF\\ proposed\end{tabular}} & \multicolumn{1}{c}{\begin{tabular}[c]{@{}c@{}}FCMVC+AF\\ proposed\end{tabular}} & \begin{tabular}[c]{@{}c@{}}FCMVC-IV\\ proposesd\end{tabular} \\ \hline
\multicolumn{1}{c}{Param. num.}             & \multicolumn{1}{c}{3}          & \multicolumn{1}{c}{2}           & \multicolumn{1}{c}{2}                   & \multicolumn{1}{c}{3}                   & \multicolumn{1}{c}{1}                & \multicolumn{1}{c}{0}                   & \multicolumn{1}{c}{0}                   & 0                    \\ \hline
\multicolumn{9}{c}{ACC}                                                                                                                                                                                                                                                                                                                                     \\ \hline
\multicolumn{1}{c}{ORL}                     & \multicolumn{1}{c}{29.86±8.74} & \multicolumn{1}{c}{41.76±10.31} & \multicolumn{1}{c}{{\ul 56.68±12.63}}   & \multicolumn{1}{c}{46.70±8.07}          & \multicolumn{1}{c}{52.32±8.33}       & \multicolumn{1}{c}{53.05±9.80}          & \multicolumn{1}{c}{52.66±10.05}         & \textbf{66.49±7.22}  \\ 
\multicolumn{1}{c}{proteinFold}             & \multicolumn{1}{c}{18.11±0.87} & \multicolumn{1}{c}{11.53±0.00}  & \multicolumn{1}{c}{23.67±1.12}          & \multicolumn{1}{c}{{\ul 27.58±1.74}}    & \multicolumn{1}{c}{18.96±2.60}       & \multicolumn{1}{c}{27.39±3.42}          & \multicolumn{1}{c}{27.37±3.25}          & \textbf{31.19±2.89}  \\ 
\multicolumn{1}{c}{uci-digit}               & \multicolumn{1}{c}{36.38±7.15} & \multicolumn{1}{c}{10.05±0.00}  & \multicolumn{1}{c}{{\ul 70.03±1.91}}    & \multicolumn{1}{c}{48.32±7.34}          & \multicolumn{1}{c}{48.47±5.51}       & \multicolumn{1}{c}{59.09±12.41}         & \multicolumn{1}{c}{59.36±11.22}         & \textbf{72.94±6.89}  \\ 
\multicolumn{1}{c}{Wiki}                    & \multicolumn{1}{c}{-}          & \multicolumn{1}{c}{41.87±4.55}  & \multicolumn{1}{c}{38.86±2.07}          & \multicolumn{1}{c}{{\ul 44.79±7.45}}    & \multicolumn{1}{c}{29.63±4.20}       & \multicolumn{1}{c}{43.43±6.72}          & \multicolumn{1}{c}{43.48±6.70}          & \textbf{45.38±6.63}  \\ 
\multicolumn{1}{c}{Reuters}                 & \multicolumn{1}{c}{-}          & \multicolumn{1}{c}{-}           & \multicolumn{1}{c}{-}                   & \multicolumn{1}{c}{-}                   & \multicolumn{1}{c}{39.70±6.57}       & \multicolumn{1}{c}{{\ul 51.16±4.50}}    & \multicolumn{1}{c}{44.92±6.87}          & \textbf{53.76±2.56}  \\ 
\multicolumn{1}{c}{Caltech256}              & \multicolumn{1}{c}{-}          & \multicolumn{1}{c}{-}           & \multicolumn{1}{c}{-}                   & \multicolumn{1}{c}{-}                   & \multicolumn{1}{c}{6.77±1.07}        & \multicolumn{1}{c}{7.82±1.15}           & \multicolumn{1}{c}{{\ul 7.94±1.14}}     & \textbf{8.10±1.13}   \\ 
\multicolumn{1}{c}{VGGFace2}                & \multicolumn{1}{c}{-}          & \multicolumn{1}{c}{-}           & \multicolumn{1}{c}{-}                   & \multicolumn{1}{c}{-}                   & \multicolumn{1}{c}{3.94±0.27}        & \multicolumn{1}{c}{{\ul 5.83±0.84}}     & \multicolumn{1}{c}{5.10±0.75}           & \textbf{5.94±0.84}   \\ 
\multicolumn{1}{c}{YouTubeFace10}           & \multicolumn{1}{c}{-}          & \multicolumn{1}{c}{-}           & \multicolumn{1}{c}{-}                   & \multicolumn{1}{c}{-}                   & \multicolumn{1}{c}{63.67±7.10}       & \multicolumn{1}{c}{{\ul 68.07±12.18}}   & \multicolumn{1}{c}{65.36±12.79}         & \textbf{89.33±2.94}  \\ \hline
\multicolumn{9}{c}{NMI}                                                                                                                                                                                                                                                                                                                                     \\ \hline
\multicolumn{1}{c}{ORL}                     & \multicolumn{1}{c}{53.88±6.66} & \multicolumn{1}{c}{59.80±8.92}  & \multicolumn{1}{c}{{\ul 73.29±9.47}}    & \multicolumn{1}{c}{65.96±6.82}          & \multicolumn{1}{c}{69.71±7.12}       & \multicolumn{1}{c}{70.40±7.17}          & \multicolumn{1}{c}{69.09±7.68}          & \textbf{79.88±5.99}  \\ 
\multicolumn{1}{c}{proteinFold}             & \multicolumn{1}{c}{26.97±1.71} & \multicolumn{1}{c}{8.40±0.00}   & \multicolumn{1}{c}{30.78±1.80}          & \multicolumn{1}{c}{{\ul 35.73±2.32}}    & \multicolumn{1}{c}{25.16±3.53}       & \multicolumn{1}{c}{34.46±4.01}          & \multicolumn{1}{c}{34.62±3.79}          & \textbf{38.26±3.61}  \\ 
\multicolumn{1}{c}{uci-digit}               & \multicolumn{1}{c}{36.55±5.03} & \multicolumn{1}{c}{0.89±0.00}   & \multicolumn{1}{c}{\textbf{56.94±3.04}} & \multicolumn{1}{c}{43.62±5.87}          & \multicolumn{1}{c}{38.29±7.21}       & \multicolumn{1}{c}{42.36±11.96}         & \multicolumn{1}{c}{51.67±8.10}          & {\ul 56.42±8.63}     \\ 
\multicolumn{1}{c}{Wiki}                    & \multicolumn{1}{c}{-}          & \multicolumn{1}{c}{29.21±4.32}  & \multicolumn{1}{c}{30.51±0.64}          & \multicolumn{1}{c}{\textbf{39.36±6.58}} & \multicolumn{1}{c}{12.21±3.11}       & \multicolumn{1}{c}{34.32±6.73}          & \multicolumn{1}{c}{34.30±6.77}          & {\ul 36.44±6.81}     \\ 
\multicolumn{1}{c}{Reuters}                 & \multicolumn{1}{c}{-}          & \multicolumn{1}{c}{-}           & \multicolumn{1}{c}{-}                   & \multicolumn{1}{c}{-}                   & \multicolumn{1}{c}{11.50±5.87}       & \multicolumn{1}{c}{{\ul 23.04±4.29}}    & \multicolumn{1}{c}{19.16±4.51}          & \textbf{25.81±2.78}  \\ 
\multicolumn{1}{c}{Caltech256}              & \multicolumn{1}{c}{-}          & \multicolumn{1}{c}{-}           & \multicolumn{1}{c}{-}                   & \multicolumn{1}{c}{-}                   & \multicolumn{1}{c}{25.13±2.45}       & \multicolumn{1}{c}{{\ul 26.80±1.89}} & \multicolumn{1}{c}{26.78±1.98}          & \textbf{26.97±1.97}     \\ 
\multicolumn{1}{c}{VGGFace2}                & \multicolumn{1}{c}{-}          & \multicolumn{1}{c}{-}           & \multicolumn{1}{c}{-}                   & \multicolumn{1}{c}{-}                   & \multicolumn{1}{c}{7.36±0.73}        & \multicolumn{1}{c}{{\ul 10.49±1.65}}    & \multicolumn{1}{c}{10.21±1.36}          & \textbf{10.72±1.67}  \\ 
\multicolumn{1}{c}{YouTubeFace10}           & \multicolumn{1}{c}{-}          & \multicolumn{1}{c}{-}           & \multicolumn{1}{c}{-}                   & \multicolumn{1}{c}{-}                   & \multicolumn{1}{c}{60.83±7.44}       & \multicolumn{1}{c}{56.95±12.20}         & \multicolumn{1}{c}{{\ul 60.89±10.06}}   & \textbf{80.28±5.36}  \\ \hline
\multicolumn{9}{c}{Purity}                                                                                                                                                                                                                                                                                                                                  \\ \hline
\multicolumn{1}{c}{ORL}                     & \multicolumn{1}{c}{31.90±9.40} & \multicolumn{1}{c}{44.18±10.60} & \multicolumn{1}{c}{{\ul 59.23±12.35}}   & \multicolumn{1}{c}{49.76±8.23}          & \multicolumn{1}{c}{54.99±8.29}       & \multicolumn{1}{c}{55.73±9.86}          & \multicolumn{1}{c}{55.37±9.87}          & \textbf{69.08±7.22}  \\ 
\multicolumn{1}{c}{proteinFold}             & \multicolumn{1}{c}{22.81±0.91} & \multicolumn{1}{c}{14.84±0.00}  & \multicolumn{1}{c}{27.70±1.38}          & \multicolumn{1}{c}{{\ul 33.22±1.92}}    & \multicolumn{1}{c}{24.12±2.78}       & \multicolumn{1}{c}{32.42±3.56}          & \multicolumn{1}{c}{32.52±3.59}          & \textbf{36.45±3.06}  \\ 
\multicolumn{1}{c}{uci-digit}               & \multicolumn{1}{c}{38.24±7.46} & \multicolumn{1}{c}{10.45±0.00}  & \multicolumn{1}{c}{{\ul 70.26±1.80}}    & \multicolumn{1}{c}{48.98±7.06}          & \multicolumn{1}{c}{49.11±5.88}       & \multicolumn{1}{c}{59.19±12.28}         & \multicolumn{1}{c}{59.45±11.17}         & \textbf{72.94±6.89}  \\ 
\multicolumn{1}{c}{Wiki}                    & \multicolumn{1}{c}{-}          & \multicolumn{1}{c}{44.71±4.72}  & \multicolumn{1}{c}{42.21±1.37}          & \multicolumn{1}{c}{{\ul 47.21±7.17}}    & \multicolumn{1}{c}{31.16±4.03}       & \multicolumn{1}{c}{46.05±6.19}          & \multicolumn{1}{c}{46.08±6.25}          & \textbf{48.07±6.20}  \\ 
\multicolumn{1}{c}{Reuters}                 & \multicolumn{1}{c}{-}          & \multicolumn{1}{c}{-}           & \multicolumn{1}{c}{-}                   & \multicolumn{1}{c}{-}                   & \multicolumn{1}{c}{39.78±6.63}       & \multicolumn{1}{c}{{\ul 54.38±2.85}}    & \multicolumn{1}{c}{48.98±5.19}          & \textbf{56.63±1.74}  \\ 
\multicolumn{1}{c}{Caltech256}              & \multicolumn{1}{c}{-}          & \multicolumn{1}{c}{-}           & \multicolumn{1}{c}{-}                   & \multicolumn{1}{c}{-}                   & \multicolumn{1}{c}{10.34±1.47}       & \multicolumn{1}{c}{11.71±1.66}          & \multicolumn{1}{c}{\textbf{11.94±1.60}} & {\ul 11.93±1.72}     \\ 
\multicolumn{1}{c}{VGGFace2}                & \multicolumn{1}{c}{-}          & \multicolumn{1}{c}{-}           & \multicolumn{1}{c}{-}                   & \multicolumn{1}{c}{-}                   & \multicolumn{1}{c}{4.50±0.37}        & \multicolumn{1}{c}{{\ul 6.39±0.90}}     & \multicolumn{1}{c}{5.70±0.80}           & \textbf{6.50±0.90}   \\ 
\multicolumn{1}{c}{YouTubeFace10}           & \multicolumn{1}{c}{-}          & \multicolumn{1}{c}{-}           & \multicolumn{1}{c}{-}                   & \multicolumn{1}{c}{-}                   & \multicolumn{1}{c}{64.60±6.90}       & \multicolumn{1}{c}{{\ul 69.09±11.26}}   & \multicolumn{1}{c}{67.03±11.27}         & \textbf{89.33±2.94}  \\ \hline
\multicolumn{9}{c}{F-score}                                                                                                                                                                                                                                                                                                                                 \\ \hline
\multicolumn{1}{c}{ORL}                     & \multicolumn{1}{c}{16.25±6.31} & \multicolumn{1}{c}{19.01±9.25}  & \multicolumn{1}{c}{{\ul 40.37±14.99}}   & \multicolumn{1}{c}{28.75±9.61}          & \multicolumn{1}{c}{33.43±12.53}      & \multicolumn{1}{c}{36.48±11.96}         & \multicolumn{1}{c}{29.96±13.53}         & \textbf{53.38±10.42} \\ 
\multicolumn{1}{c}{proteinFold}             & \multicolumn{1}{c}{11.40±1.43} & \multicolumn{1}{c}{10.02±0.00}  & \multicolumn{1}{c}{11.69±0.55}          & \multicolumn{1}{c}{{\ul 15.92± 1.54}}   & \multicolumn{1}{c}{9.79±1. 87}       & \multicolumn{1}{c}{14.85±2.90}          & \multicolumn{1}{c}{14.18±3.21}          & \textbf{17.58±2.87}  \\ 
\multicolumn{1}{c}{uci-digit}               & \multicolumn{1}{c}{32.25±3.89} & \multicolumn{1}{c}{18.10±0.00}  & \multicolumn{1}{c}{{\ul 55.04±3.33}}    & \multicolumn{1}{c}{35.05±7.16}          & \multicolumn{1}{c}{35.42±6.96}       & \multicolumn{1}{c}{40.75±12.51}         & \multicolumn{1}{c}{40.00±12.19}         & \textbf{55.68±8.87}  \\ 
\multicolumn{1}{c}{Wiki}                    & \multicolumn{1}{c}{-}          & \multicolumn{1}{c}{28.79±4.82}  & \multicolumn{1}{c}{26.38±2.33}          & \multicolumn{1}{c}{30.35±8.53}          & \multicolumn{1}{c}{18.84±2.18}       & \multicolumn{1}{c}{31.08±7.83}          & \multicolumn{1}{c}{{\ul 31.24±7.87}}    & \textbf{33.21±7.87}  \\ 
\multicolumn{1}{c}{Reuters}                 & \multicolumn{1}{c}{-}          & \multicolumn{1}{c}{-}           & \multicolumn{1}{c}{-}                   & \multicolumn{1}{c}{-}                   & \multicolumn{1}{c}{38.97±1.98}       & \multicolumn{1}{c}{{\ul 39.00±3.95}}    & \multicolumn{1}{c}{35.17±4.21}          & \textbf{41.56±2.57}  \\ 
\multicolumn{1}{c}{Caltech256}              & \multicolumn{1}{c}{-}          & \multicolumn{1}{c}{-}           & \multicolumn{1}{c}{-}                   & \multicolumn{1}{c}{-}                   & \multicolumn{1}{c}{3.59±2.02}        & \multicolumn{1}{c}{{\ul 5.36±1.46}}     & \multicolumn{1}{c}{5.01±1.73}           & \textbf{5.64±1.50}   \\ 
\multicolumn{1}{c}{VGGFace2}                & \multicolumn{1}{c}{-}          & \multicolumn{1}{c}{-}           & \multicolumn{1}{c}{-}                   & \multicolumn{1}{c}{-}                   & \multicolumn{1}{c}{1.83±0.08}        & \multicolumn{1}{c}{{\ul 2.24±0.35}}     & \multicolumn{1}{c}{1.34±0.31}           & \textbf{2.29±0.36}   \\ 
\multicolumn{1}{c}{YouTubeFace10}           & \multicolumn{1}{c}{-}          & \multicolumn{1}{c}{-}           & \multicolumn{1}{c}{-}                   & \multicolumn{1}{c}{-}                   & \multicolumn{1}{c}{{\ul 54.63±9.51}} & \multicolumn{1}{c}{53.93±14.04}         & \multicolumn{1}{c}{48.80±15.65}         & \textbf{82.16±14.83} \\ \bottomrule
\end{tabular}
\end{table*}
\subsubsection{Clustering Performance Comparison on Complete Datasets} 
Table \ref{comlete_result} demonstrates the results of FCMVC-IV and the competitors on eight complete datasets based on the four clustering metrics mentioned above. Note that '\--{}' indicates the algorithm cannot be executed smoothly due to the out-of-memory error. From this table, we have the following observations:
\begin{enumerate}
\item As a method of MF-based MVC, FCMVC can handle situations where the number of views is not fixed and can grow over time. Furthermore, FCMVC significantly improves over MFMVC and achieves the best clustering performance among the compared algorithms. For instance,  it exceeds the second-best algorithm by 3.41\%, 3.38\%, 2.77\%, 1.42\%, 3.48\%, 1.08\%, 22.12\%, and 25.29\% in terms of ACC on all the datasets, and the improvements on other metrics are similar, illustrating that FCMVC can effectively combine the information between different views.
\item Compared with FMR, PMSC,  MNMF, LMVSC, FMCNOF, 
and CMVC, FCMVC is parameter-free. These methods consume considerable resources to select suitable hyperparameters by conducting the algorithm repeatedly in practical applications. In contrast, our proposed method is free of parameters and more suitable for large-scale data. Meanwhile, the proposed method is superior to the compared ones, demonstrating its effectiveness.
\item Different from these compared methods, the proposed algorithm works on both regular and large-scale datasets. Also, FCMVC-IV can handle incomplete views. Thus it has a broader range of application scenarios.
\end{enumerate}
\begin{figure*}[htbp]
	\centering
	\vspace{-0.4cm}
	\subfigure{
		\includegraphics[width=0.8\textwidth]{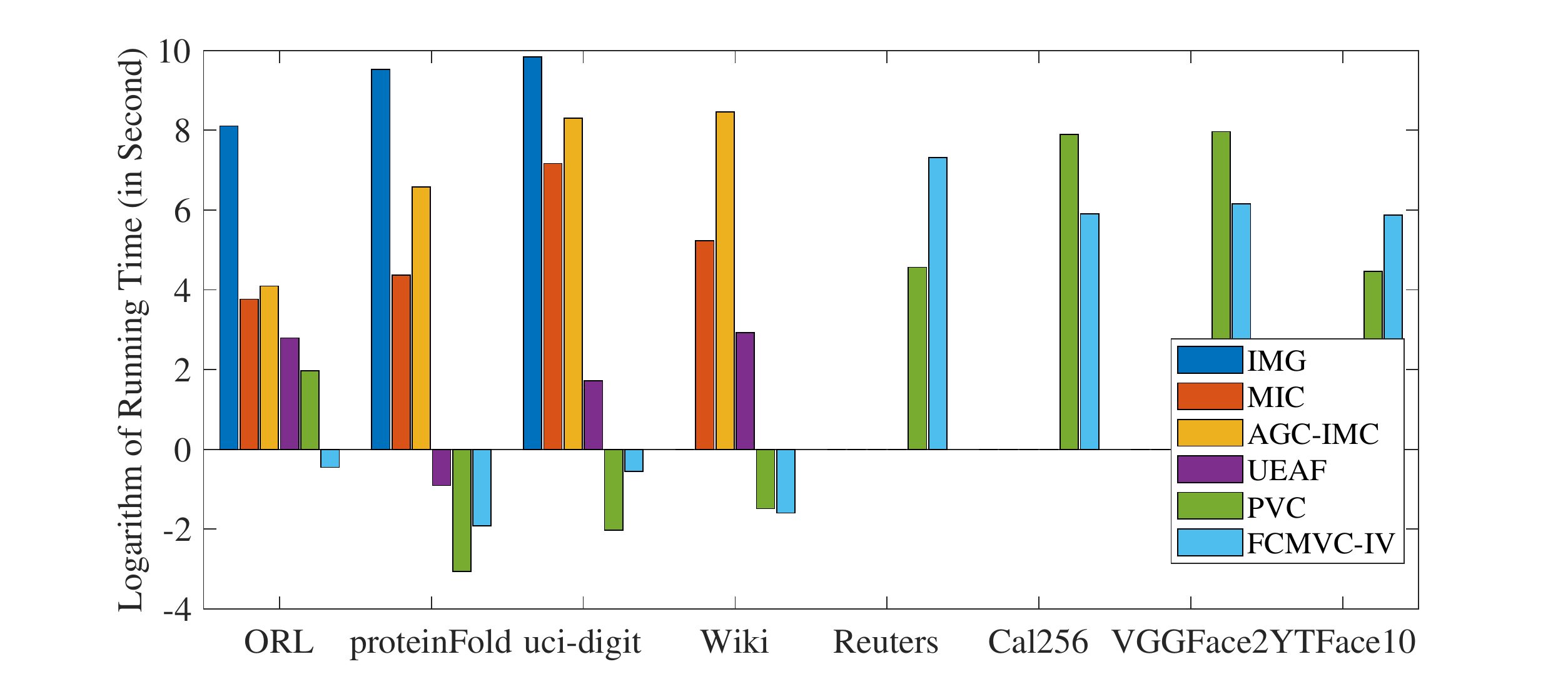}}
	\hspace{-0.3cm}

	\caption{The relative running time of the compared algorithms on the incomplete benchmark datasets. The empty bar indicates that the respective method is out of memory on the dataset.}
	\label{running_time}
\end{figure*}

\begin{figure*}[h]
	\centering
	\vspace{-0.2cm}
        \subfigure{
		\includegraphics[width=0.24\textwidth]{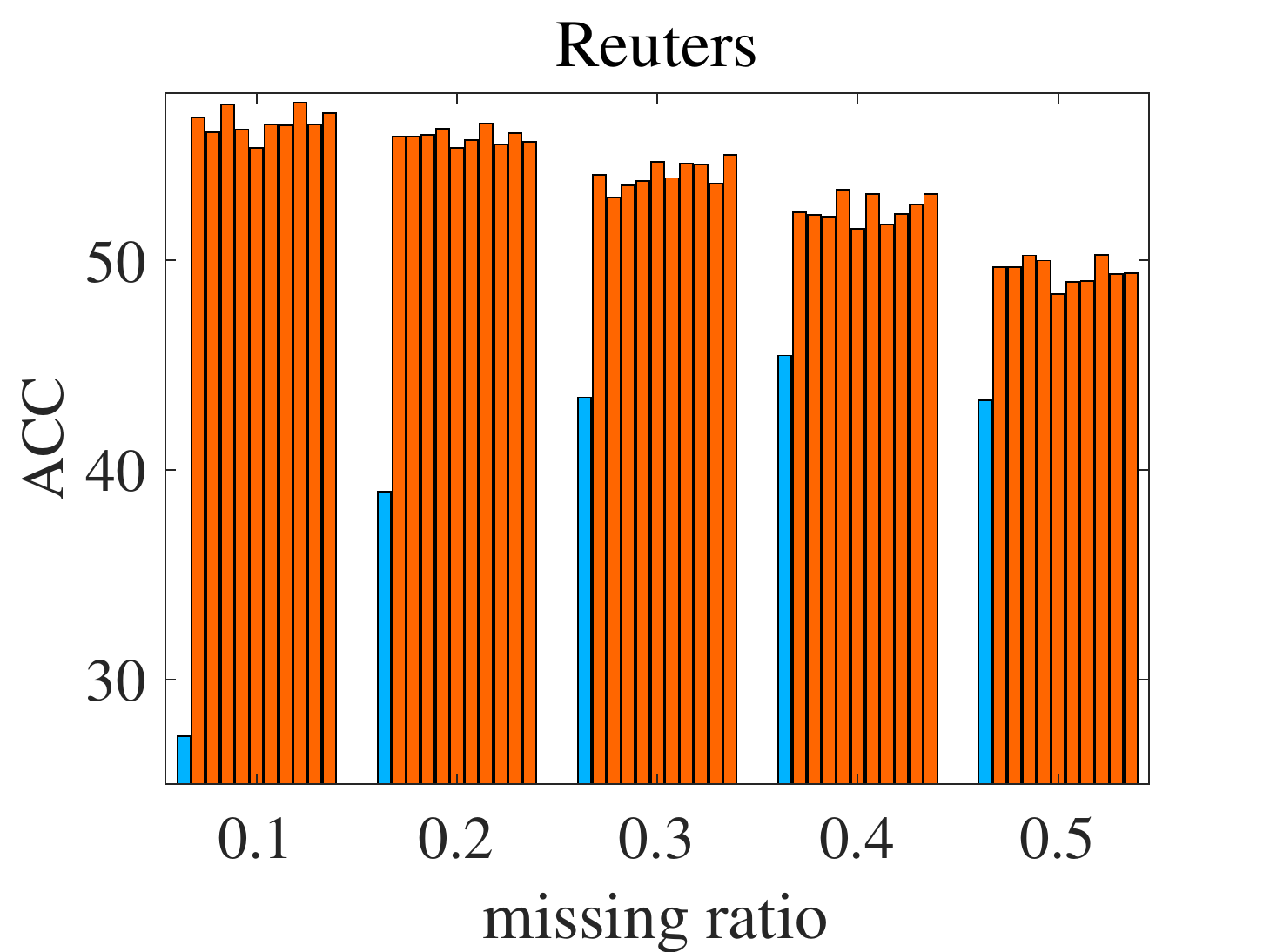}}
	\hspace{-0.2cm}
	\subfigure{
		\includegraphics[width=0.24\textwidth]{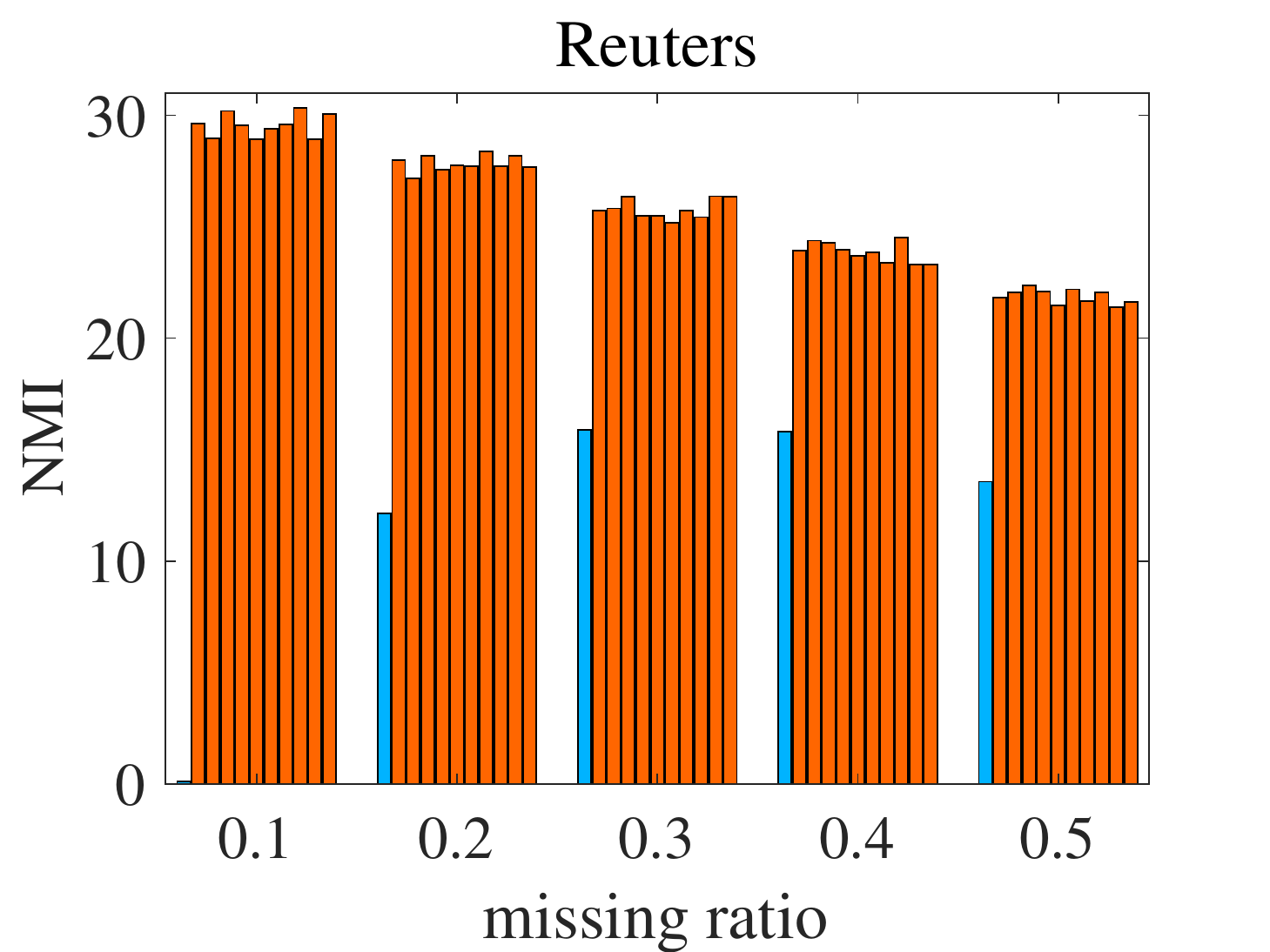}}
	\hspace{-0.2cm}
	\subfigure{
		\includegraphics[width=0.24\textwidth]{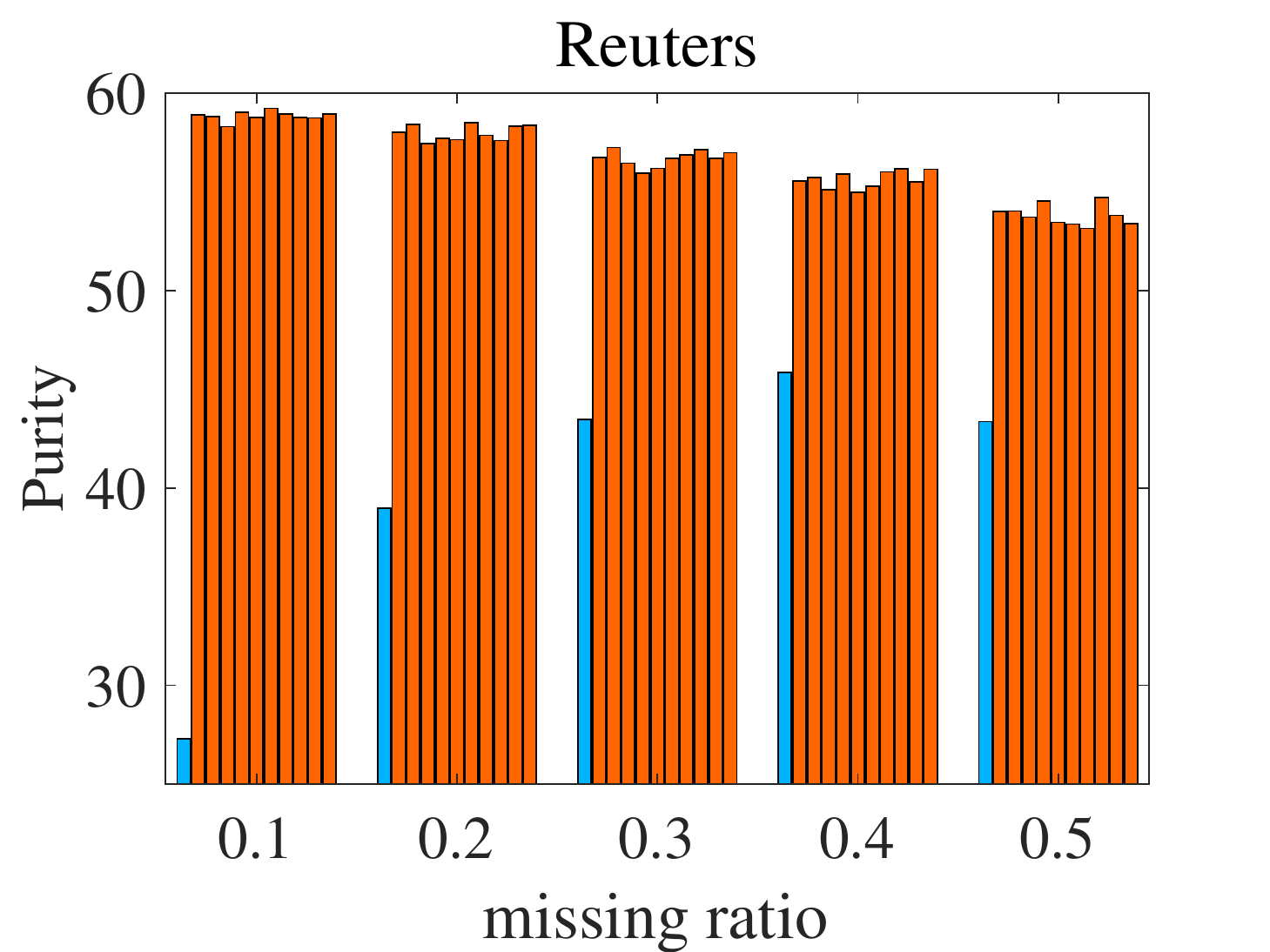}}
	\hspace{-0.2cm}
	\subfigure{
		\includegraphics[width=0.24\textwidth]{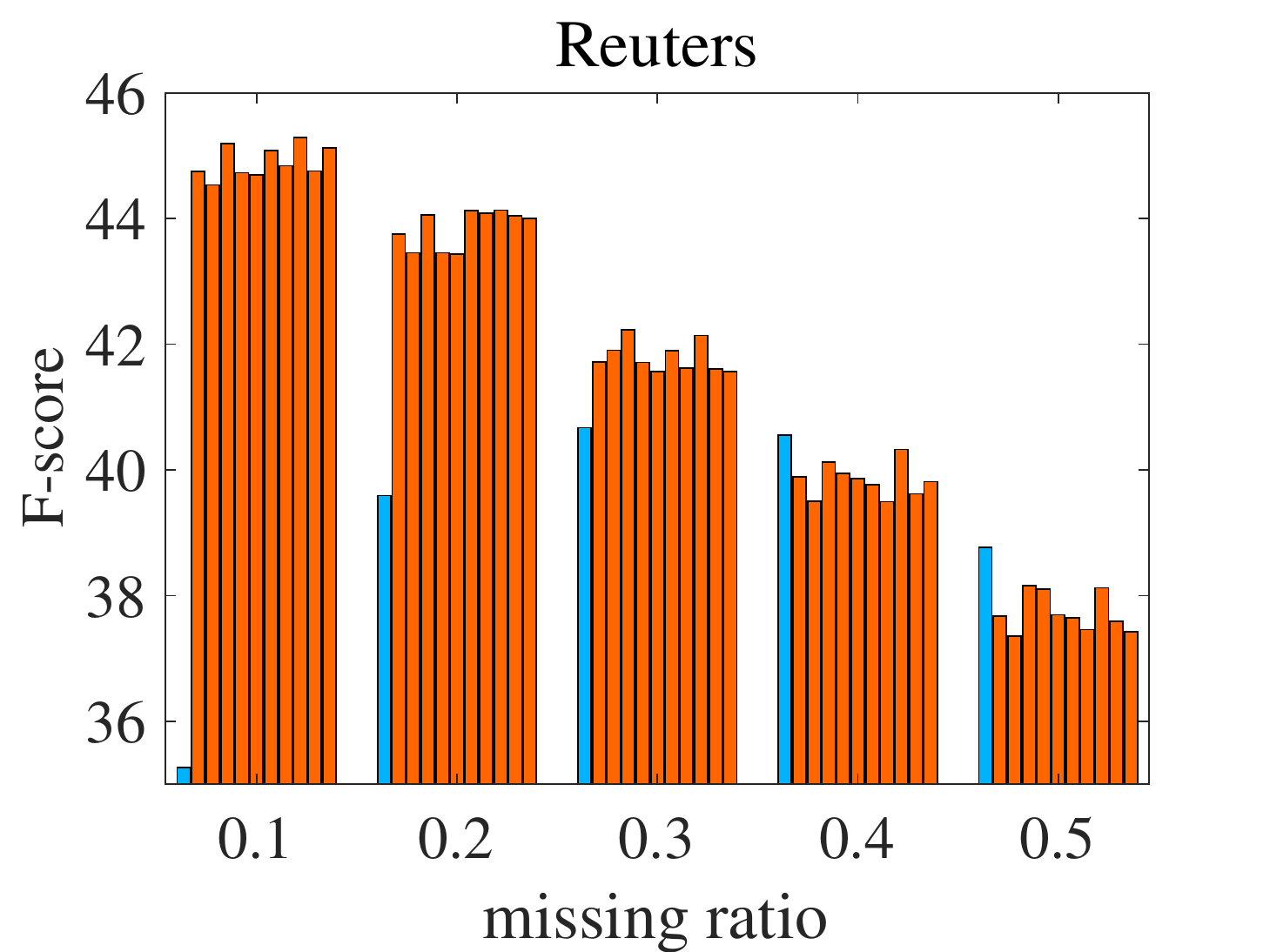}}
	\hspace{-0.2cm}
	\subfigure{
		\includegraphics[width=0.24\textwidth]{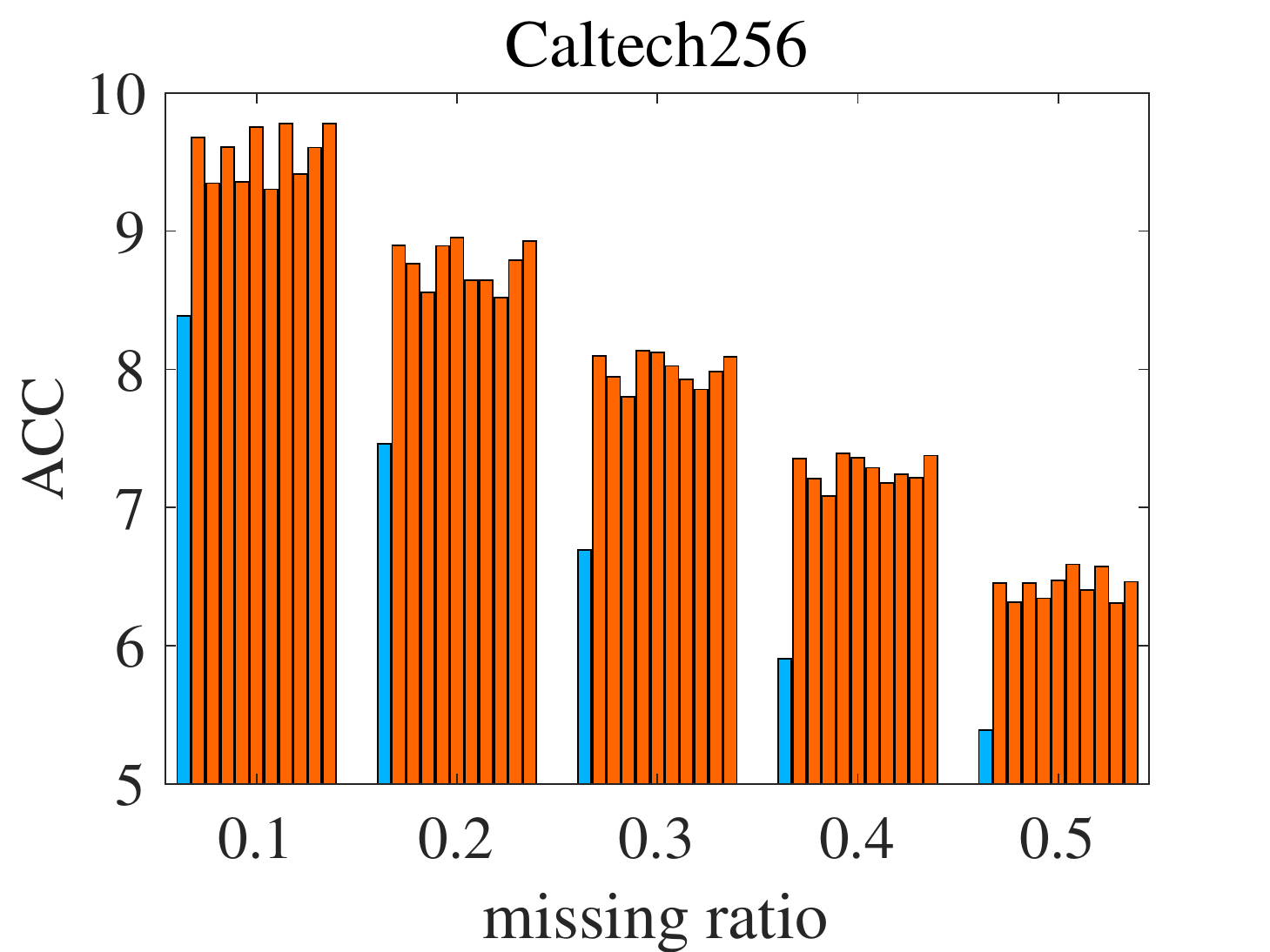}}
	\hspace{-0.2cm}
	\subfigure{
		\includegraphics[width=0.24\textwidth]{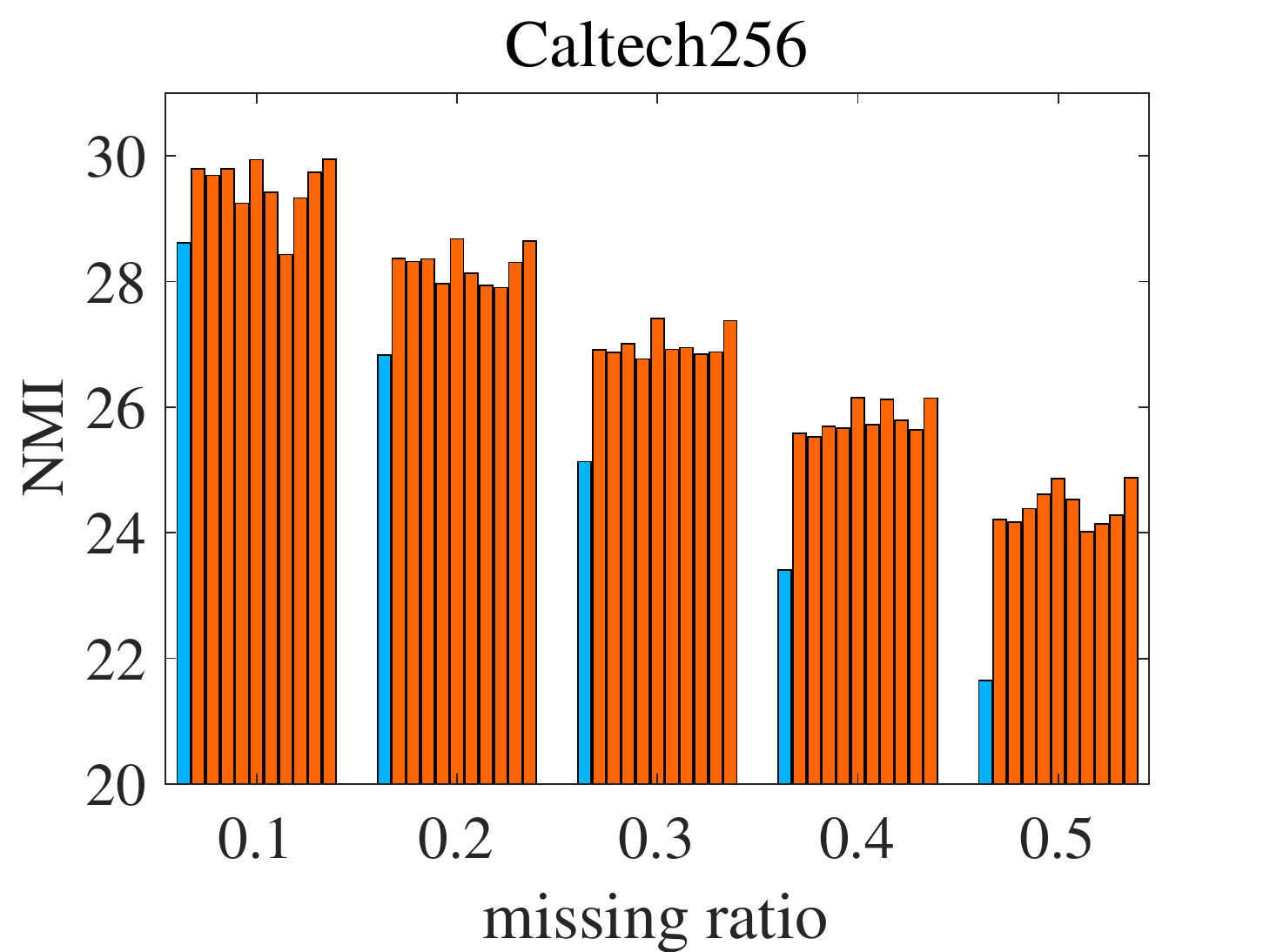}}
	\hspace{-0.2cm}
	\subfigure{
		\includegraphics[width=0.24\textwidth]{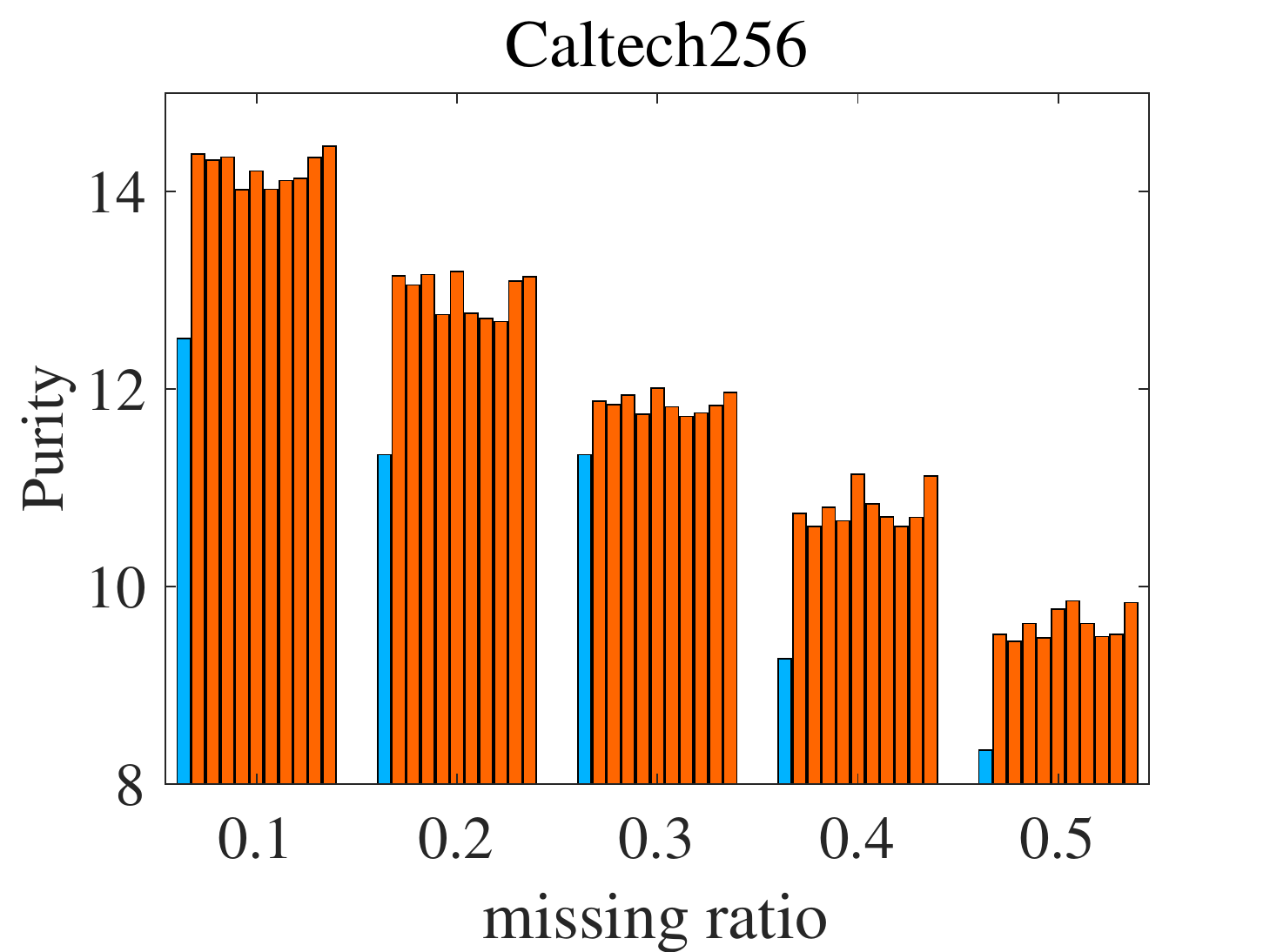}}
	\hspace{-0.2cm}
	\subfigure{
		\includegraphics[width=0.24\textwidth]{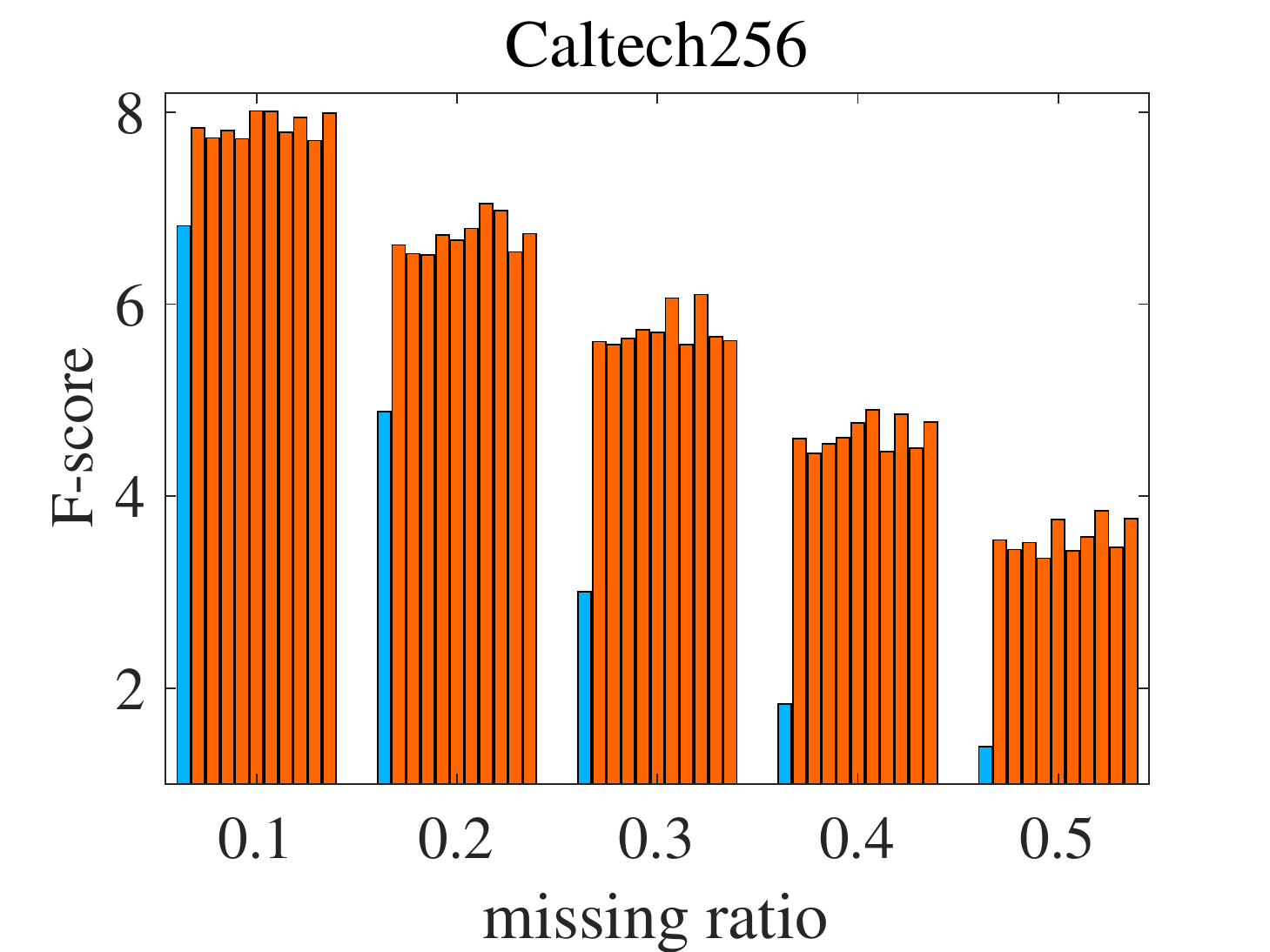}}
	\hspace{-0.2cm}
	\subfigure{
		\includegraphics[width=0.24\textwidth]{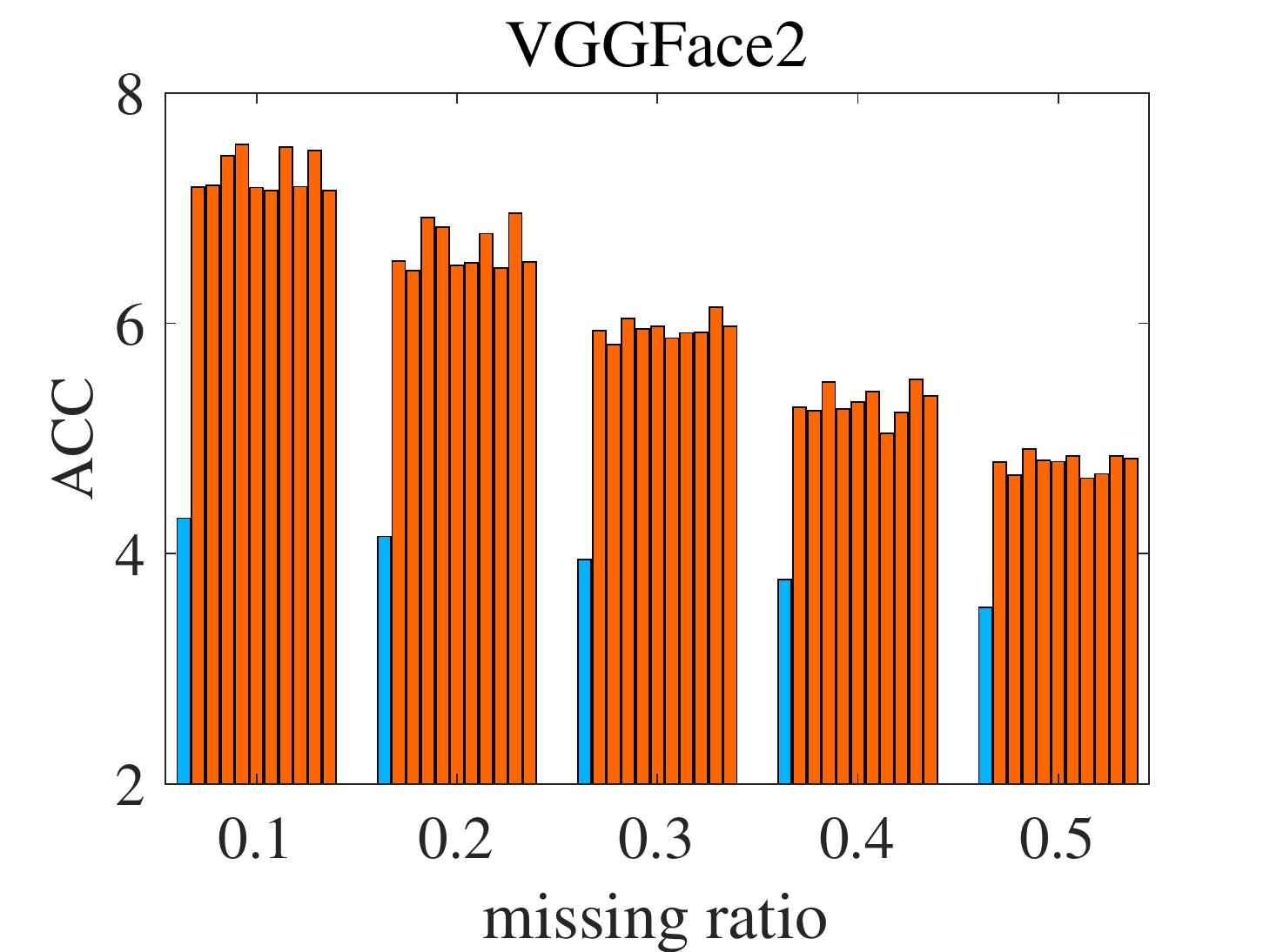}}
	\hspace{-0.2cm}
	\subfigure{
		\includegraphics[width=0.24\textwidth]{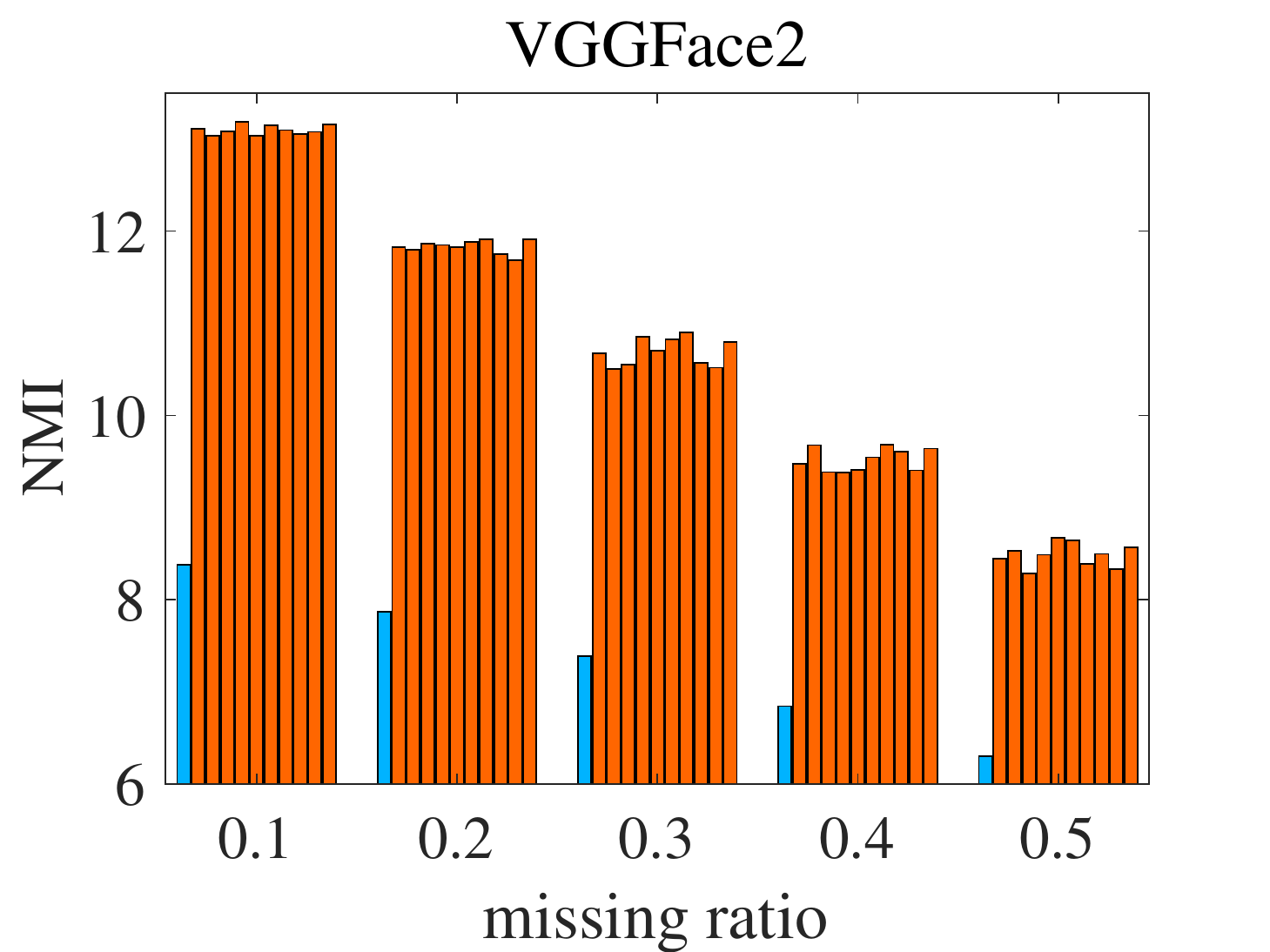}}
	\hspace{-0.2cm}
	\subfigure{
		\includegraphics[width=0.24\textwidth]{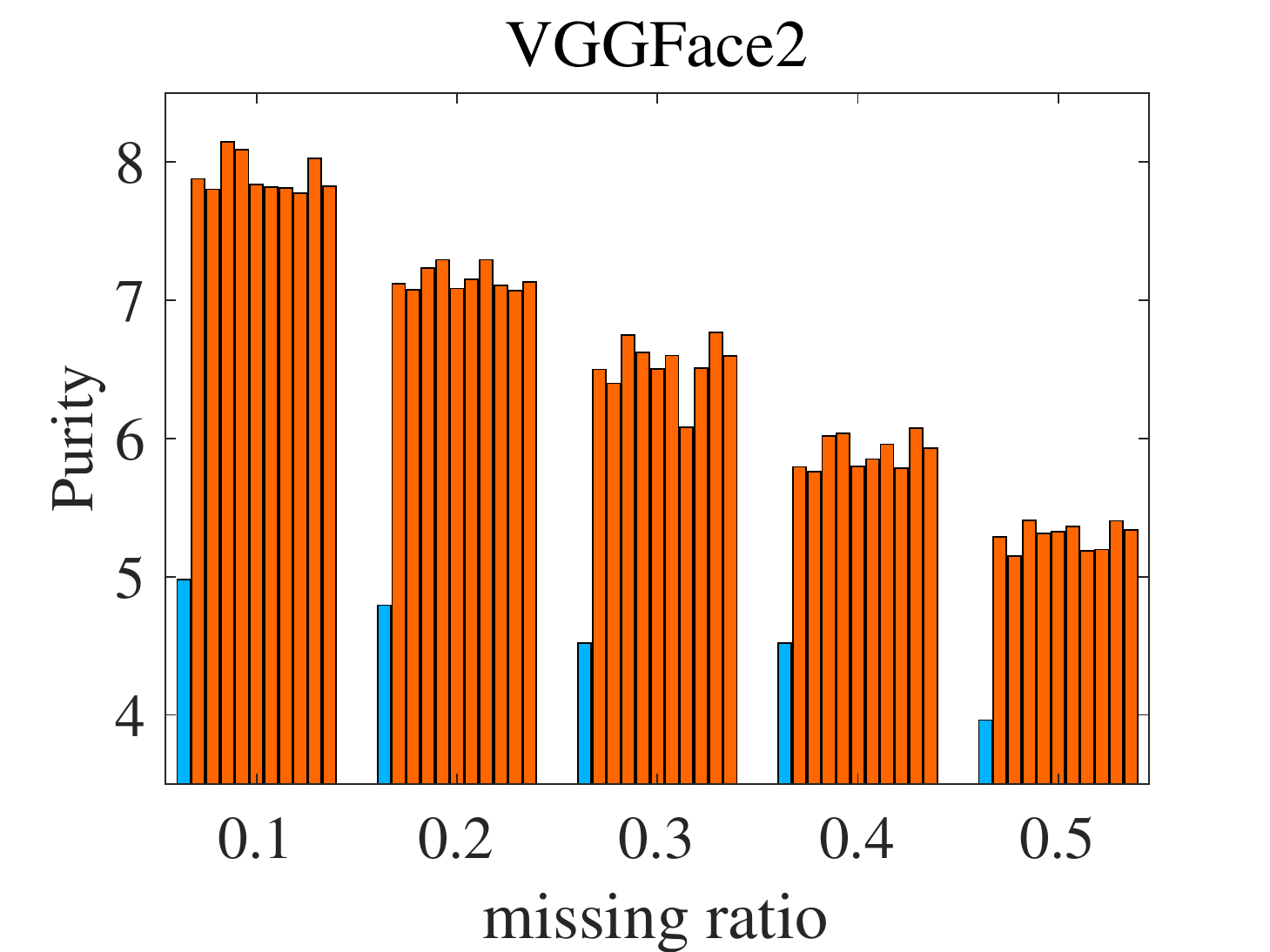}}
	\hspace{-0.2cm}
	\subfigure{
		\includegraphics[width=0.24\textwidth]{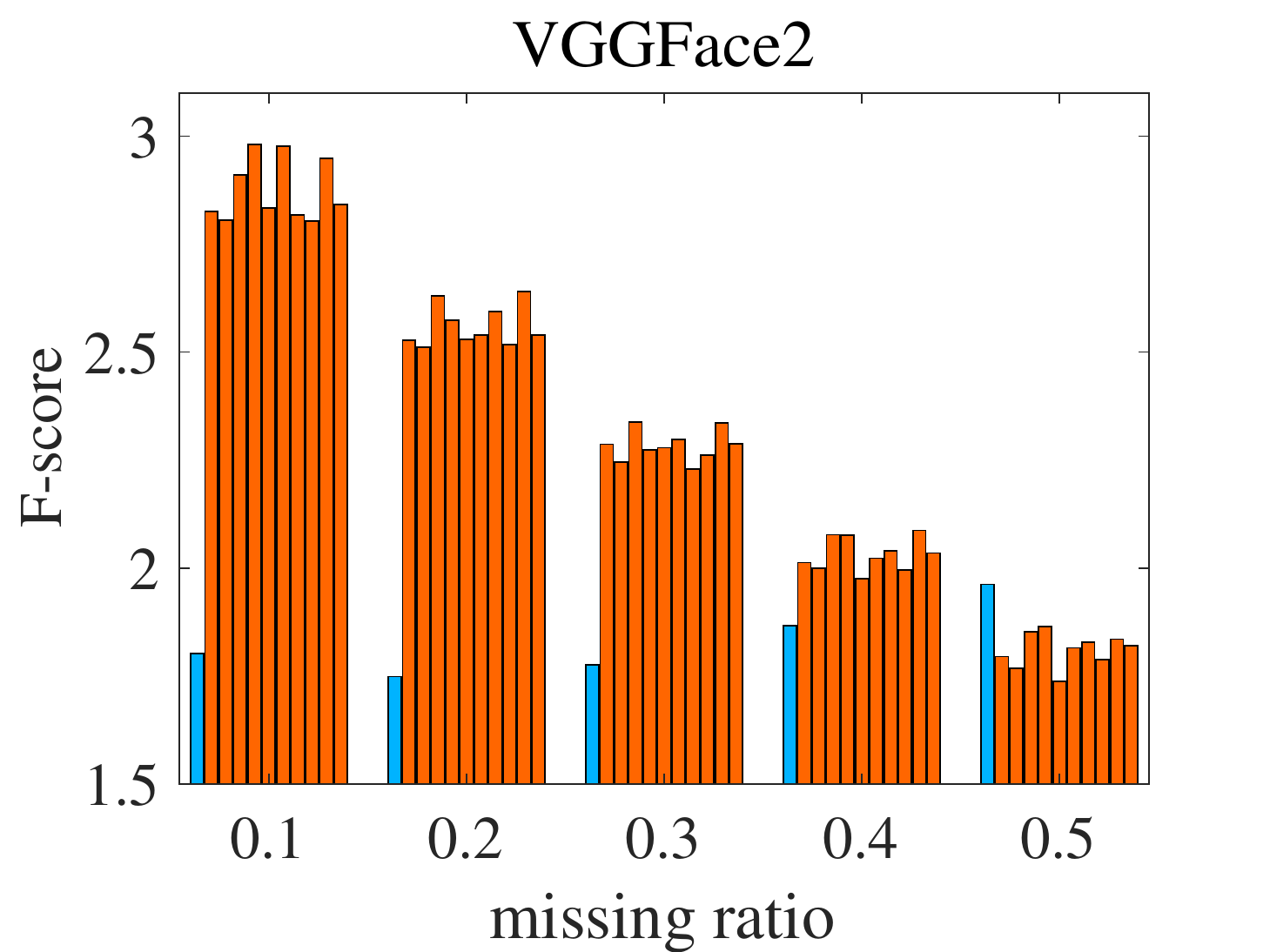}}
	\hspace{-0.2cm}
		\subfigure{
		\includegraphics[width=0.24\textwidth]{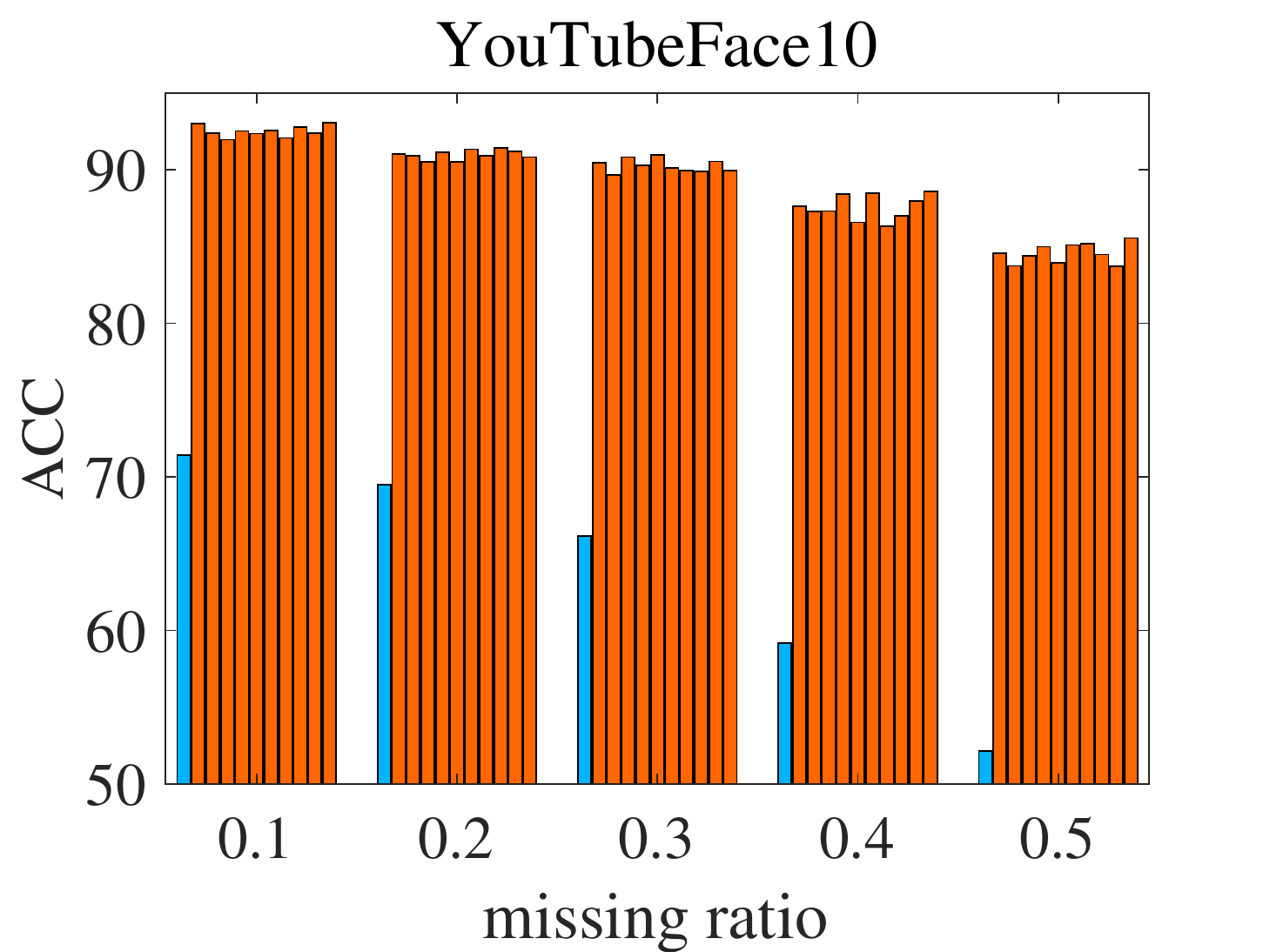}}
	\hspace{-0.2cm}
	\subfigure{
		\includegraphics[width=0.24\textwidth]{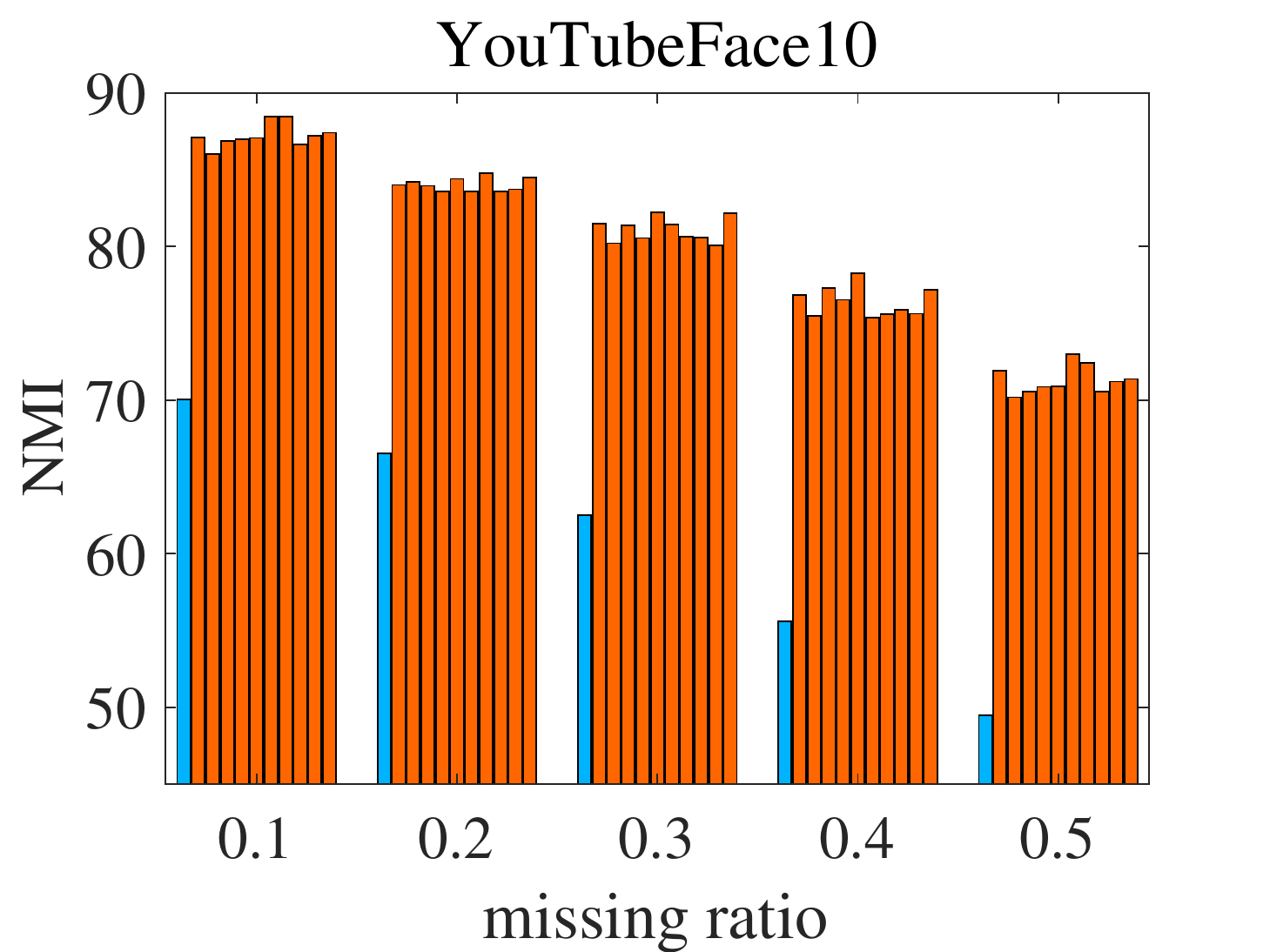}}
	\hspace{-0.2cm}
	\subfigure{
		\includegraphics[width=0.24\textwidth]{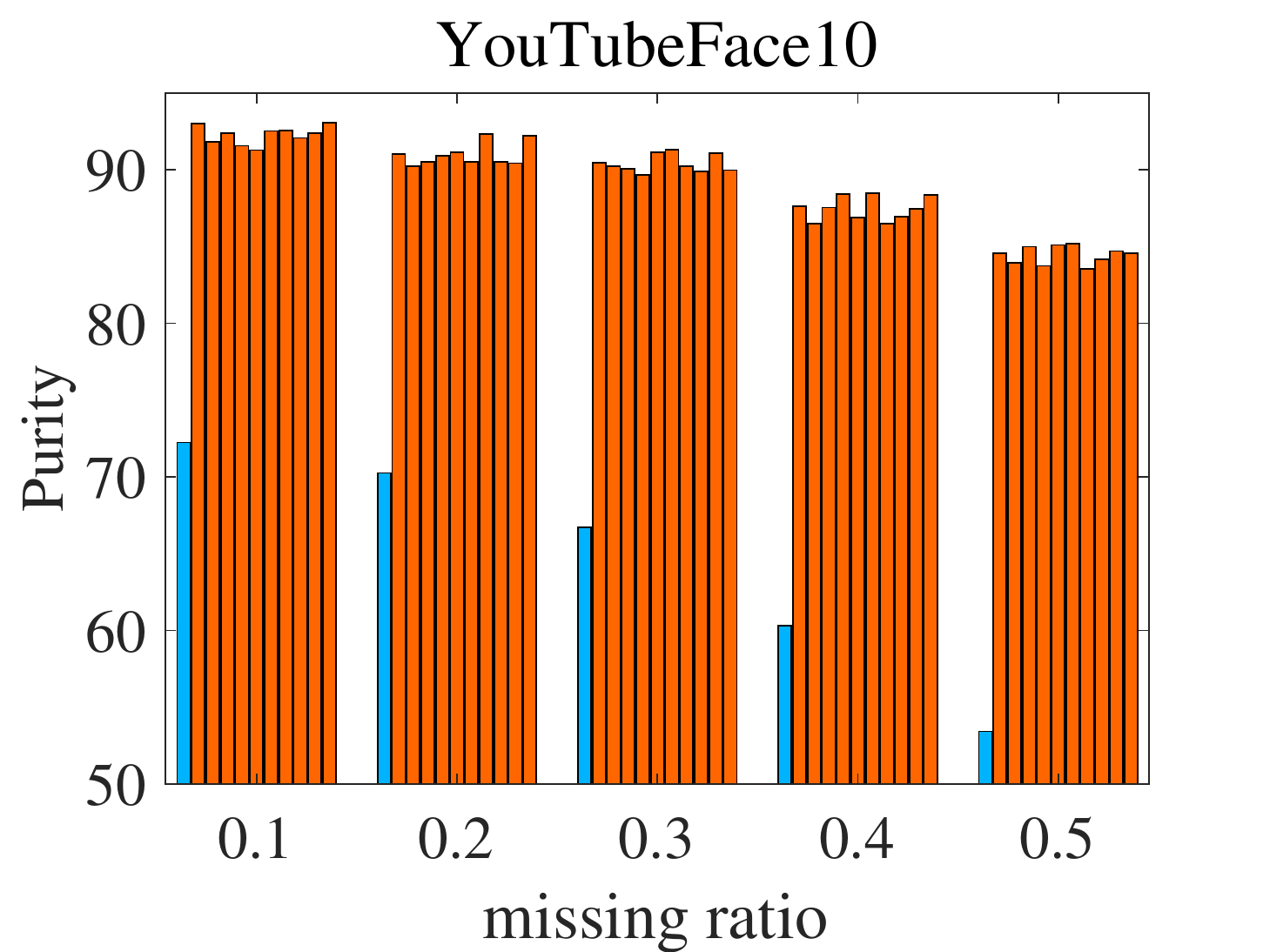}}
	\hspace{-0.2cm}
	\subfigure{
		\includegraphics[width=0.24\textwidth]{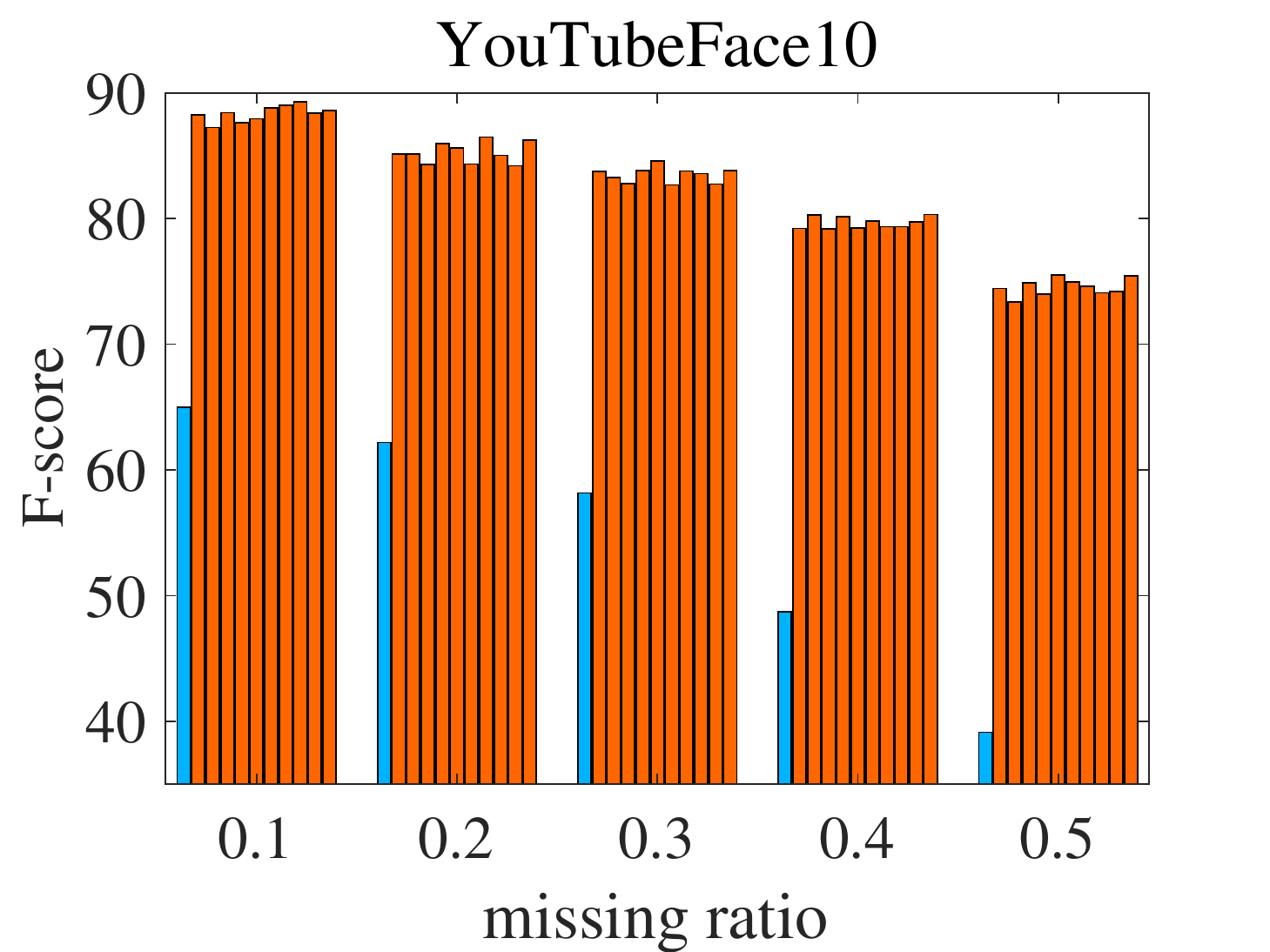}}
	\hspace{-0.2cm}
	
	\caption{The results of FCMVC-IV vary with different fusing orders of views compared with PVC on four large-scale datasets.}
	\label{fig_order}
\end{figure*}

\subsubsection{Clustering Performance Comparison on Incomplete Datasets}
We depict the clustering results vary with missing ratios of each dataset in Fig \ref{fig_incom_res1} and \ref{fig_incom_res2}. To further clarify the excellence of our proposed method, we demonstrate the aggregated results of FCMVC-IV and the comparison algorithms on eight incomplete datasets in Table \ref{incomlete_result}. For each dataset, we report the mean and standard deviation (std) concerning the results of different missing rates. Based on the figure and table, we observe that:
\begin{enumerate}
\item Our proposed method exhibits superiority on all clustering indicators compared to incomplete MVC algorithms. For instance, our approach improves  17.31\%, 13.09\%, 4.16\%, 1.32\%, 5.08\%, 2.02\%, 1.89\%, and 31.23\% over the second-best method on all datasets regarding ACC, separately. Meanwhile, FCMVC-IV almost consistently outperforms the compared methods on these datasets on different missing ratios. Therefore, FCMVC-IV is a unique pioneering algorithm of incomplete continual MVC.
\item Caused by the square or cubic time complexity, most state-of-the-art incomplete algorithms fail to handle large-scale data. For example, IMG, MIC, AGC-IMC, and UEAF suffer from out-of-memory errors when dealing with Reuters, Caltech256, VGGFace2, and YouTubeFace10. Moreover, the intractable hyper-parameters further limit their applications. Our proposed algorithm addresses the above drawbacks owing to its linear complexity and parameter-free property.
\item Compared with other competitors, our naive framework FPMVC+ZF and FPMVC+AF demonstrate a comparable clustering performance, illustrating the effectiveness of our fusion strategy. However, they are inferior to FPMVC-IV. Thus, the attempt to integrate the knowledge between the incomplete previous and the incoming views is valid. Also, the strategy of merely utilizing appearing samples is effective.
\end{enumerate}

\subsection{Running Time Comparison}
We conduct experiments to evaluate the time efficiency of our proposed algorithm, and the results are shown in Figure \ref{running_time}. Considering that FCMVC+ZF and FCMVC+AF share a similar framework to FCMVC-IV and the difference in running time between them is slight, we plot the other five competitors in our figure. Note that the running time of FCMVC-IV is obtained from the arrival of the first view until the fusion of the last view. From the figure, it is obtained that FCMVC-IV consumes fewer time resources than the compared methods. In addition, FCMVC-IV is more flexible because it can deal with continual multi-view data, while the compared algorithms have to recompute with all data once a new view arrives. Although PVC costs less time on several datasets, it needs to repeat the implementations when a new view is available. Therefore, the real-time processing capacity makes FCMVC-IV superior to PVC in terms of time efficiency.

In sum, both the theoretical and the experimental results show the low time complexity of our method, enabling it to handle real-time large-scale incomplete data. Moreover, the clustering performance indicates its superiority to existing algorithms.

\subsection{Experiment on the effect of view order}
Considering that the algorithm deals with scenarios where the number of views can increase, we conduct experiments to investigate the effect of different fusion orders on the clustering results. For instance, suppose there are three views $\left\{V_1, V_2, V_3\right\}$ in total. If $V_1$, $V_2$, and $V_3$  are fused in turn, the final experimental results may differ from other fusion sequences, such as the fusing order of $V_3$, $V_2$ and $V_1$. Suppose that $m$ denotes the number of views, there are $m!=m*(m-1)*\cdots*1$ fusing orders. Figure \ref{fig_order} reports the results of FCMVC-IV compared with PVC with respect to ten different orders on three large-scale datasets. From the figure, it is obtained that our proposed method is not sensitive to the fusing order.

\section{Conclusion}
In this paper, we propose a novel algorithm termed Fast Continual Multi-View Clustering with Incomplete Views (FCMVC-IV) to handle the incomplete continual data problem (ICDP). By maintaining a consensus coefficient matrix, FCMVC-IV updates it when a new view is available. We match the consensus information and the new knowledge of different sample sizes with two matrices, i.e., the indicator matrix and the rotation matrix. In addition, comprehensive experiments demonstrate the excellent efficiency of the proposed method. Meanwhile, FCMVC-IV is not sensitive to fusing orders, enabling it to handle real-time data with stable results.

\ifCLASSOPTIONcompsoc
  \section*{Acknowledgments}
\else
  \section*{Acknowledgment}
\fi
This work was supported by the National Key R\&D Program of China 2020AAA0107100, the Natural Science Foundation of China (project no. 61922088, 61773392, 62006237 and 61976196).

\ifCLASSOPTIONcaptionsoff
  \newpage
\fi



%
\bibliographystyle{IEEEtran}
\bibliography{myreference}
%


\begin{IEEEbiography}[{\includegraphics[width=1in,height=1.10in,clip,keepaspectratio]{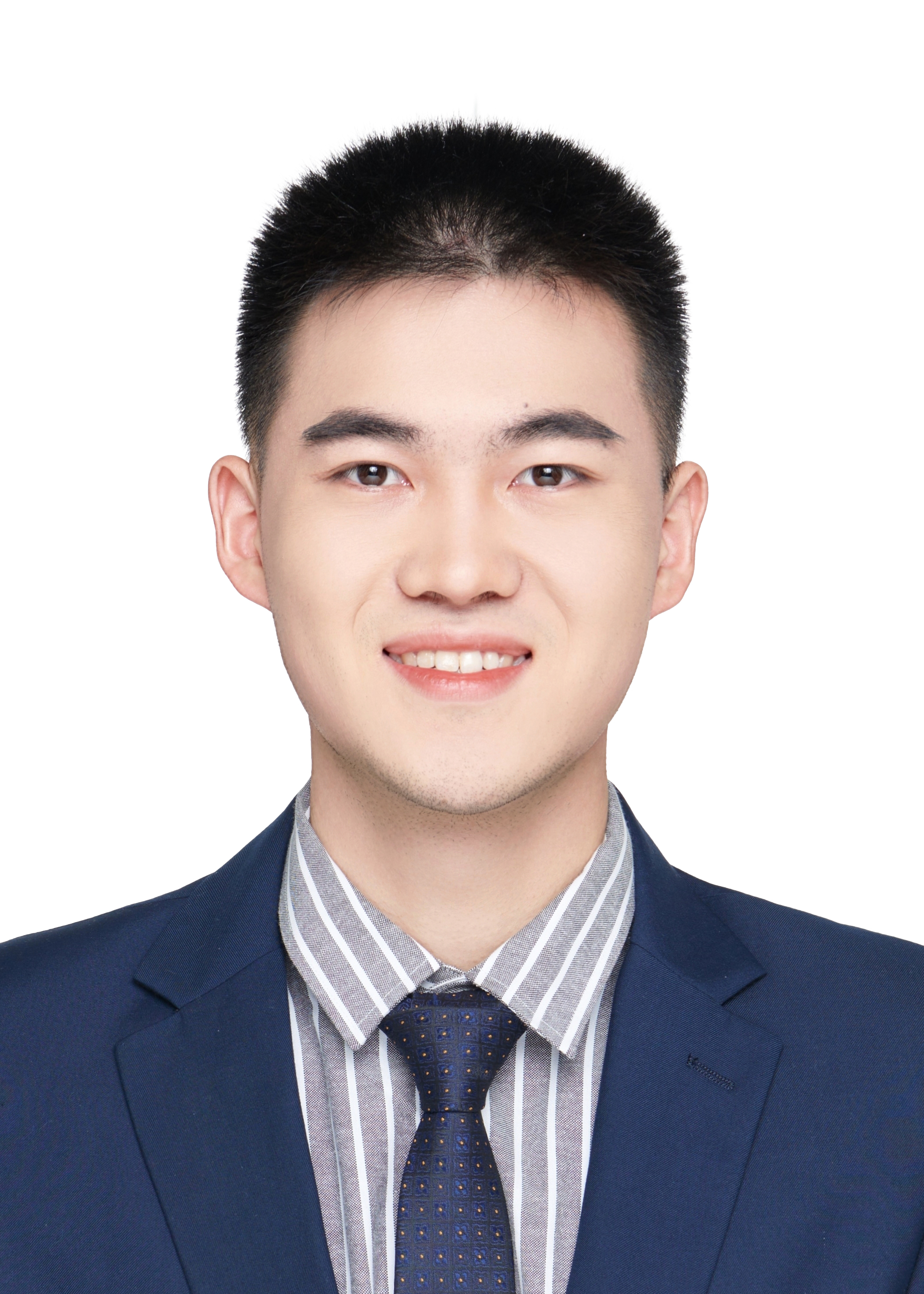}}]{Xinhang Wan} received the B.E degree in Computer Science and Technology from Northeastern University, Shenyang, China, in 2021. He is currently working toward the Ph.D. degree with the National University of Defense Technology (NUDT), Changsha, China.
He has published papers in conferences, such as ACM MM and AAAI. His current research interests include multi-view learning, incomplete multi-view clustering, and continual clustering.
\end{IEEEbiography}

\begin{IEEEbiography}[{\includegraphics[width=1in,height=1.25in,clip,keepaspectratio]{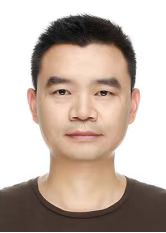}}]
{Bin Xiao} received the B.S. and M.S. degrees in electrical engineering from Shanxi Normal University, Xi'an, China, in 2004 and 2007, respectively, and the Ph.D. degree in computer science from Xidian University, Xi'an. He is currently a Professor with the Chongqing University of Posts and Telecommunications, Chongqing, China. His research interests include image processing and pattern recognition. 
\end{IEEEbiography}

\begin{IEEEbiography}[{\includegraphics[width=1in,height=1.10in,clip,keepaspectratio]{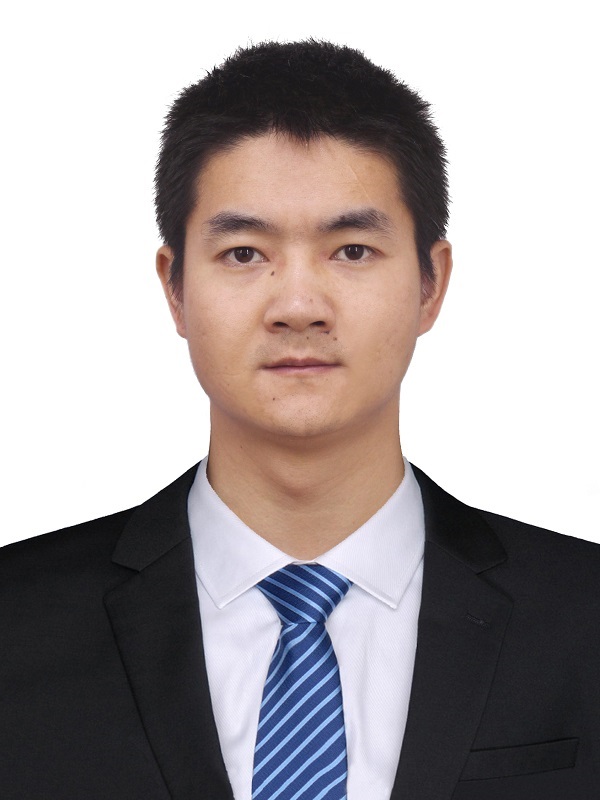}}]{Xinwang Liu} received his PhD degree from National University of Defense Technology (NUDT), China. He is now Professor of School of Computer, NUDT. His current research interests include kernel learning and unsupervised feature learning. Dr. Liu has published 60+ peer-reviewed papers, including those in highly regarded journals and conferences such as IEEE T-PAMI, IEEE T-KDE, IEEE T-IP, IEEE T-NNLS, IEEE T-MM, IEEE T-IFS, ICML, NeurIPS, ICCV, CVPR, AAAI, IJCAI, etc. He serves as the associated editor of Information Fusion Journal. More information can be found at \url{https://xinwangliu.github.io/}.
\end{IEEEbiography}

\begin{IEEEbiography}[{\includegraphics[width=1in,height=1.10in,clip,keepaspectratio]{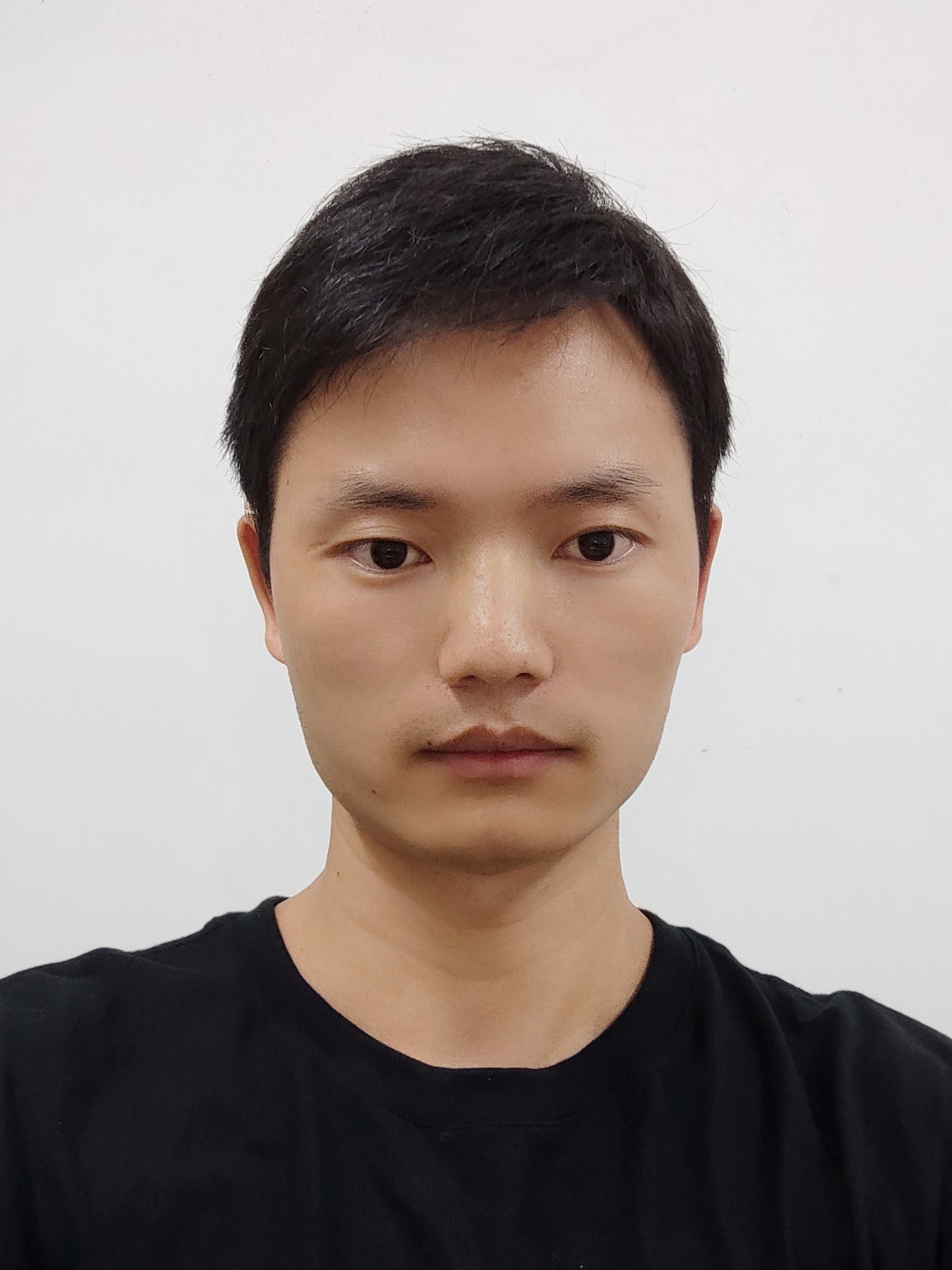}}]{Jiyuan Liu} received his PhD from National University of Defense Technology (NUDT), China, in 2022. He is now a lecturer with the College of Systems Engineering, NUDT. His current research interests include multi-view clustering, federated learning and anomaly detection. Dr. Liu has published papers in journals and conferences such as IEEE T-KDE, IEEE T-NNLS, ICML, NeurIPS, CVPR, ICCV, ACMMM, AAAI, IJCAI, etc. He serves as program committee member and reviewer on IEEE T-KDE, IEEE T-NNLS, ICML, NeurIPS, CVPR, ICCV, ACMMM, AAAI, IJCAI, etc. More information can be found at \url{https://liujiyuan13.github.io/}.
\end{IEEEbiography}

\begin{IEEEbiography}
[{\includegraphics[width=1in,height=1.15in,clip,keepaspectratio]{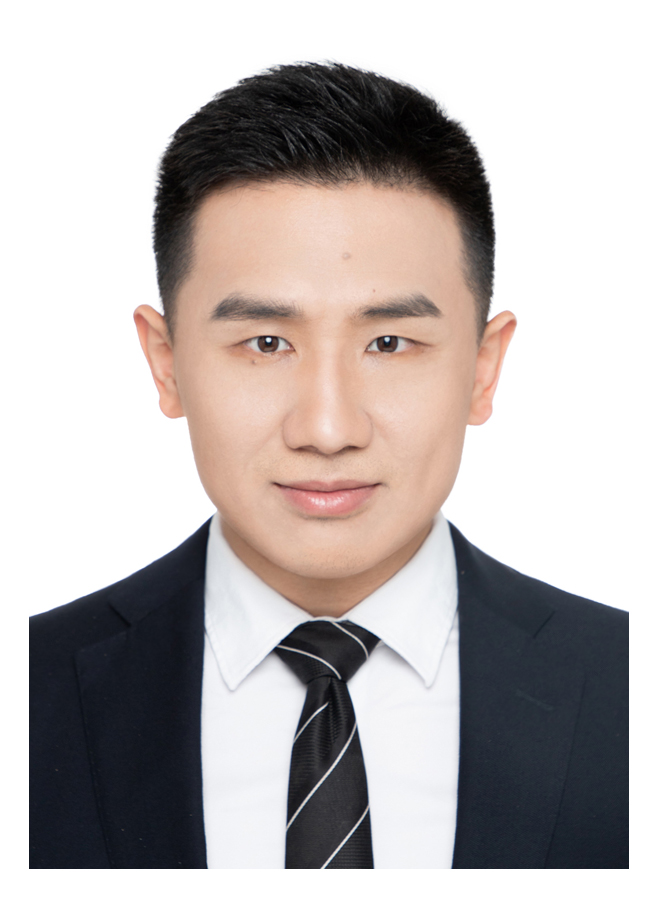}}]{Weixuan Liang} is pursuing his P.H.D degree in National University of Defense Technology (NUDT), China. He has authored or coauthored papers in journals and conferences, such as the IEEE Transactions on Knowledge and Data Engineering (TKDE), Annual Conference on Neural Information Processing Systems (NeurIPS), AAAI Conference on Artificial Intelligence (AAAI), and ACM Multimedia (ACM MM). His current research interests include multi-view clustering, learning theory, graph-based learning, and kernel methods.
\end{IEEEbiography}

\begin{IEEEbiography}[{\includegraphics[width=1in,height=1.10in,clip,keepaspectratio]{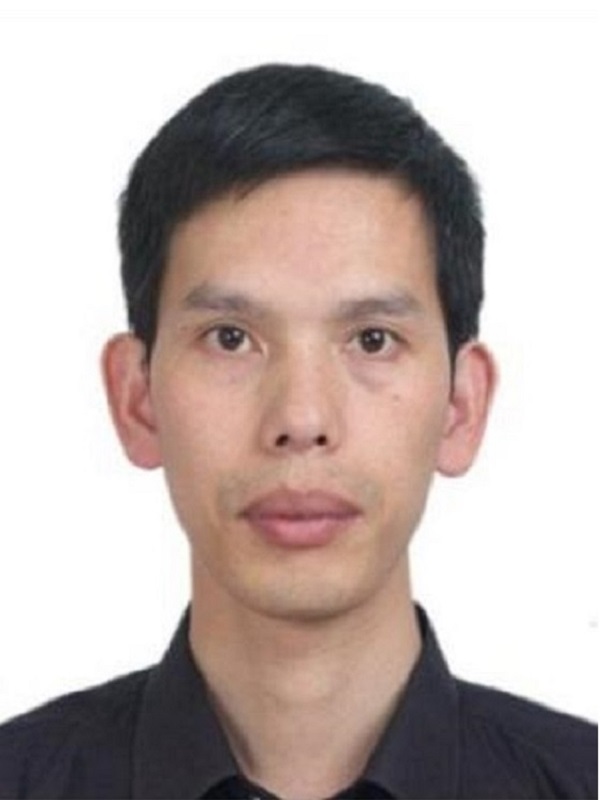}}]{En Zhu} received the Ph.D. degree from the National University of Defense Technology (NUDT), China. He is currently a Professor with the School of Computer Science, NUDT. He has published more than 60 peer-reviewed papers, including IEEE TRANSACTIONS ON CIRCUITS AND SYSTEMS FOR VIDEO TECHNOLOGY (TCSVT), IEEE TRANSACTIONS ON NEURAL NETWORKS AND LEARNING SYSTEMS (TNNLS), PR, AAAI, and IJCAI. His main research interests include pattern recognition, image processing, machine vision, and machine learning. He was awarded the China National Excellence Doctoral Dissertation.
\end{IEEEbiography}






\end{document}